\documentclass[Afour,sageh,times]{sagej}

\usepackage{moreverb,url}

\usepackage[draft, colorlinks,bookmarksopen,bookmarksnumbered,citecolor=black,urlcolor=black]{hyperref}

\usepackage{mathrsfs}
\usepackage[flushleft]{threeparttable}
\usepackage{tablefootnote}
\usepackage{subfigure}
\usepackage{hanging}
\usepackage{soul}

\usepackage{graphicx}
\usepackage[noend]{algpseudocode}
\usepackage{color}
\usepackage[dvipsnames]{xcolor}
\usepackage{amsmath,amssymb,amsfonts,dsfont}
\usepackage{bm}
\usepackage{setspace}
\usepackage{multicol}
\usepackage{multirow}
\usepackage{fancyhdr}
\usepackage{authblk}
\usepackage{caption}
\usepackage{tikz}
\usetikzlibrary{calc,arrows,decorations.markings}
\usepackage{gensymb}

\usepackage{algorithm}
\usepackage{algorithmicx}
\usepackage{algpseudocode}

\usepackage{gensymb}

\graphicspath{{../},{image}}

\newcommand{\depthIm}{\mathcal{D}}
\newcommand{\grasp}{\mathcal{G}}
\newcommand{\primSeg}{\mathcal{P}}
\newcommand{\primSegi}{{\mathcal{P}}_i}
\newcommand{\primSet}{I}
\newcommand{\primInd}{i}
\newcommand{\primShape}{P}
\newcommand{\graspInd}{\alpha}

\newcommand{\PS}{PS-CNN}
\newcommand{\algName}{\PS}

\begin{document}

\runninghead{Lin et al.}

\title{\LARGE \bf
Primitive Shape Recognition for Object Grasping}


\author{Yunzhi Lin, Chao Tang, Fu-Jen Chu, Ruinian Xu, and Patricio A.  Vela\affilnum{1}}

\affiliation{\affilnum{1}Intelligent Vision and Automation Laboratory
(IVALab), School of Electrical and Computer Engineering, Institute for Robotics
and Intelligent Machines; Georgia Institute of Technology, GA, USA.}

\corrauth{Yunzhi Lin, Intelligent Vision and Automation Laboratory (IVALab),
Georgia Institute of Technology,
North Ave NW,
Atlanta, GA 30332.}

\email{yunzhi.lin@gatech.edu}

%
%
\begin{abstract}
Shape informs how an object should be grasped, both in terms of 
where and how. As such, this paper describes a segmentation-based
architecture for decomposing objects sensed with a depth camera
into multiple primitive shapes, along with a post-processing pipeline
for robotic grasping. 
Segmentation employs a deep network, called {\em PS-CNN}, trained on
synthetic data with 6 classes of primitive shapes and generated using a
simulation engine.  Each primitive shape is designed with parametrized
grasp families, permitting the pipeline to identify multiple grasp
candidates per shape region. The grasps are rank ordered, with the first
feasible one chosen for execution. For task-free grasping of individual
objects, the method achieves a 94.2\% success rate placing it amongst
the top performing grasp methods when compared to top-down and
$SE(3)$-based approaches.  Additional tests involving variable
viewpoints and clutter demonstrate robustness to setup.  For
task-oriented grasping, {\em PS-CNN} achieves a 93.0\% success rate.
Overall, the outcomes support the hypothesis that explicitly encoding
shape primitives within a grasping pipeline should boost grasping
performance, including task-free and task-relevant grasp prediction.
\end{abstract}


\maketitle

\section{Introduction}

%
Manipulation is traditionally a multi-step task consisting of sequential
actions applied to an object--as determined from perception and planning
modules--to be executed by a robotic arm with a gripper. 
Although it may be taken for granted that object grasping is an easy
task based on human grasping capabilities at a young age, high accuracy
robot grasping remains a challenging problem due to the diversity of
objects that a robotic arm could grasp and the contact dynamics
associated to specific robot hand designs.  
Deep learning has emerged as a strong approach for addressing these
issues, however additional research can hopefully better illuminate how
to design grasping strategies for robotic manipulators.

In contrast to object-centric approaches, which require creating 3D models
or scanning large quantities of real objects 
\citep{wohlkinger20123dnet,calli2017yale} and also require a high
accuracy detector, people intuit that objects with similar shapes can
be grasped in an object-agnostic manner and that household objects seen
in daily life are composed of a limited set of canonical shapes.  Based
on these two observations, this paper explores the related assertion
that {\em shape is an important property to explicitly encode within a
grasping pipeline.} Doing so will promote grasping success.



To validate the assertion, this paper describes the design and
evaluation of a more geometric and explicitly shape-centric approach to
grasp recognition based on primitive shapes.
Primitive shapes offer a powerful means to alleviate data inefficiency
and annotation insufficiency
\citep{yamanobe2010grasp,jain2016grasp,tobin2017domain}
by abstracting target objects to primitive shapes with {\em a priori} known
grasp options. 
Previous primitive shape methods represented objects as a single shape
from a small library \citep{jain2016grasp,eppner2013grasping},
applied model-based rules to deconstruct objects \citep{yamanobe2010grasp} 
or employed Reeb graphs for decomposition \citep{aleotti20123d}. 
As a result, they do not handle novel objects with unmodeled geometry or
being the union of primitive shapes. Thus, traditional shape methods
neither obtain a high object grasping success rate, nor show the
potential to solve more advanced tasks including 
grasping in clutter and bin picking \citep{mahler2019learning},
and task-oriented grasping \citep{fang2018learning}. Meanwhile, current
state-of-the-art grasping methods using deep networks relegate shape to
an implicitly derived internal concept captured through training.

\begin{figure}[t]
  \centering
  \vspace*{0.06in}
  \begin{tikzpicture}[inner sep = 0pt, outer sep = 0pt]
    \node[anchor=south west] (fnC) at (0in,0in)
      {\includegraphics[height=1.7in,clip=true,trim=0in 0.2in 0in 0.35in]{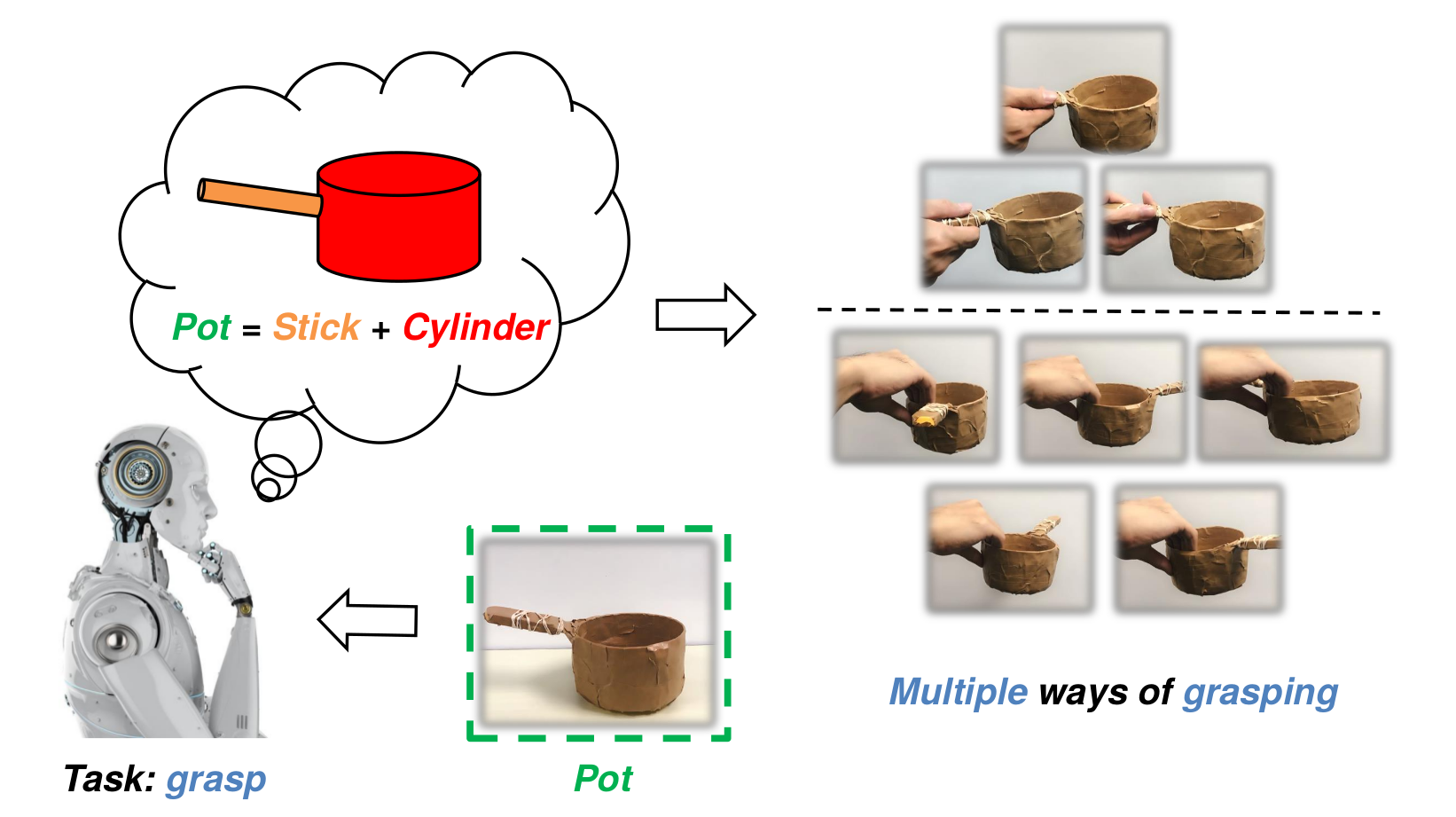}};
  \end{tikzpicture}
  \caption{Concept graph of the proposed approach: the robot first infers the primitive shape decomposition of captured household object (e.g. a pot is decomposed into a stick and a cylinder). Grasp family is then applied to each shape, enabling to grasp the object part appropriately under specific contexts. \label{fig:concept}} 
  \vspace*{-0.25in}
\end{figure}

Working from \cite{yamanobe2010grasp}'s model decomposition idea, 
this paper studies the role of a primitive shapes detector in supporting
generalized grasp strategies for household objects. 
Figure \ref{annotated} depicts the general concept, with a robot whose
objective is to grasp the pot. The pot consists of a handle attached to
a container, which maps to stick and cylinder object classes.
Grasping the handle can be performed through any grasp from a
continuously parametrized set of grasp poses, and likewise for the
container component. Awareness of shape informs knowledge of potential
grasps for an object. More explicitly stated, {\em this manuscript explores
the effectiveness of primitive shape object decompositions towards
the task of grasping objects by leveraging recent results in deep
network segmentation to achieve the decomposition. It aims to show that
explicit recognition of shape provides an effective means to generate
grasp candidates.}

%
\begin{figure*}[t!]
  \centering
  \begin{tikzpicture}[inner sep = 0pt, outer sep = 0pt]
    \node[anchor=south west] at (0in,0in)
      {{\includegraphics[width=1\textwidth]{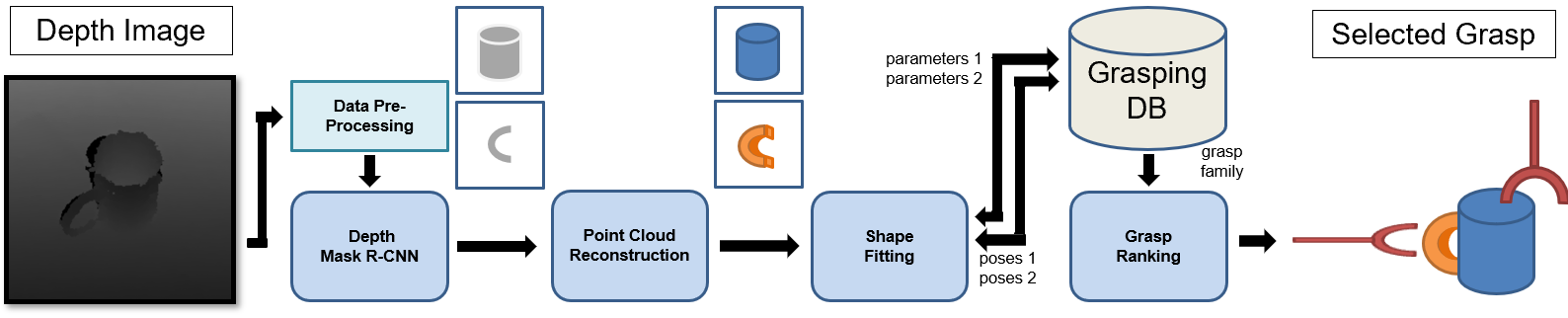}}};
   \node[yshift=-4pt] at (1.6in,0in) {\small (a)};
    \node[yshift=-4pt] at (3.9in,0in) {\small (b)};
    \node[yshift=-4pt] at (5.05in,0in) {\small (c)};
    \node at (2.6in,1.2in) {$\primSeg_{i_1}$};
    \node at (2.6in,0.8in) {$\primSeg_{i_2}$};
    \node at (3.75in,1.2in) {$\primShape_{j_1}$};
    \node at (3.75in,0.8in) {$\primShape_{j_2}$};
  \end{tikzpicture}
  \caption{The proposed deep network, segmentation-based pipeline. 
    (a) From monocular depth input, objects are segmented into primitive
    shape classes, with the object to grasp extracted and converted into
    primitive shape point clouds. 
    (b) Shape pose and parameter estimation, grasp scoring, and grasp
    selection processes follow. The best matching shape pose and parameter
    per primitive shape are identified. 
   \textbf{(c)} Candidate grasps are priority ranked and tested for
     feasibility with the first feasible grasp chosen for physical robot
     executions.
   \label{fig:pipeline}}
\end{figure*}
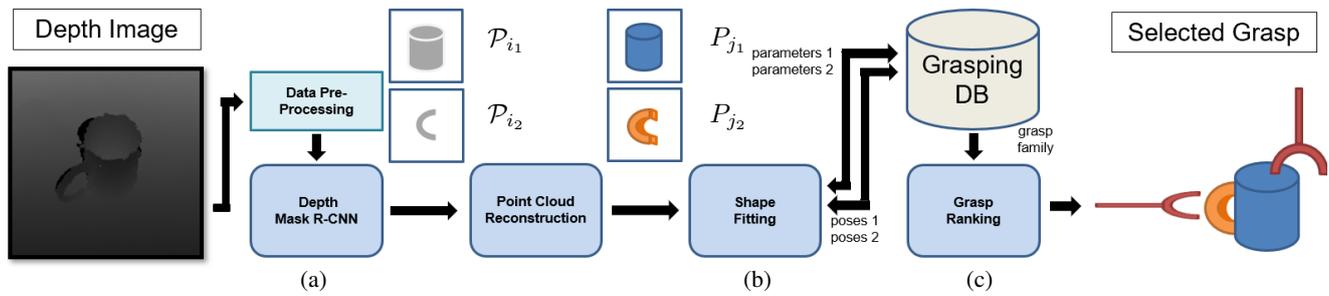

\subsubsection{Overall Approach.} 
Figure \ref{fig:pipeline} depicts the processing flow for the primitive shape informed grasping pipeline described in this manuscript.
Upon receipt of an input depth image, the image contents are segmented
into distinct primitive shape categories. Detection and segmentation of
these categories exploits the Mask R-CNN \citep{he2017mask} instance
segmentation deep network trained to segment a depth image according to
the primitive shapes it contains.  The processing that follows employs
traditional grasp methods to generate a grasp candidate from a
segmented primitive shape. The post-segmentation process involves shape
parameter estimation, pose recovery, grasp family specification, and
grasp prioritization or selection via rank scoring.  After this final
step, the selected grasp is planned and executed.  When coupled to a
robotic arm, the pipeline identifies primitive shapes within the scene
to recover suitable grasp options for objects associated to the
primitive shapes.

\subsubsection{Training Data.} 
Deep learning for visual processing is data-intensive and requires
annotated input/output datasets for training. Consequently, deep
learning solutions are fully specified only when the deep network
structure and learning processes are specified.  Robotics research
mitigates the cost of manual annotation for deep learning through the
use of programmed simulations \citep{tremblay2018falling} with automated,
known ground-truth annotation capabilities. When combined with domain
randomization \citep{tobin2017domain,tobin2018domain}, the resulting
training datasets permit translation of the learnt relationships to
input data capture from actual sensors. 
In some cases, there is still a performance gap arising from the distribution mismatch between the training data set and the deployment
input data.  Reducing the mismatch is known as domain alignment
\citep{sankaranarayanan2018learning}.  Here, data distribution gaps are tackled through both
randomization and alignment.

Alignment is addressed in a bi-directional manner by corrupting the
simulated data and denoising the depth image data. The intent is to
introduce real sensor artifacts that cannot be removed in the simulated
images, and to denoise the real images to match the simulated images.
Randomization is addressed through the use of synthetic ground truth
data based on parametrized sets of primitive shape classes and their
parametrically defined grasp families.  The parametric models permit
automatic synthesis of diverse input scenes with their matching output
grasps, thereby avoiding extensive manual annotation.
The shape classes are sufficiently representative of object parts
associated to household objects yet low enough in cardinality that grasp
family modeling is quick. 
Simulated domain randomization and data corruption generates a large
synthetic dataset composed of different primitive shapes combinations,
quantities, and layouts.  The automated ground truth generation
strategy rapidly generates input/output data.

\subsubsection{Contribution.}
This study extends a previous conference version \citep{lin2020using}
beyond proof-of-concept and focuses on the value of explicitly encoding
shape information. It improves the dataset generation strategy and
replaces the post-processing components downstream of shape
segmentation with improved model-based approaches based on best
practice.  Experimental confirmation includes more categories of objects and
additional grasping tests that demonstrate the robustness and value of
primitive shapes.

%
The main contribution is the testing and confirmation of the hypothesis
that explicitly encoding shape primitives within a grasping pipeline
boosts grasping performance.
In particular, executing the deep-learning enabled pipeline of
Figure \ref{fig:pipeline}, which is detailed in Sections
\ref{sec:exp_grasping}\,-\,\ref{sec:exp_result}, on an actual 7-DOF
robotic manipulator (Section \ref{sec:exp_grasping}) confirms 
that shape primitives provide an effective means to generate
grasp candidates.
{The grasping pipeline achieves one of the highest success rates
on individual objects amongst parallel plate gripper, or equivalent,
manipulation systems using a single-shot grasp recognition strategy.}
Published works with higher success rates employ an eye-in-hand visual
servoing strategy (using multiple images), a suction cup,
or a dual-gripper strategy with a suction cup and a parallel plate gripper,
with the suction cup being the dominant factor leading to the high
success rates \citep{mahler2019learning}.  
{As such, the first of them
exploits the temporal regularity possible through repeated measurements
and predictions, while the latter two address the grasping and grasp
closure problem through an alternative grasping mechanism.}  
{Additional studies demonstrate primitive shapes have some robustness under different operational settings (i.e., camera viewpoint and light clutter). More importantly, primitive shapes enable the recognition of specific shape regions, which paves the way for task-oriented grasping or purposeful manipulation.}
The studies provide evidence in favor of an explicit shape
primitive recognition module for grasp recognition strategies. They also
suggest areas for improvement in the primitive shape grasping
pipeline.

\section{Related Work}
%
Grasping is a mechanical process involving intentional contact between a
robot end-effector and an object. It can be mathematically described
from prior knowledge of the target object's properties (geometry, hardness,
etc.), the hand contact model, and the hand dynamics 
\citep{Murray_Li1994,bicchi2000robotic}. 
Mechanics-based approaches with analytical solutions work well for some
objects but cannot successfully apply to other, often novel, target objects
\citep{Tung_Kak_1996,Prattichizzo_Malvezzi_Gabiccini_Bicchi_2012,Rosales_Suarez_Gabiccini_Bicchi_2012}.
Grasp scoring methods augmented by point cloud processing overcome some
of these limitations \citep{aleotti20123d,ten2017grasp}. However, analytical
models cannot cover the parametric variation associated to all potential
objects to grasp.  In hopes of permitting more robust generalization over
model-based methods, efforts have gone into data-driven approaches using
machine learning \citep{Bohg_Morales_Asfour_Kragic_2014} as well as
combined approaches employing analytic scoring with {\em image-to-grasp}
learning \citep{mahler2017dex}.  Contemporary state-of-the-art grasping
solutions employ deep learning \citep{caldera2018review} and leverage
available training data. 

\subsection{Deep Learning Strategies}
%
%
Deep learning strategies primarily take one of three types.
%
%
The first type exploits the strong detection or classification capabilities of
deep networks to recognize candidate structured grasping representations
\citep{watson2017real,Park_Chun_2018,Chu_Xu_Vela_2018,%
satish2019onpolicy}. The most
common representation is the $SE(2) \times \mathbb{R}^2$ grasp
representation associated to a parallel plate gripper. 
As an oriented rectangle~\citep{asif2019densely} or a pair of
keypoints~\citep{wang2021double, xu2021gknet}, there is an underlying
assumption of a top-down view such that the grasp representation
directly maps to a top-down grasp to execute.
As a perception problem, recognition accuracy is high (around 95\% or higher), 
with a nearly comparable performance during robotic implementation
(around 90\% and up). Training involves image/grasp datasets obtained
from manual annotation \citep{lenz2015deep} or 
simulated grasping \citep{depierre2018jacquard,satish2019onpolicy}. 
Related to this category is, DexNet \citep{mahler2017dex}, which uses
random sampling and analytical scoring followed by deep network regression
to output refined, learnt grasp quality scores for grasp selection. By
using simulation with an imitation learning methodology, tens of thousands
to millions of annotations support DexNet regression training. 
Success rates vary from 80\% to 93\% depending on the task.
Another trend is to work in the full $SE(3) \times \mathbb{R}^3$ space.
It removes the constraint of 2D grasping that requires grasping aligned
with the tabletop frame or the image plan and optical axis.
As a result, 3D grasp supports greater grasp diversity and more advanced
manipulation tasks. Recent approaches process the 3D point cloud input
either in a discriminative or a generative manner. Discriminative
methods~\citep{ten2017grasp, liang2019pointnetgpd} sample grasp
candidates and then rank them according to grasp quality network or
other heuristic evaluations. On the other hand, generative
approaches~\citep{mousavian20196, qin2020s4g, ni2020pointnet++,
wu2020grasp} directly regress 6-DoF grasp configurations. 
They claim to be a robust choice in cluttered scenes. 

The second type replaces annotation or simulation with actual
experiential data coupled to deep network reinforcement learning methods
\citep{levine2018learning, zeng2018learning}, when sufficient resources
are available.  It is a preferred option when the ground truth
annotations are hard to obtain. Grasping solutions based on simulation
or experience tend to be configuration-dependent; they usually learn for
specific robot and camera setups. 

The third type is based on object detection or recognition
\citep{Bohg_Morales_Asfour_Kragic_2014}. Again, recent work employs deep
learning to detect objects and relative poses to inform grasp
planning~\citep{tremblay2018deep, xiang2018posecnn,
peng2019pvnet, wang2019normalized}. Others may perform object-agnostic
scene segmentation to differentiate objects followed by a DexNet grasp
selection process~\citep{danielczuk2019segmenting} or a GraspNet grasp
generation process~\citep{mousavian20196}.  Like
\cite{danielczuk2019segmenting}, this paper focuses on where to find
candidate grasps as opposed to grasp quality scoring.

%
%

\begin{figure}[t]
  \centering
  \vspace*{0.06in}
  \begin{tikzpicture}[inner sep = 0pt, outer sep = 0pt]
    \node[anchor=south west] (fnC) at (0in,0in)
      {\includegraphics[height=0.95in,clip=true,trim=0in 0.25in 0in 0.35in]{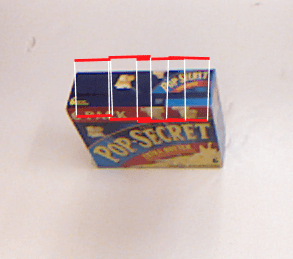}};
    \node[anchor=south west,xshift=2pt] (fnJ) at (fnC.south east)
      {\includegraphics[height=0.95in,clip=true,trim=0in 0.33in 0in 0.27in]{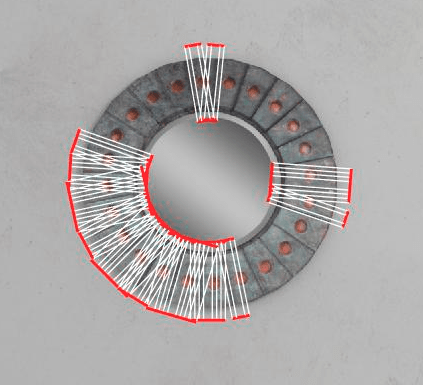}};
    \node[anchor=north west,xshift=2pt,yshift=1em] at (fnC.south west) 
      {\small Cornell};
    \node[anchor=north west,xshift=2pt,yshift=1em] at (fnJ.south west) 
      {\small Jacquard};
  \end{tikzpicture}
  \caption{Grasp annotation data with missing grasp candidates. \label{annotated}} 
  \vspace*{-0.25in}
\end{figure}

%
%
%
%
Although deep learning grasp methods have achieved great success shown in the aforementioned discussion, they still suffer from two related problems:
sparse grasp annotations or insufficiently rich data (i.e., covariate shift).  
The former can be seen in Figure \ref{annotated}, which shows an image from
the Cornell dataset \citep{jiang2011efficient} and another from
the Jacquard dataset \citep{depierre2018jacquard}. Both lack annotations in
graspable regions due to missing manual annotation or a false negative in
the simulated scenario (either due to poor sampling or incorrect physics).
Sampling insufficiency can be seen in \cite{satish2019onpolicy},
where the DexNet training policy was augmented with an improved (on-policy)
oracle to provide a richer sampling space. Yet, sampling from a continuous
space is bound to under-represent the space of possible options, especially for higher parametric grasp space dimensions.

%
\subsection{Primitive Shapes}
This paper proposes to more fully consider shape primitives
\citep{miller2003automatic} due to their known, parametrized grasp families
\citep{yamanobe2010grasp,shiraki2014modeling}.  The parameterized families
provide a continuum of grasp options rather than a sparse sampling. A
complex object can be decomposed into parts representing distinct surface
categories based on established primitives.

The primitive shapes idea has a long history dating back to the 1970s,
when \cite{marr1978representation} illustrated the
concept of decomposing a human 3D model into combinations of cylinders.
Since then, researchers in the computer vision community have advanced on
the idea \citep{kaiser2019survey}, with approaches based 
on RANSAC-enabled extensions with randomized fits with maximal consensus \citep{schnabel2007efficient}, 
primitive-driven region growing algorithms that extract connected
components with neighborhood information \citep{attene2010hierarchical},
and 
clustering approaches for detecting simple geometric primitives \citep{holz2011real}. 
More recent progress further demonstrating the potential of primitive
shapes introduces different deep neural network designs
such as a vowel-based network to generate primitives
\citep{tulsiani2017learning}, an image-based network combined with
Conditional Random Field (CRF) \citep{kalogerakis20173d}, and point-based
network with differentiable primitive model estimator
\citep{li2019supervised}. However, those methods generally took a 3D model
or a point cloud randomly sampled on a model's surface as the input
\citep{paschalidou2019superquadrics,paschalidou2020learning}, even
required fine-grained labels to train the networks
\citep{sharma2020parsenet}, which is not as feasible to achieve for
grasping.

Past research in the robotic grasping field explored shape primitive approaches in
the context of point cloud processing \citep{aleotti20123d} 
and fitting for the cases of superquadric~\citep{Goldfeder_Allen_Lackner_Pelossof_2007,vezzani2017grasping,xia2018reasonable, hachiuma2020pose} and box
surfaces~\citep{huebner2008selection}.  
In most cases, the entire sensor-scanned object point cloud is required for shape segmentation and grasp selection.
Existing approaches also model objects as individual primitive shapes~\citep{eppner2013grasping,jain2016grasp,vezzani2017grasping,fang2018learning},
which do not exploit the potential of primitive shapes to generalize to
unseen/novel objects. 
Deep network approaches for shape primitive segmentation to
inform grasping do not appear to be well studied.  


\subsection{Domain Randomization}

Deep learning is data-intensive, which translates to unwieldy manual
annotation efforts. As noted earlier, robotics research employs simulation
and synthetic data generation whenever possible 
\citep{levine2018learning,zeng2018learning}, e.g., 
DexNet uses simulation to generate grasp quality test data
\citep{mahler2017dex,satish2019onpolicy}.
Thus, the data generation component is considered part of the solution.
Improvements in simulation and rendering fidelity
~\citep{denninger2019blenderproc, morrical2021nvisii} permit the
generation of large scale photo-realistic datasets with demonstrated
performance improvements~\citep{hodavn2019photorealistic}.
However, simulation introduces a domain gap, or distribution shift,
because the training signals and the true input signals differ. The gap
can be overcome through domain randomization and consideration of how
the two signals differ \citep{prakash2019structured}.  

Domain randomization is a prevalent method in the robotics simulation
field. It strives to permit networks trained from simulated data to 
apply to real-world domains without additional real image input~\citep{tobin2017domain,peng2018sim,OpenAI2018LearningDI}. 
Since domains can be parametrically complex, structured domain
randomization helps to regulate multiple parameters in a more
organized manner through a probability distribution map
\citep{prakash2019structured}.
In some cases, it might be possible to establish which parameters are
more useful to bridge the domain gap through a guided domain
randomization approach \citep{zakharov2019deceptionnet}. 

Typically, domain randomization solutions are designed for RGB inputs. 
This paper makes use of depth images as inputs, which requires modifying
the existing methods or identifying depth-specific approaches. 
Relative to color imagery, depth imagery has a less severe gap more
readily addressed.  One source of the mismatch for modern robotics
simulation engines \citep{rohmer2013v,blender} is that they render and
simulate virtual environments quite cleanly, often leading to sensor
imagery with less defects than real sensors.  Depth sensors like the
Kinect v1 loses details and introduces noise during the depth capture
\citep{planche2017depthsynth,Sweeney2019ASA}, while also having shadow
effects from the active illumination, leading to a persistent
distribution shift between the simulated depth data and sensed depth
data.  Mechanisms to correct for this shift are needed for deep learning
to be used for primitive shape segmentation.  Here, addressing domain
shift focuses on the variation of the placed objects as a randomization
strategy and employs bi-directional alignment to reduce sensing-based
gaps. By incorporating a specially-designed primitive shapes family and
targeting household objects, the ground truth generation strategy
reduces the domain shift.

%
%

\section{Grasping from Primitive Shapes Recognition}

\begin{figure*}[t]
  \begin{tikzpicture}[inner sep=0pt, outer sep=0pt]
     \node[anchor=south west] (Full) at (0in,0in)
       {\includegraphics[height=1.10in,clip=true,trim=0.25in 0.25in 0in 0.25in]{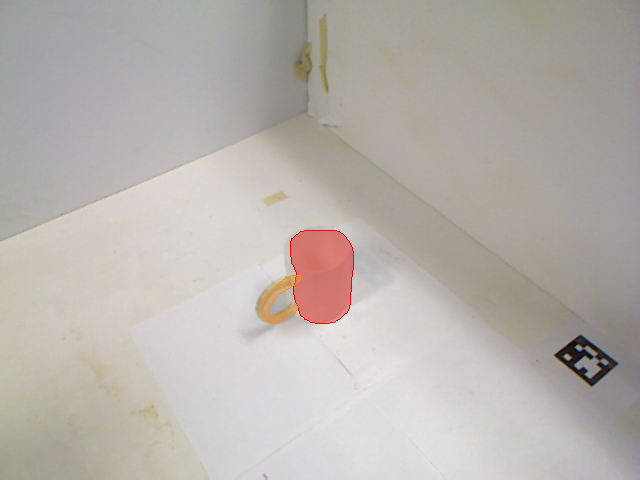}};
     \node[anchor=south west,xshift=3pt,yshift=1pt] at (Full.south west)
     {\small Setup (no zoom)};

     \node[anchor=north west,xshift=2pt] (SS) at (Full.north east)
       {\includegraphics[height=0.535in,clip=true,trim=3.75in 1.2in 2.63in 3.5in]{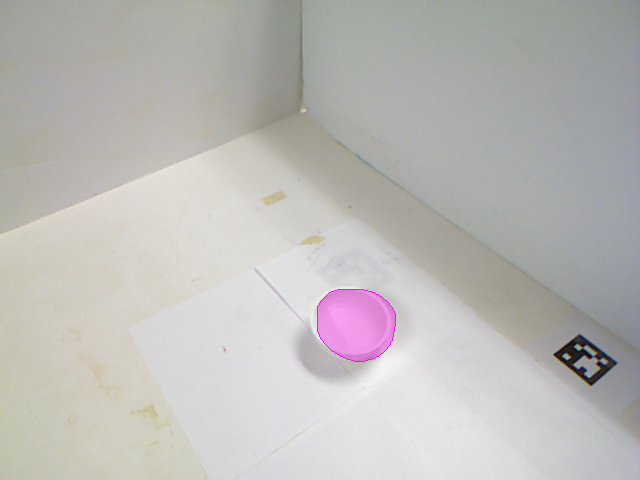}};
     \node[anchor=south west,xshift=2pt] (Cyl) at (Full.south east)
       {\includegraphics[height=0.535in,clip=true,trim=3.2in 1.2in 2.93in 3.3in]{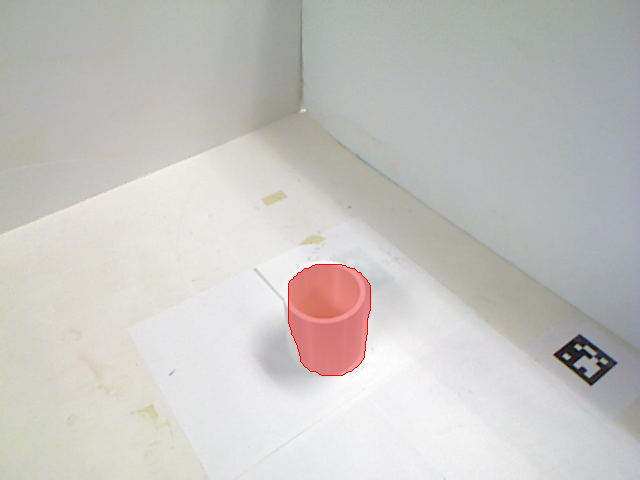}};
      \node[anchor=south west,xshift=2pt] (Bask) at (Cyl.south east)
       {\includegraphics[height=0.535in,clip=true,trim=2.7in 1.6in 3.1in 2.3in]{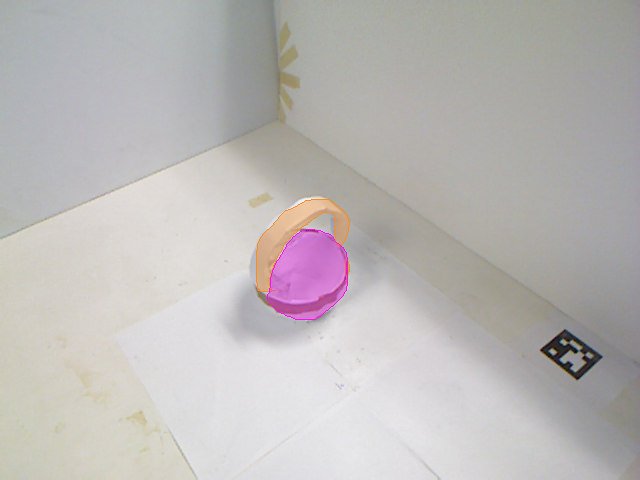}};      
     \node[anchor=north west,xshift=2pt] (Pot) at (SS.north east)
       {\includegraphics[height=0.535in,clip=true,trim=3.2in 2.2in 3.15in 2.2in]{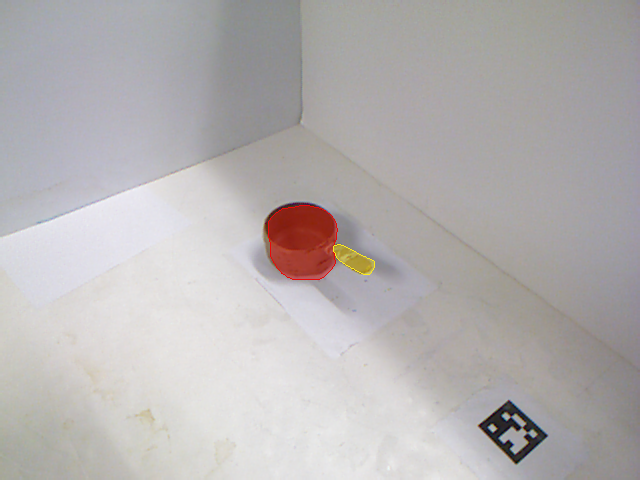}};

     \node[anchor=south east] (M2) at (0.99\textwidth,0in)
       {\includegraphics[height=1.1in,clip=true,trim=1.1in 0in 2.55in 2in]{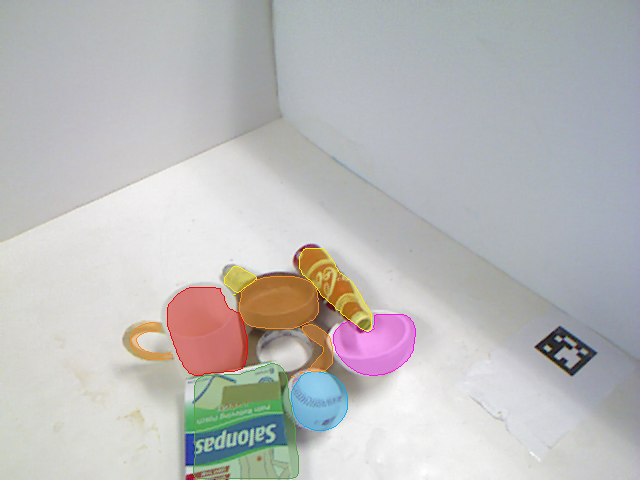}};
     \node[anchor=south east,xshift=-2pt] (M1) at (M2.south west)
       {\includegraphics[height=1.1in,clip=true,trim=2.6in 1.0in 1.8in 2in]{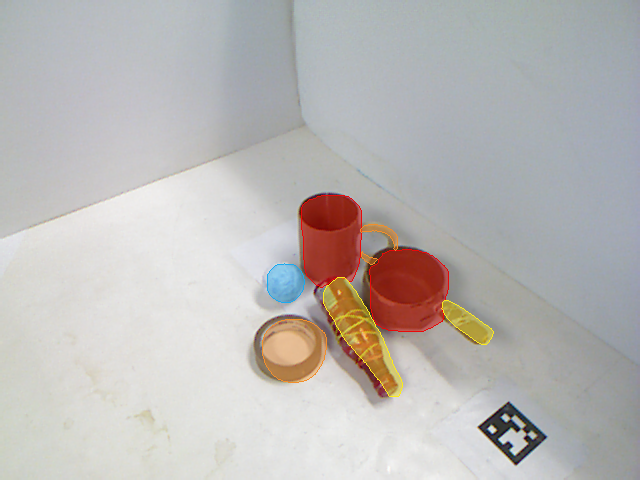}};
     \node[anchor=south east,xshift=-2pt] (M0) at (M1.south west)
       {\includegraphics[height=1.1in,clip=true,trim=2.2in 0in 2.45in 2.9in]{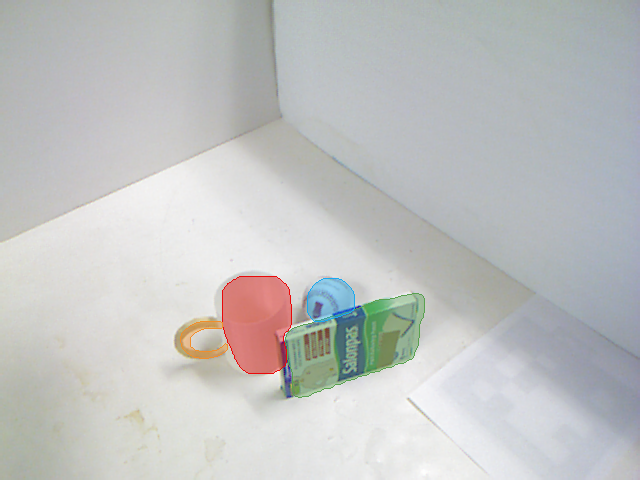}};
   \end{tikzpicture}
   \caption{Sample segmentation outcomes for test scenarios consisting of
   individual and multiple objects (zoomed and cropped images). 
     \label{fig:segdemo}}
    \vspace*{-0.20in}
\end{figure*}

The intent behind this investigation is to explore the potential value of
using deep networks to segment a scene according to the surface primitives
contained within it, thereby establishing {\em where} to grasp.
Once the object or region to grasp is known, post-processing recovers
the shape geometry and the grasp family associated to the shape, to
establish {\em how} to grasp.  
The state-of-the-art instance segmentation deep network Mask R-CNN
\citep{he2017mask} serves as the backbone network for converting depth
images into primitive shape segmentation images. 
Importantly, a synthetically generated training set using only shape
primitives in concert with domain randomization \citep{tobin2017domain}
covers a large set of scene visualizations. 
The ability to decompose unseen/novel objects into distinct shape regions,
often with explicitly distinct manipulation affordances, permits
task-oriented grasping \citep{fang2018learning}.

The vision-based robotic grasping problem here presumes the existence of a
depth image $\depthIm$ $\in$ $\mathbb{R}^{H\times W}$ 
($H$ and $W$ are image height and width) capturing a scene containing an
object to grab.  The objective is to abstract the scene into a set of
primitive shapes and generate grasp configurations from them.  
A complete solution involves establishing a routine or process, $f$,
mapping the depth image $\depthIm$ to a grasp 
$\mathcal{G} = f(\mathcal{D}) \in SE(3)$.
The grasp configuration $\grasp \in SE(3)$ specifies the final pose in the
world frame of the end-effector. 

%
Per Figure \ref{fig:pipeline}, the process is divided into three stages. In
the first stage, the depth image $\depthIm$ gets segmented according to
defined primitive shape categories indexed by the set $\primSet$. 
The primitive shape segmentation images are $\primSegi$ for $\primInd \in
\primSet$.
The segmentation $\primSegi$ and the depth image $\depthIm$ generate
segmented point clouds in 3D space for the primitive surfaces attached to
the label $\primInd$. In the second stage, when the grasp target is
established, the surface primitives attached to the target grasp region
are converted into a corresponding set of primitive shapes $\primShape_{j}$
in 3D space, where $j$ indexes the different surface primitive segments. 
In the third stage, the parametrized grasp families of the surface primitives
are used to generate grasp configurations $\grasp_{\graspInd}$ for
$\alpha \in \mathbb{N}^+$.  A prioritization process leads to rank-ordered
grasps with the first feasible grasp being the one to execute. This
section details the three stages and the deep network training method.


%
%
\section{Primitive Shape Segmentation}
%
%
The proposed approach hypothesizes that commonly seen household objects
can be decomposed into one or more primitive shapes. After studying
several household object datasets
\citep{calli2017yale,chen2003visual,funkhouser2003search,shilane2004princeton}, 
the chosen primitive shapes were:
{\em Cylinder}, 
{\em Ring}, 
{\em Stick}, 
{\em Sphere},
{\em Semi-sphere}, and 
{\em Cuboid}.
A deep learning procedure will be described for creating a network based
on Mask-RCNN \citep{he2017mask} with primitive shape segmentation outputs
from depth image inputs. Doing so requires providing both the annotated
training set, and the training technique.
Sections \ref{datagen} and \ref{domain_alignment} detail method used to
synthetically generate depth images and known segmentations from the
parametric shape classes.  
Section \ref{training} describes the training procedure for generating
the Mask-RCNN network that will decompose an input depth image into a set
of segmentations reflecting hypothesized primitive shapes. 
To gain a sense for the output structure of the trained primitive shape
segmentation network (\PS), Figure \ref{fig:segdemo}\, depicts several
segmentations for different input depth images overlaid on the
corresponding, cropped RGB images. The color coding is 
red: {\em Cylinder}, 
orange: {\em Ring}, 
yellow: {\em Stick}, 
blue: {\em Sphere},
purple: {\em Semi-sphere}, 
and
green: {\em Cuboid}.
For individual, sparsely distributed, and clustered objects, 
{\PS} captures the primitive shape regions and classes of the sensed objects. 

\begin{table}[t]
  \centering
  \caption{Primitive Shape Classes\label{tab:primtive_shape_design}}
  \setlength\tabcolsep{2pt}
  \begin{tabular}{|c|c|c|}
    \hline
    Class  & Parameters  & Range (unit: cm) \\ 
    \hline 
    %
    \hline
     Cylinder& 
    $(r_{in},r_{out}^{*},h)$  & 
    \parbox{1.35in}{
      \vspace*{-1.1ex}
      \begin{equation} \nonumber
      \begin{split}
        r_{in} &\in [3,7], \ \sigma_{1} = 0.3 \\ 
        h & \in [5,10], \ \sigma_{2} = 0.5 
      \end{split}
      \end{equation}
      \vspace*{-2.2ex}} 
 
    \\ 
    \hline
    Ring & 
      $(r_{in},r_{out}^{*},h)$ & 
    \parbox{1.35in}{
      \vspace*{-1.1ex}
      \begin{equation} \nonumber
      \begin{split}
        r_{in}  &\in [1.4,4], \ \sigma_{1} = 0.2 \\ 
        h   &  \in [0.8,2.4], \ \sigma_{2} = 0.1
      \end{split}
      \end{equation}
    \vspace*{-2.2ex}} 
    \\
    \hline
    Stick & 
    $(r_{in},r_{out}^{*},h)$ & 
      \parbox{1.35in}{
      \vspace*{-1.1ex}
      \begin{equation} \nonumber
      \begin{split}
        r_{in} & \in [0.8, 1.5],\ \sigma_{1} = 0.1\\ 
        h & \in [4, 10],\ \sigma_{2} = 0.5 
      \end{split}
      \end{equation} 
    \vspace*{-1.6ex}} \\

    \hline
    Sphere  & 
    $r$ & $r \in [2, 5],\ \sigma=0.2$\rule{0pt}{2.25ex}
    \\ \hline
    Semi-sphere & 
      $r$ & $r \in [2, 5],\ \sigma=0.2$\rule{0pt}{2.25ex}
      \\ 
      
    \hline
   Cuboid  & 
    $(h,w,d)$ & 
    \parbox{1.5in}{
      \vspace*{-1.1ex}
      \begin{equation} \nonumber
      \begin{split}
        h, w, d &\in [2, 12], \ \sigma =0.5
      \end{split}
      \end{equation} } \\

      \hline
        
  \end{tabular}
  \centerline{$r$ - radius, $r_{in}$ - inner radius, 
              $r_{out}$ - outer radius}
  \centerline{$h$ - height, $w$ - width, $d$ - depth} 
  \centerline{*To simplify generation, $r_{out}$ is set to be 1.15 times $r_{in}$.} 
\end{table}
%


\subsection{Dataset Generation}
\label{datagen}
%
%

\begin{figure}[t]
  \centering
    \fbox{\includegraphics[width=0.93\columnwidth]{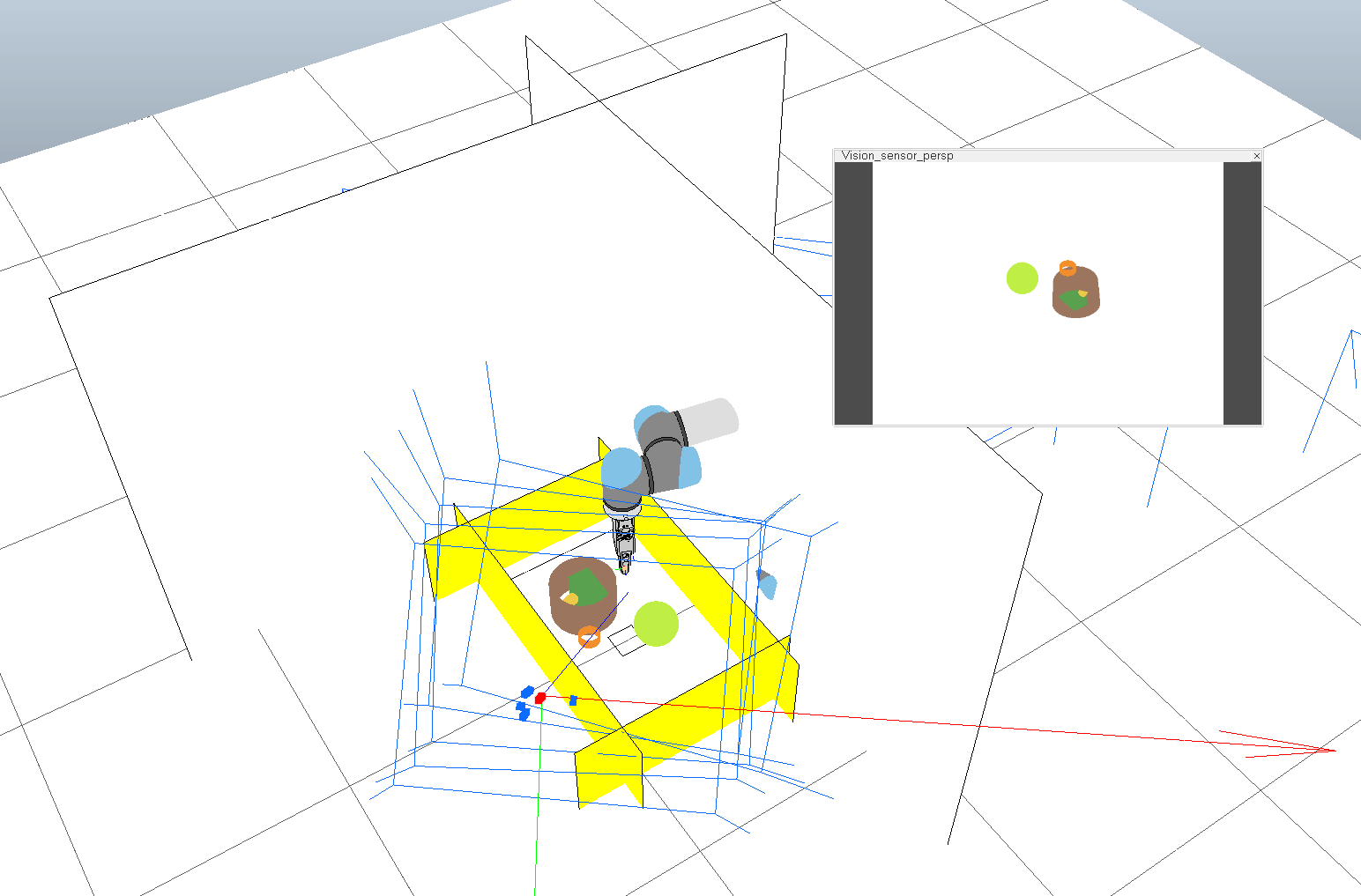}}
  \caption{Experimental setup in the simulator, where the red represents the principle camera and the blue represents the assistant cameras.\label{fig:exp_simulation_setting}}
\end{figure}

The value of using primitive shapes is in the ability to automatically
synthesize a vast library of shapes through gridded sampling within the
parametric domain of each class.  This section describes what parameters
are varied to synthetically generate input imagery with known
segmentation outcomes, from the V-REP simulation
software~\citep{rohmer2013v}, see Figure \ref{fig:exp_simulation_setting}.
Based on the hypothesis that a dataset with diverse combinations of primitive 
shapes could induce learning generalizable to household objects, 
the dataset generation procedure consists of the following degrees of
freedom: 
(1) camera pose setup: a principle camera fixed to a specific
pose or canonical view while four other assistant cameras are
instantiated with random relative poses around the principle camera; 
(2) primitive shape parameters; 
(3) placement order; 
(4) placement flag, for which there are two options: one to place all
the objects close together, one to scatter them; 
(5) initial placement location assigned in a Gaussian distribution; 
(6) mode of placement, for which there are three modes: 
{\em free-fall}, {\em straight up from the table-top}, and 
{\em floating in the air}; and 
(7) initial orientation. 
%
%
%




\begin{figure*}[t]
  \centering
\tikzstyle{vecArrow} = [semithick, decoration={markings,mark=at position 1
with {\arrow[scale=1.25,semithick]{open triangle 60}}},
   double distance=3.4pt, shorten >= 6.75pt,
   preaction = {decorate},
   postaction = {draw,line width=3.4pt, white,shorten >= 6.5pt}]
\tikzstyle{innerWhite} = [semithick, white,line width=1.4pt, shorten >=
6.25pt]

  \begin{tikzpicture}[inner sep = 0pt, outer sep= 0pt]
    \node[anchor=south west] (SimOrig) at (0in,0in)
      {\includegraphics[width=0.15\textwidth]{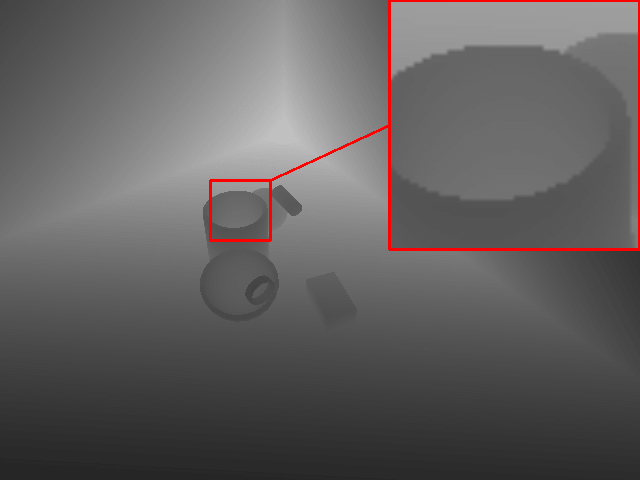}};
    \node[anchor=south west, xshift=0.75in] (SimFilt) at (SimOrig.south east)
      {\includegraphics[width=0.15\textwidth]{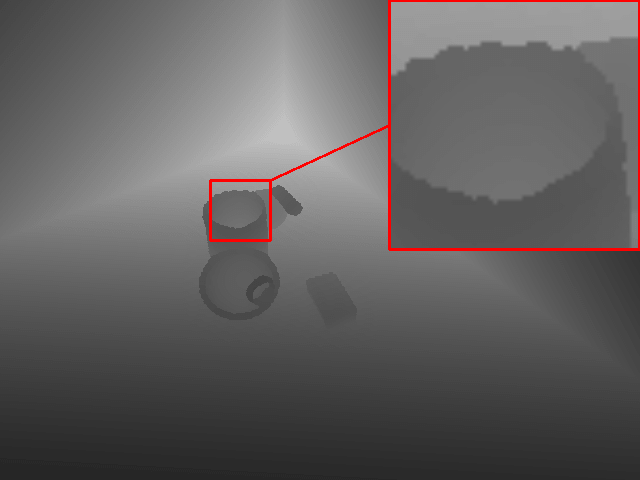}};

    \node[anchor=south east] (KinOrig) at ($(SimOrig.south
    east) + (0.75\textwidth,0in)$)
      {\includegraphics[width=0.15\textwidth]{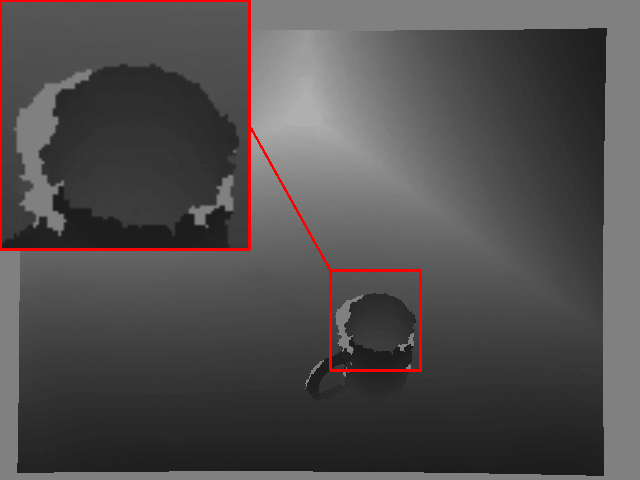}};
    \node[anchor=south east, xshift=-0.75in] (KinFilt) at (KinOrig.south west) 
      {\includegraphics[width=0.15\textwidth]{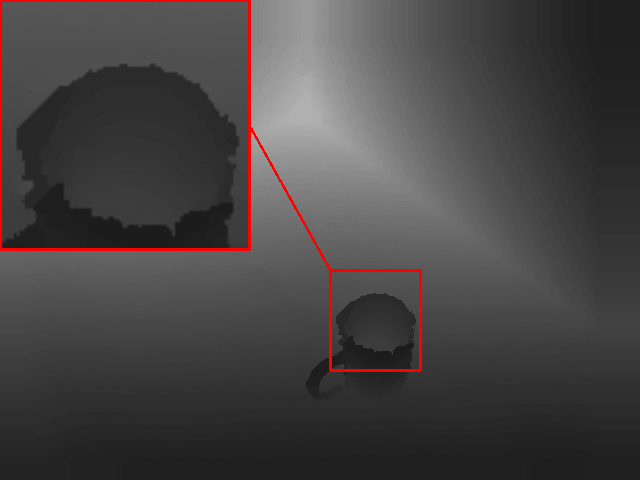}};

    \node[anchor=south west,xshift=2pt,yshift=2pt] at (SimOrig.south west)
      {\footnotesize \textcolor{white}{Simulated}};
    \node[anchor=south west,xshift=2pt,yshift=2pt] at (KinOrig.south west)
      {\footnotesize \textcolor{white}{Kinect}};
    \node[anchor=base,rotate=90,yshift=7pt] (clean) at (SimOrig.west)
      {\small Too Clean};
    \node[anchor=north,rotate=90,yshift=-5pt] (noisy) at (KinOrig.east)
      {\small Too Noisy};
    \node[anchor=center,rotate=90] at 
      ($(KinFilt.west)!0.5!(SimFilt.east)$) {\small Better Matched};

    \draw[vecArrow] ([xshift=5pt]SimOrig.east) to 
      node[above,xshift=-5pt,yshift=5pt,anchor=base]{\small Corrupt} 
      ([xshift=-5pt]SimFilt.west);
    \draw[vecArrow] ([xshift=-5pt]KinOrig.west) to 
      node[above,xshift=2pt,yshift=5pt,anchor=base]{\small Denoise} 
      ([xshift=5pt]KinFilt.east);

  \end{tikzpicture}
  \caption{Bi-directional image filtering to align training data and
    real data. An oil painting filter applied to training imagery simulates
    the noise of the Kinect depth sensor.  Temporal averaging and spatial
    median filtering regularize the Kinect depth image during run-time.
    \label{fig:oilpainting}}
\end{figure*}
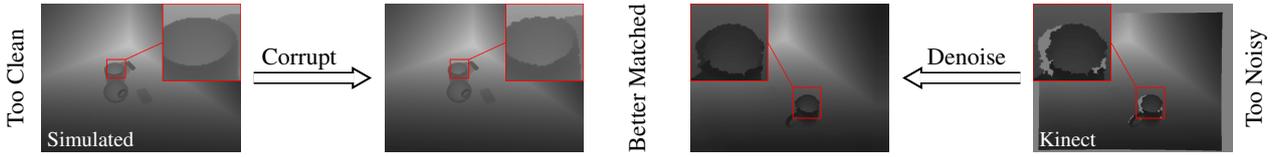

Regarding the camera pose parameters, the principle camera view provides
an input image from the expected perspective associated to the nominal
configuration. The addition of four assistant cameras generates varied
viewpoints should the configuration be off. It also provides some
variation equivalent to translating the object placement by a small
amount.
For each primitive shape, a specific parametric model determines its geometry.
Table \ref{tab:primtive_shape_design} lists the parameter coordinates
(middle column) for each primitive shape class. 
The {\em Cylinder}, {\em Ring}, and {\em Stick} are modeled to have
$r_{in}$, $r_{out}$ and $h$ parameters, where $r_{out}$ is fixed to be
1.15 times $r_{in}$ for simplicity. Those classes could be considered as
three subsets of cylinder-like shapes with different radius-to-height
ratios. 
The {\em Sphere} and {\em Semi-sphere} are only defined by a radius value. 
The {\em Cuboid} is modeled by its three main edges.
All of the parameters are limited to a particular range, grid discretized
according to $\sigma$.
Every scene generated has one randomly generated instance of each
primitive class, with the insertion order randomly determined. 

Once the core parameters for the camera and the primitive shapes have
been established, the next step is to place the shapes within the scene.
The placement flag is randomly chosen between the two options
following a ratio of 4:1, indicating a preference for objects to be
placed close together to generate more complex combinations. The
scatter option often isolates the objects to create loose clutter. 
To place objects close to each other, the first option assigns all the
objects the same mean placement location, while the second option would
individually assign the objects a mean placement location within the
scene. The actual placement of the objects will be based on 
The initial placement location is generated according to a Gaussian
distribution where the mean is set to be the placement location center
and the variance is set to be $0.15$m.  The placement mode is randomly
determined for each shape inserted. The initial orientation is uniformly
randomly determined within a bounding volume above the table-top.

Through random selection, 300k scenes of different primitive shapes
combinations are generated. 
A typical scene is shown in Figure \ref{fig:exp_simulation_setting}.
For each instance, RGB and depth image pairs
are collected. Shape color coding provides segmentation ground truth and primitive shape ID. 
The process samples a sufficiently rich set of visualized shapes once
self-occlusion and object-object occlusion effects are factored in.

%
%

\subsection{Domain Alignment between Simulation and Reality%
\label{domain_alignment}}

State-of-the-art simulators \citep{rohmer2013v,blender} benefit
data generation by automating data collection in virtual environments,
but do so using idealized physics or sensing. Some physical effects are
too burdensome to model. To alleviate this problem, the images from both
sources, the simulation and the depth sensor, are modified to better match.
The objective is to minimize the corrections applied, therefore the first
step was to reduce or eliminate the sources of discrepancies.
Discrepancy reduction involves configuring both environments to match,
which includes the camera's intrinsic and extrinsic parameters, and the
background scene. Comparing images from both sources, the main gap
remaining is the sensing noise introduced by the low-fidelity Kinect v1
depth sensor \citep{planche2017depthsynth,Sweeney2019ASA}.  
The Kinect has occlusion artifacts arising from the baseline between the
active illuminator and the imaging sensor, plus from measurement noise. The
denoising process includes temporal averaging, boundary cropping, and 
median filtering, in that order.
Once the Kinect depth imagery is denoised, the next step is to corrupt the
simulated depth imagery to better match the visual characteristics of the
Kinect. The primary source of uncertainty is at the depth edges or object
boundaries due to the properties of the illuminator/sensor combination. The
simulated imagery should be corrupted at these same locations. The
simulated environment has both a color image and a depth image. The color
image is designed to provide both the shape primitive label and the object
ID, thereby permitting the extraction of object-wise boundaries.  After
establishing the object boundary pixels, they are dilated to obtain an
enlarged object boundaries region, then an oil painting filter
\citep{sparavigna2010cld,mukherjee2014study} corrupts the depth data in
this region.  Considering that manipulation is only possible within a
certain region about the robotic arm, the depth values from both sources
were clipped and scaled to map to a common interval.  Figure
\ref{fig:oilpainting} depicts this bi-directional process showing how it
improves alignment between the two sources.

\subsection{Data Preprocessing and Training\label{training}}

Per \S \ref{domain_alignment}, the simulated depth images are re-scaled
then corrupted by a region-specific oil-painting filter.  To align with the
input of Mask R-CNN, the single depth channel is duplicated across the
three input channels. The ResNet-50-FPN as the backbone is trained from scratch on corrupted depth images in PyTorch 1.1.
It runs for 250,000 iterations with 4 images per mini-batch. The
primitive shape dataset is divided into a 83\%/17\% training and testing
splits.  The learning rate is set to 0.02 and divided by 10 at iterations
50000, 100000, and 180000. The workstation consists of a single NVIDIA
1080Ti (Pascal architecture) with cudnn-7.5 and cuda-9.0. Dataset
generation takes 95 hours, and training takes 55 hours (150 hours in total).

\subsection{Vision Evaluation Metrics}
Evaluation of the proposed pipeline consists of testing on novel input data
as purely a visual recognition problem, followed by testing on the
experimental system. For the visual segmentation evaluation, the
segmentation accuracy is computed by $F_{\beta}^{\omega}$ \citep{Margolin2014HowTE}:
\begin{equation} \label{F_measure}
  F_{\beta}^{\omega} = {(1 + \beta^{2})}
    \,{{Pr^{\omega}\cdot Rc^{\omega}}
    \Big/
    {(\beta^{2}\cdot Pr^{\omega}+Rc^{\omega})}}.
\end{equation}

where $Pr^{\omega}$ is weighted precision and $Rc^{\omega}$ is weighted recall. $\beta$ signifies the effectiveness of detection, which is set to 1 by default.


\subsection{Vision Result}


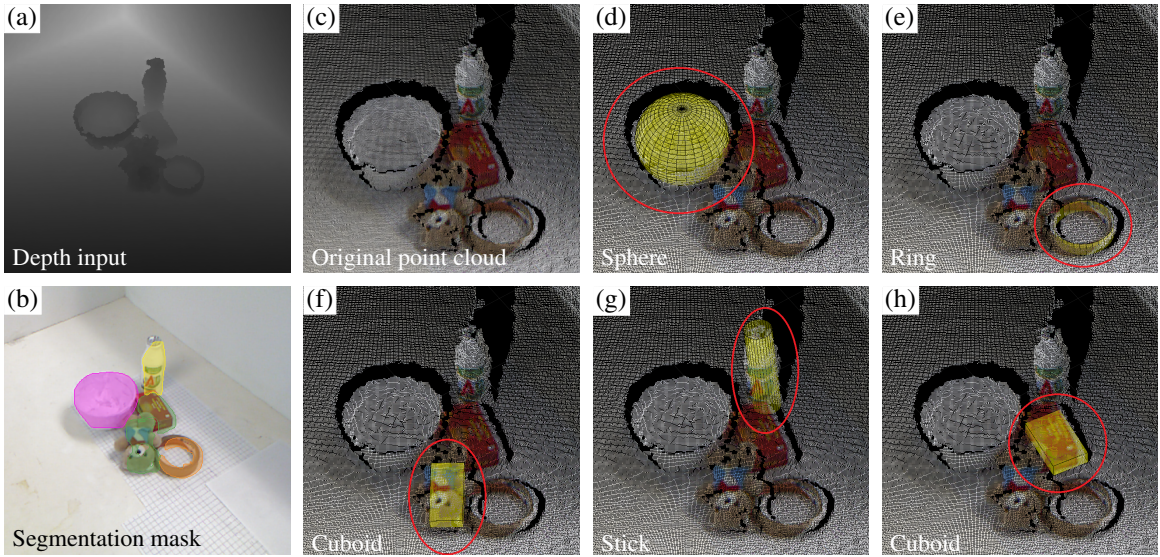
\begin{figure*}[t]
  \centering
  \begin{tikzpicture}[inner sep=0pt, outer sep=0pt]
    \node[anchor=south west] (depth) at (0in,0in)
      {\includegraphics[height=1.4in,clip=true,trim=2in 0.3in 1in 0.80in]{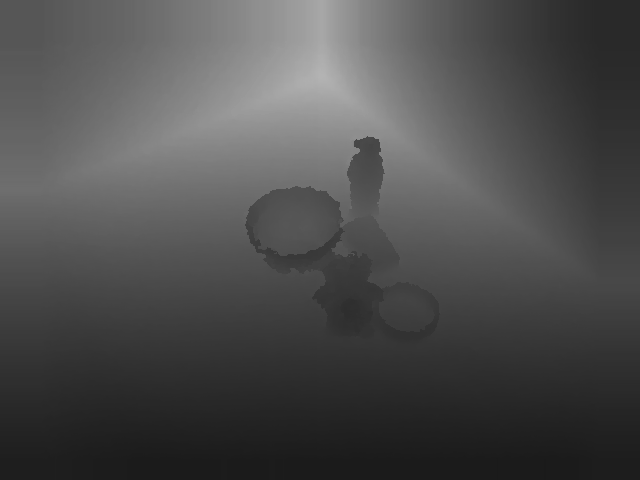}};
     \node[anchor=south west,text=white,xshift=3pt,yshift=1pt] at (depth.south west)
     {\small Depth input};
     
    \node[anchor=south west,xshift=5pt] (scene) at (depth.south east)
      {\includegraphics[height=1.4in,clip=true,trim=2in 0.3in 1in 0.80in]{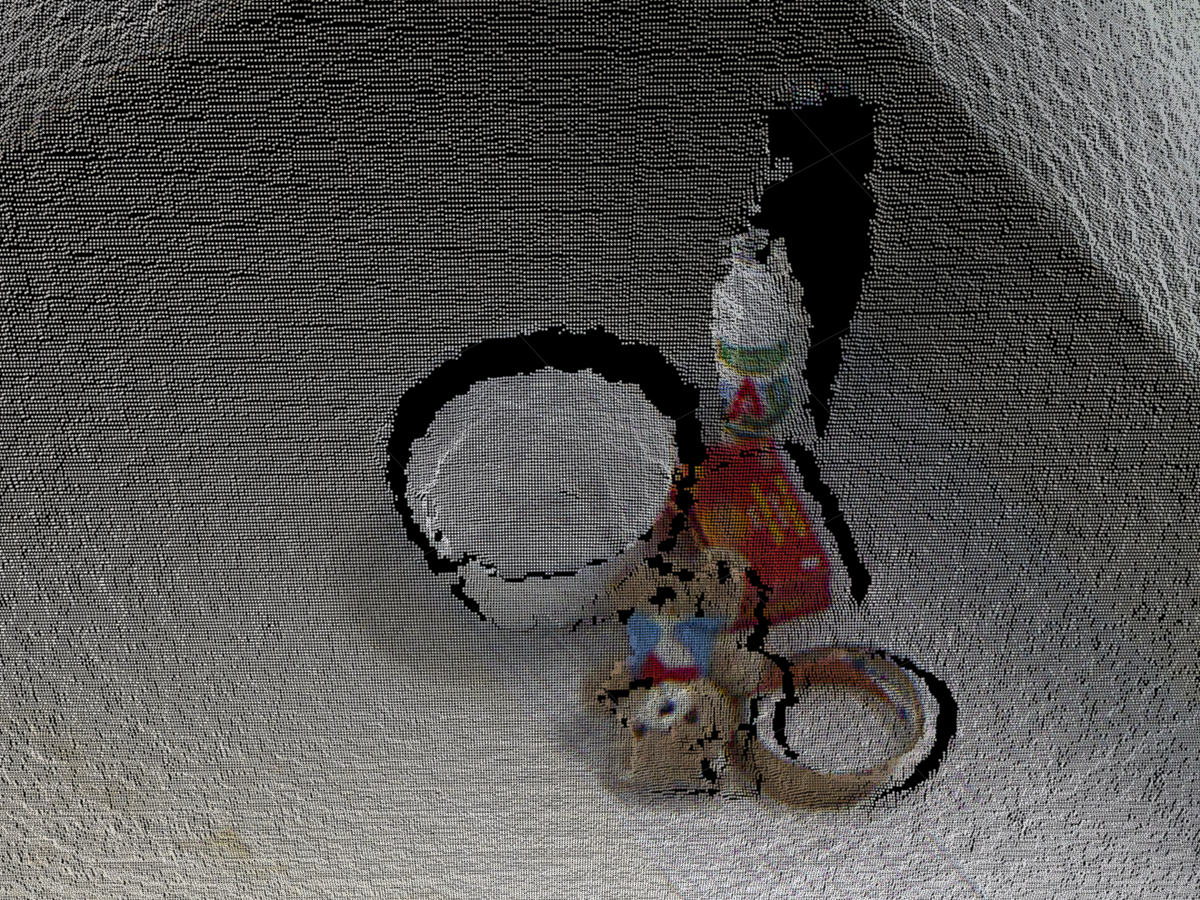}};
     \node[anchor=south west,text=white,xshift=3pt,yshift=1pt] at (scene.south west)
     {\small Original point cloud};
     
    \node[anchor=south west,xshift=5pt] (Bowl_0) at (scene.south east)
      {\includegraphics[height=1.4in,clip=true,trim=2in 0.3in 1in 0.80in]{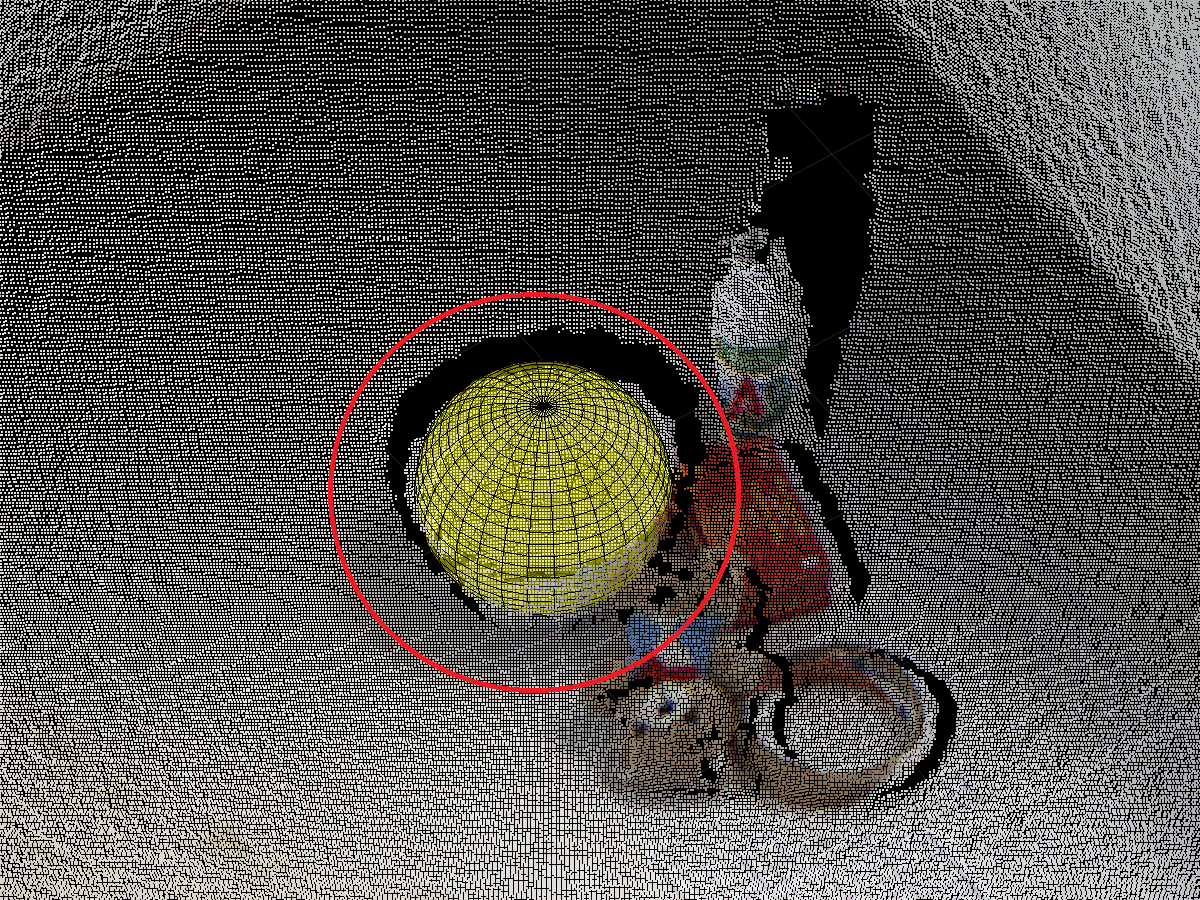}};
     \node[anchor=south west,text=white,xshift=3pt,yshift=1pt] at (Bowl_0.south west)
     {\small Sphere};

    \node[anchor=south west,xshift=5pt] (Ring_1) at (Bowl_0.south east)
      {\includegraphics[height=1.4in,clip=true,trim=2in 0.3in 1in 0.80in]{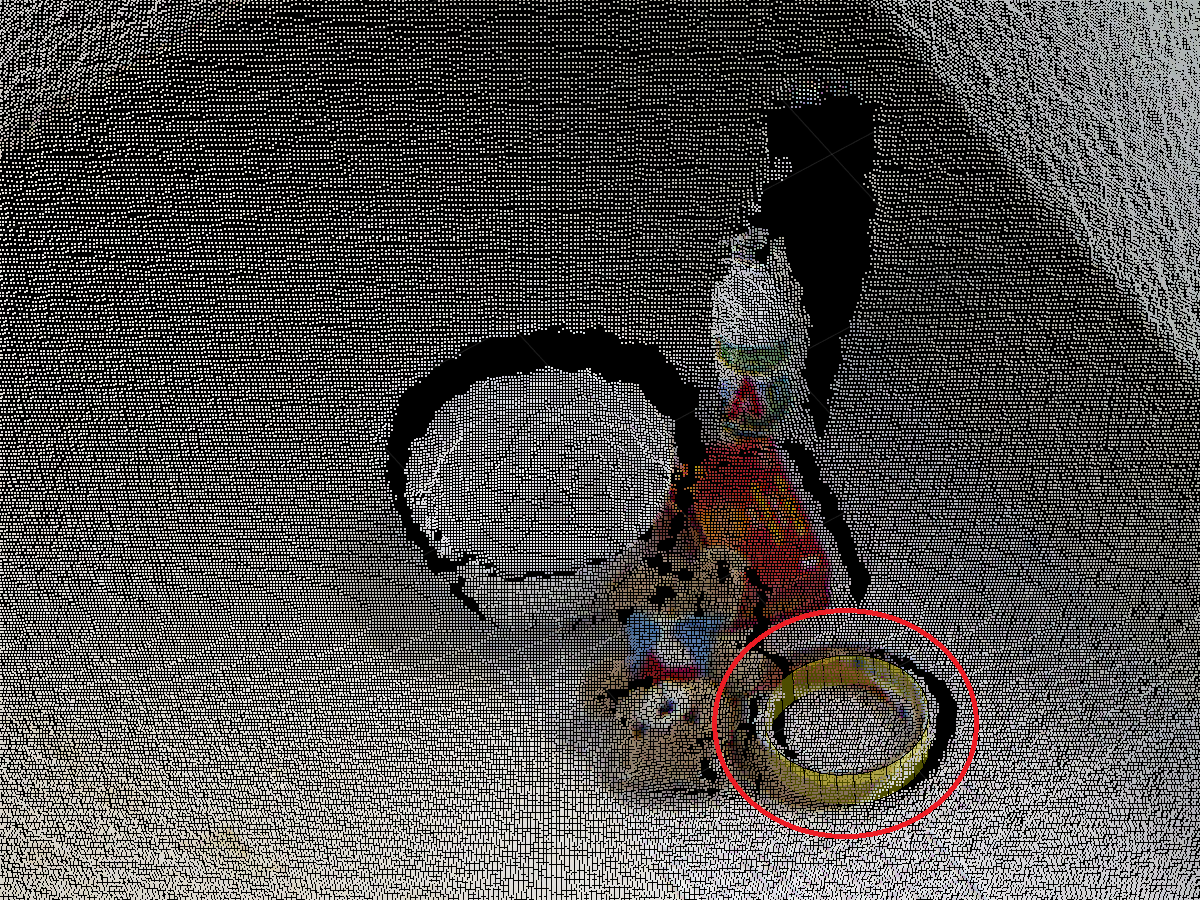}};
      \node[anchor=south west,text=white,xshift=3pt,yshift=1pt] at (Ring_1.south west)
     {\small Ring};

     \node[anchor=north west, yshift=-5pt] (visual) at (depth.south west)
      {\includegraphics[height=1.4in,clip=true,trim=2in 0.3in 1in 0.80in]{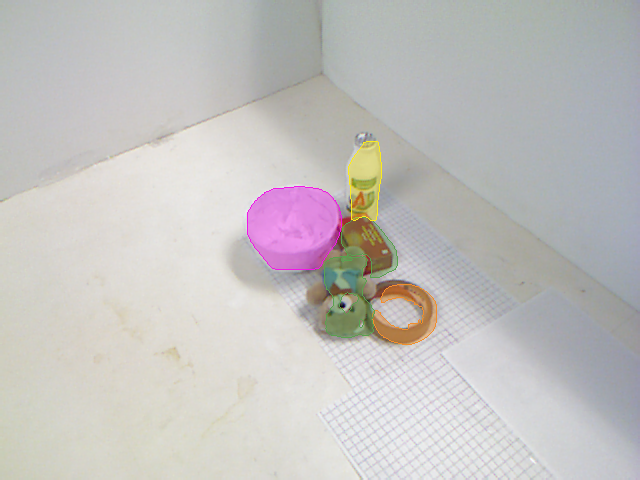}};
     \node[anchor=south west,text=black,xshift=3pt,yshift=1pt] at (visual.south west)
     {\small Segmentation mask};

     \node[anchor=south west,xshift=5pt] (Cuboid_2) at (visual.south east)
      {\includegraphics[height=1.4in,clip=true,trim=2in 0.3in 1in 0.80in]{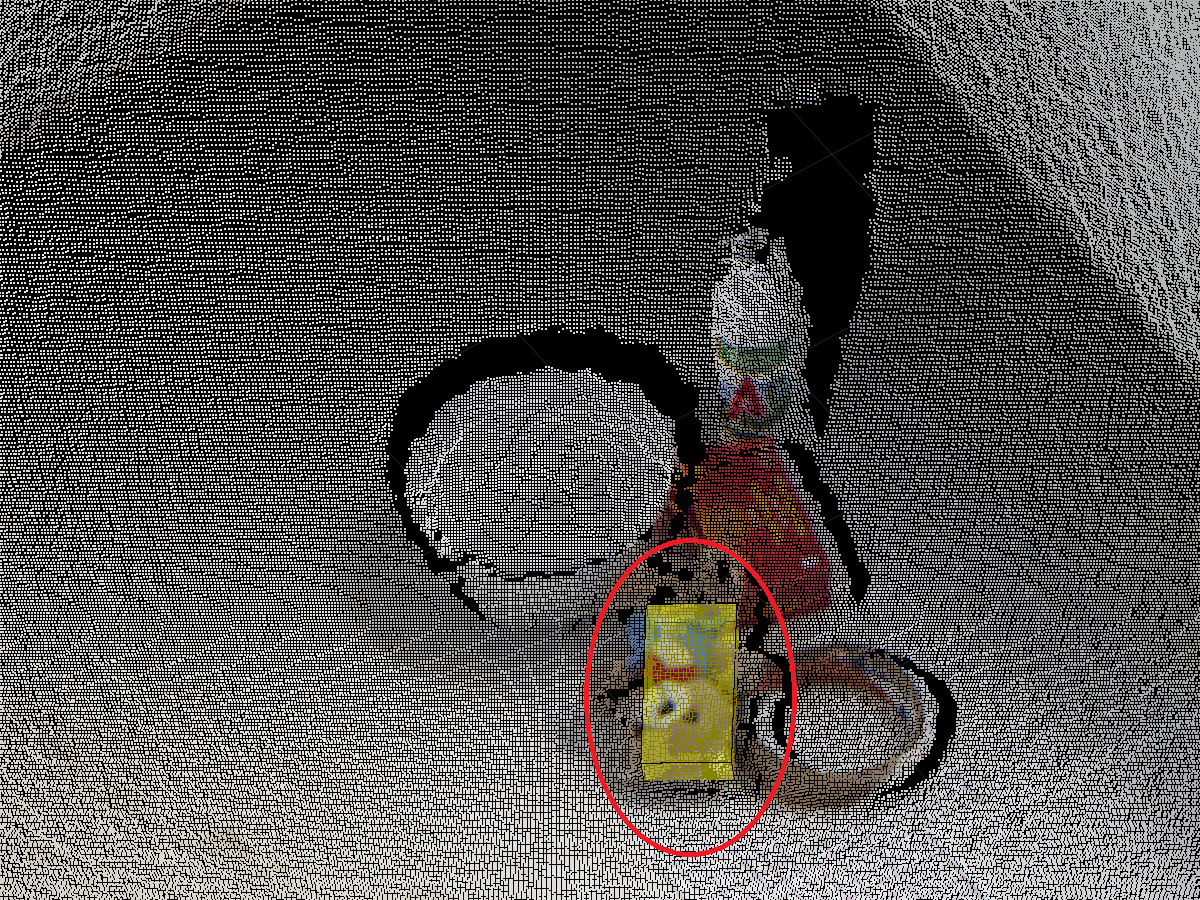}};
       \node[anchor=south west,text=white,xshift=3pt,yshift=1pt] at (Cuboid_2.south west)
     {\small Cuboid};

     \node[anchor=south west,xshift=5pt] (Stick_3) at (Cuboid_2.south east)
      {\includegraphics[height=1.4in,clip=true,trim=2in 0.3in 1in 0.80in]{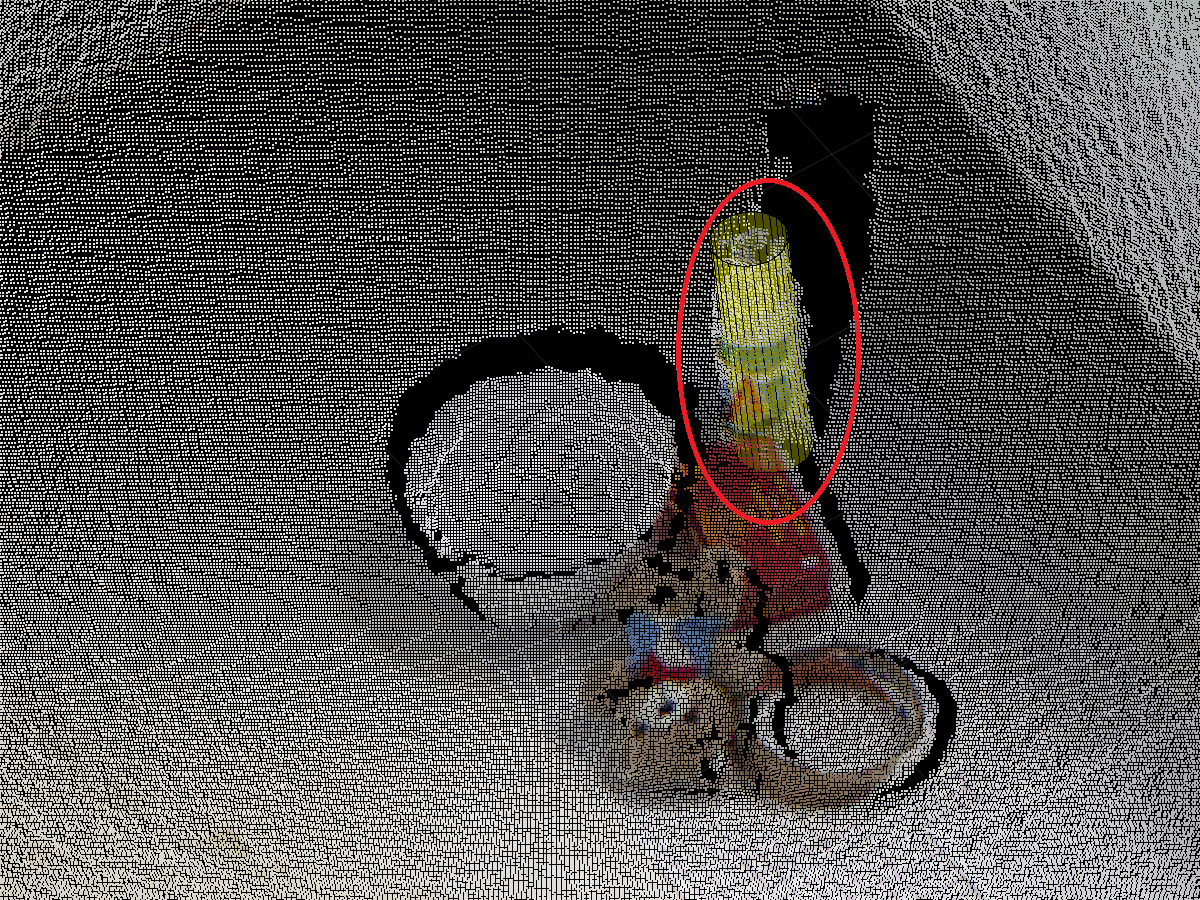}};
      \node[anchor=south west,text=white,xshift=3pt,yshift=1pt] at (Stick_3.south west)
     {\small Stick};

    \node[anchor=south west,xshift=5pt] (Cuboid_4) at (Stick_3.south east)
      {\includegraphics[height=1.4in,clip=true,trim=2in 0.3in 1in 0.80in]{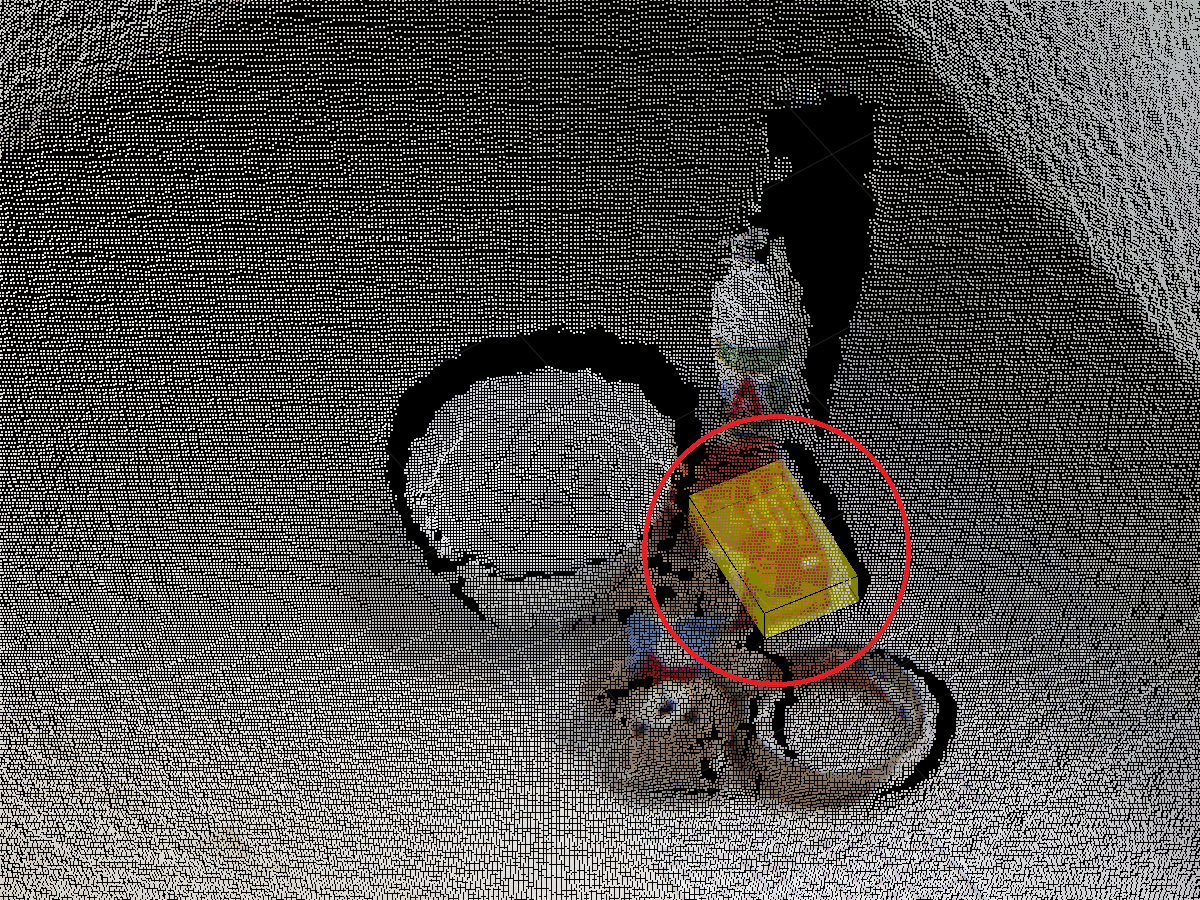}};
     \node[anchor=south west,text=white,xshift=3pt,yshift=1pt] at (Cuboid_4.south west)
     {\small Cuboid};

   \foreach \x / \y in {depth/a,visual/b,scene/c,Bowl_0/d,Ring_1/e,Cuboid_2/f,Stick_3/g,Cuboid_4/h} 
   {
    \node[anchor=north west,fill=white,draw=white,inner sep=1pt] at (\x.north west) {(\y)};
   }

  \end{tikzpicture}
    \caption{Shape fitting for a multi-object scene (zoomed and cropped images). The target shape fitting result is underlined by red circle.
    \label{fig:shape_fitting}}

\end{figure*}

\begin{table}[t]
\centering
  \caption{Performance of PS-CNN on 3D printed primitive shapes\label{tab:testseg}}
  
  \begin{tabular}{|l|c|c|}
    \hline
                & Original & Corrupted 
    \\ \hline \hline
    Cylinder    & 0.822    & {\bf 0.834}  \\ \hline
    Ring        & {\bf 0.907 }   & 0.902\\ \hline
     Stick       & 0.795    &{\bf 0.824}\\ \hline
    Sphere      & 0.591    &{\bf 0.842} \\ \hline
    Semi-sphere & 0.905    & {\bf 0.919 }\\ \hline
    Cuboid      & {\bf 0.916 }  & 0.915  \\ \hline \hline

    {\textbf All}  & 0.823    &{\bf 0.872 }
    \\ \hline
  \end{tabular}
  
\end{table}

Segmentation tests are performed on a set of 3D printed primitive shapes
and compared to manual segmentations. The tested implementations include
a network trained with the original simulated images (no corruption) and
with the oil-painting corrupted images. Per Table \ref{tab:testseg}\,, 
the $F_{\beta}^{\omega}$ segmentation accuracy improved from 0.823 to 0.872
(ranges over $[0,1]$),
with the primary improvement sources being for the
{\em sphere} shape class followed by the {\em stick class}. The
segmentation accuracy is sufficient to capture and label significant
portions of an object's graspable shape regions, see Figure
\ref{fig:segdemo}.

\section{From Primitive Shapes to Grasp Candidate}

\renewcommand{\algorithmicrequire}{\textbf{Input:}}  
\renewcommand{\algorithmicensure}{\textbf{Output:}} 

\subsection{Shape Fitting}
\label{sec:shape_fitting}
%
%
%
Given a target object region to grasp, the intersecting shape primitive
regions are collected and converted into separate partial point clouds.
For each region and its hypothesized shape class, Principal Component
Analysis (PCA) is first adopted to predict a coarse primary vector for
reference. Random Sample Consensus (RANSAC) then estimates the pose and
parameter of the partial point cloud over the shape model parameter
space of the primitive shape. The outputs inform an object shape model
for hypothesizing the candidate grasp family for each object. This matching step is depicted in Figure \ref{fig:pipeline}(b).  
{Although the estimated geometric primitives may not perfectly match
the target region, the downstream pipeline is robust to errors that do not
significantly impact the grasp's gripper properties (e.g., opening width).}
For more details on this standard shape fitting process, see the
Supplemental Material at \cite{graspPrimShape}. Figure
\ref{fig:shape_fitting} depicts the process of taking the depth image (a)
and generating a primitive shape segmentation (b), from which each shape
region is segmented out from the point cloud (c) to give a set of point
clouds (e)-(h) that correspond to an object or an object part to grasp.  In
the figure, all of the object maps to a single primitive shape. Figure
\ref{fig:graspdemo} and Section \ref{sec:res_taskoriented} includes cases
where objects consist of two primitive shape classes.

\begin{figure*}[t]
  \centering
  \vspace*{0.06in}
  \begin{tikzpicture}[inner sep=0pt, outer sep=0pt]
    
    \node[anchor=south west] (A) at (0in,0in)
      {\includegraphics[height=1.5in,clip=true,trim=1.5in 0.65in 1.5in 0.50in]{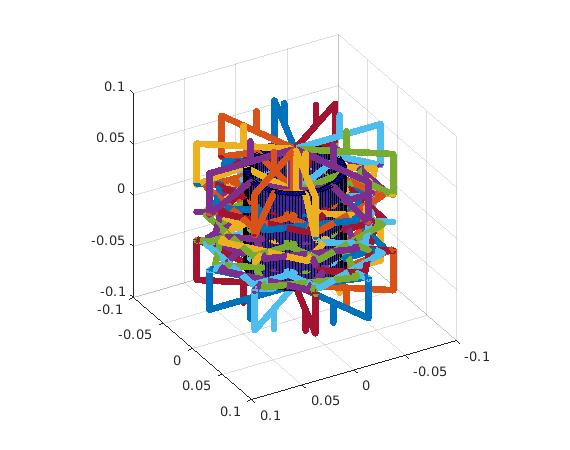}};
    
    \node[anchor=south west,xshift=5pt] (B) at (A.south east)
      {\includegraphics[height=1.5in,clip=true,trim=1.5in 0.65in 1.5in 0.50in]{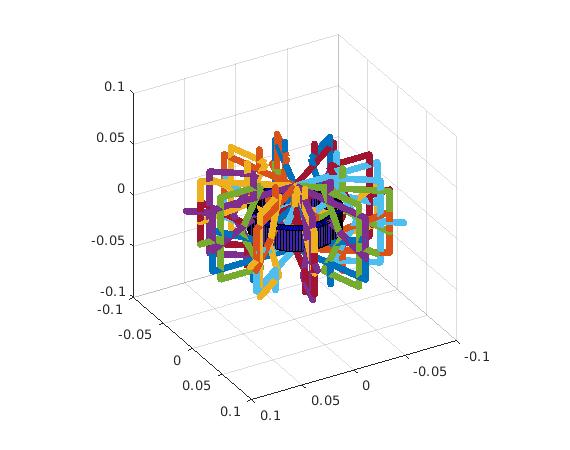}};
    
    \node[anchor=south west,xshift=5pt] (C) at (B.south east)
      {\includegraphics[height=1.5in,clip=true,trim=1.5in 0.65in 1.5in 0.50in]{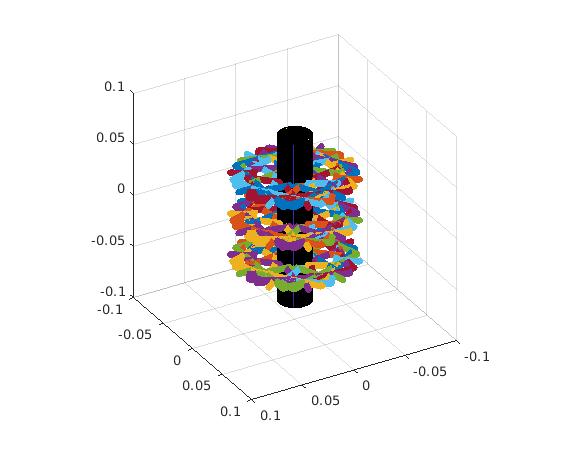}};

     \node[anchor=north west, yshift=-5pt] (D) at (A.south west)
      {\includegraphics[height=1.5in,clip=true,trim=1.5in 0.65in 1.5in 0.50in]{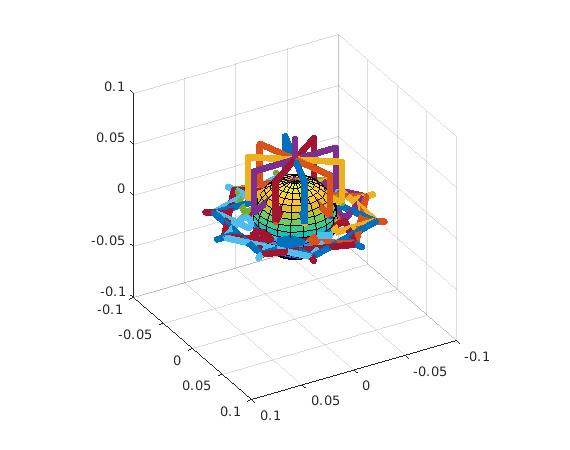}};
    
     \node[anchor=south west,xshift=5pt] (E) at (D.south east)
      {\includegraphics[height=1.5in,clip=true,trim=1.5in 0.65in 1.5in 0.50in]{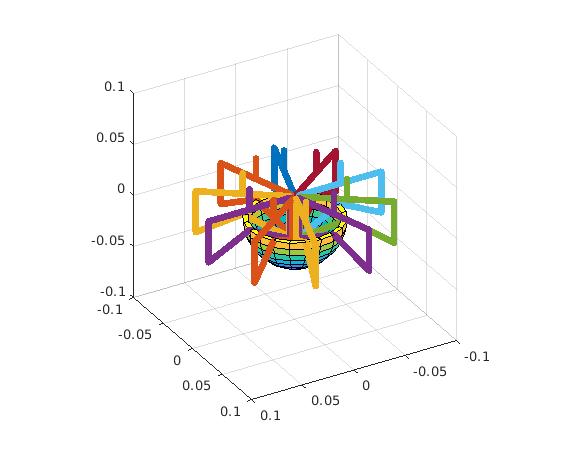}};
    
    \node[anchor=south west,xshift=5pt] (F) at (E.south east)
      {\includegraphics[height=1.5in,clip=true,trim=1.5in 0.65in 1.5in 0.50in]{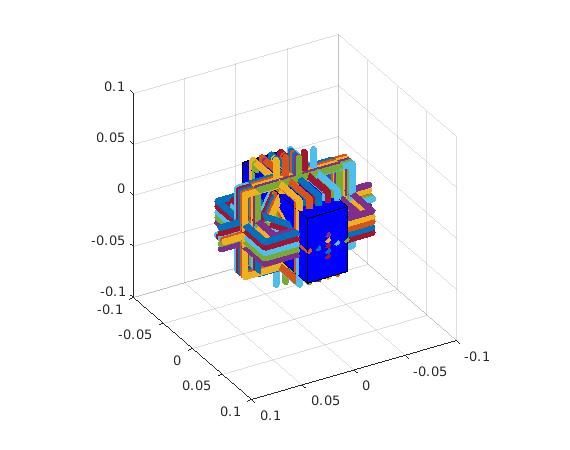}};
      
  \end{tikzpicture}
    \caption{Grasp family for Cylinder, Ring, Stick, Sphere, Semi-sphere, and Cuboid (from left to right, top to bottom).
    \label{fig:grasp_family}}
\end{figure*}

\subsection{Grasp Family}
The premise behind this work is that knowledge of shape informs grasping
opportunities. Simple shapes have known grasp options. This section
describes how estimating and registering recognized primitive shape models
in the scene permits recovery of one or more grasp families corresponding to
each recognized shape.  Each primitive shape class has a unique design for
its grasp families.  The configurations are based on the geometry of the
primitive shape and of parallel grippers, which are parametrically generated
based on shape symmetries. This idea can be extended to other types of
grippers. Example shape-based grasp configurations are depicted in Figure
\ref{fig:grasp_family} for reference.

The grasp family design is as follows: \\
1) {\em Cylinder:} There are two members corresponding to grasping from the
top or from the bottom, with the free parameter being rotation about the
cylinder axis. Another member corresponds to grasping from the side (the jaw
plane is perpendicular to the principal axis of the cylinder), with the free
parameters being the translation along and rotation about the principle
axis; \\
2) {\em Ring:} Similar to the cylinder, a ring can be grasped from the
top/bottom/side as well. However, the jaw plane is parallel to circular
plane of the ring when approaching from the side; \\
3) {\em Stick:} A restricted case of the cylinder that can only be grasped
from the side; \\
4) {\em Sphere:} We have two members for the sphere, which correspond to
grasping from the top or the side. The free parameter is the rotation about
the principal axis of the shape; \\
5) {\em Semi-sphere:} Unlike the sphere, the semi-sphere can only be grasped
from the top while the remaining parameters remain the same;
6) {\em Cuboid:} For the cuboid, the gripper can approach all six facets
vertically or horizontally, generating twelve members (6*2) in total. The
free parameter is translation along a single axis in the face plane.

Under ideal conditions (e.g., floating object, no occlusion, no collision,
and reachable), all the grasp configurations described above would be
possible. However, only a subset of grasps can be feasible in reality when
taking more factors into account. Importantly, when trying to grasp object
lying on a tabletop surface, the gripper may collide with the tabletop plane
if the grasp configuration is too close to the target object.  To deal with
this challenge, we introduce a specific post-processing step for small
objects, where the corresponding grasp family is generated away from the
target shape along the computed grasp's body z-axis. Additionally, three
options (small/medium/large) are employed to approximate the best opening
width for each grasp candidate. The association is based on the primitive
shape class and the grasp family class, while the width parameters are set
empirically according to the gripper geometry. 

Adopting the idea of grasp families for primitive shapes avoids issues
associated to incomplete annotations or sparse sampling of the grasp space.
Since the geometry of each shape is known, the predicted grasp family is
robust under different settings (modulo the weight distribution of the
object).
\newcommand{\quatR}{\mathfrak{q}}
\newcommand{\score}{s}
\newcommand{\scoreR}{s_{rot}}
\newcommand{\scoreT}{s_{trans}}
\newcommand{\costR}{C_{rot}}
\newcommand{\costT}{C_{trans}}
\newcommand{\rankMC}{\gamma_{MC}}
\newcommand{\rankTS}{\gamma_{TS}}
\newcommand{\rank}{\gamma_{g}}
\newcommand{\scoreO}{s_{occ}}
\newcommand{\scoreD}{s_{dim}}
\newcommand{\scoreC}{s_{col}}

%
%
%
\subsection{Grasp Prioritization and Selection} \label{ranking_algorithm}
%
%
The final step prior to execution is to select one grasp from the set of
candidate grasps. 
DexNet 2.0~\citep{mahler2017dex} was explored as a means to score the grasps,
but performance degraded as our camera setup was not of a top-down view. 
Instead, a simple geometric grasp prioritization scoring function was used
inspired by GPD grasp ranking~\citep{ten2017grasp}.
It considers the required pose of the hand 
$\grasp = \grasp^{\mathcal{W}}_{\mathcal{H}}$ 
relative to the world frame (which is located at the manipulator base),
the dimensions of hand versus the targeted shape, point cloud occupancy, and 
collision.
The prioritization scheme prefers dimension-compatible and
collision-free grasps to minimize translation, maximize the volume of
point cloud within the gripper closing region, and favor approaching
from above due to the manipulator elbow-up geometry.  The highest-ranked
grasp with a feasible grasping plan is the grasp selected.

%
The composite grasp prioritization score consists of contributions from
rotation, translation, occupancy, dimension-compatibility, and collision of
a candidate grasp pose. The translation score contribution depends on
the length of the translation element (i.e., the distance from the
world/base frame). Define the translation cost as $\costT(T) = \|T\|$,
where $T$ is the translation interpreted to be a vector in $\mathbb{R}^3$.
The rotational contribution regards the equivalent quaternion
$\quatR \in SO(3)$ as a vector in $\mathbb{R}^4$ and applies the following
positive, scalar binary operation to obtain the orientation grasp cost:
\begin{equation}
  \costR(\quatR) = \vec{w} \odot \vec \quatR, \quad \text{where}\ 
    \vec w = (0, \omega_1, \omega_2, \omega_3)^T\!\!\!,\ \omega_i > 0,
\end{equation}
with
$\vec a \odot \vec b \equiv |a^1 b^1| + |a^2 b^2| + |a^3 b^3| + |a^4 b^4|$.
This cost prioritizes vertical grasps by penalizing grasps that do not
point down.
Alternative weightings are possible depending on the given task, or the
robot configuration.

\begin{figure*}
  \centering
 \begin{tikzpicture}[inner sep=0pt, outer sep=0pt]
        
     \node[anchor=south west] (SM0) at (0in,0in)
       {\includegraphics[height=1.10in,clip=true,trim=1.4in 1.4in 1.4in 1.4in]{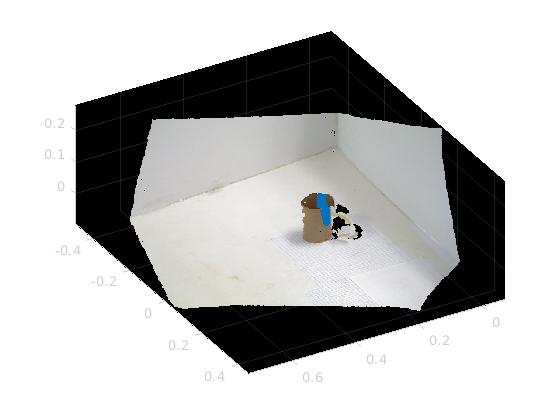}};
     \node[anchor=south east, xshift=-2pt] (SM1) at (SM0.south west)
       {\includegraphics[height=1.1in,clip=true,trim=1.4in 1.4in 1.4in 1.4in]{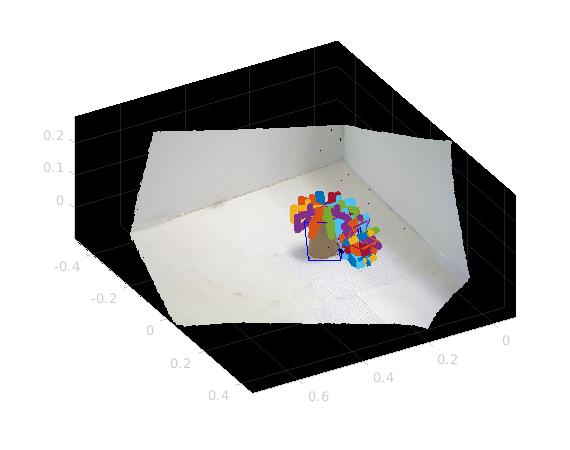}};
     \node[anchor=south east,xshift=-2pt] (SM2) at (SM1.south west)
       {\includegraphics[height=1.1in,clip=true,trim=0.5in 0.5in 0.5in 0.5in]{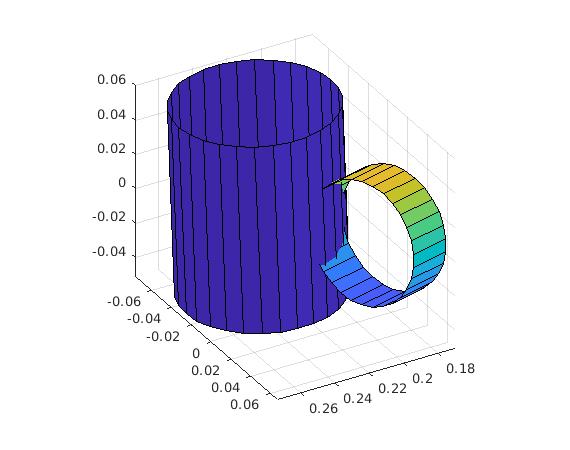}};
     \node[anchor=south east,xshift=-2pt] (SM3) at (SM2.south west)
       {\includegraphics[height=1.1in,clip=true,trim=1.4in 1.4in 1.4in 1.4in]{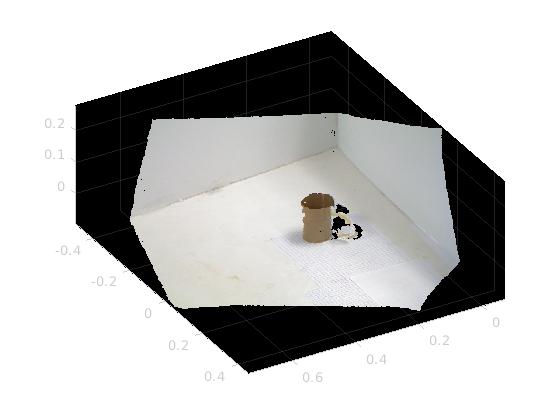}};
       
     \node[anchor=south west] (MM0) at (0in,-1.2in)
       {\includegraphics[height=1.10in,clip=true,trim=1.7in 1.4in 1.6in 1.4in]{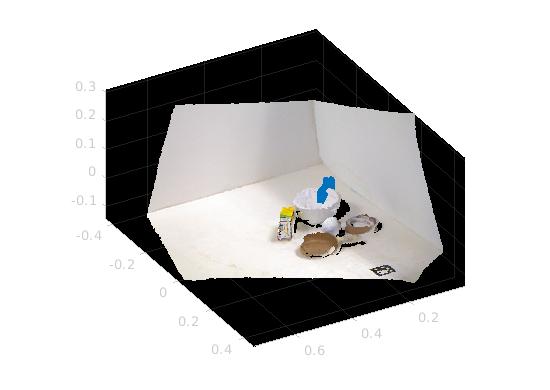}};
     \node[anchor=south east, xshift=-2pt] (MM1) at (MM0.south west)
       {\includegraphics[height=1.1in,clip=true,trim=1.4in 1.4in 1.4in 1.4in]{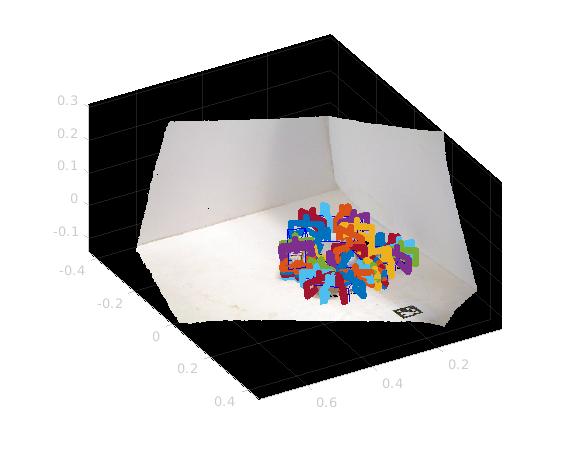}};
     \node[anchor=south east,xshift=-2pt] (MM2) at (MM1.south west)
       {\includegraphics[height=1.1in,clip=true,trim=0.5in 0.5in 0.5in 0.5in]{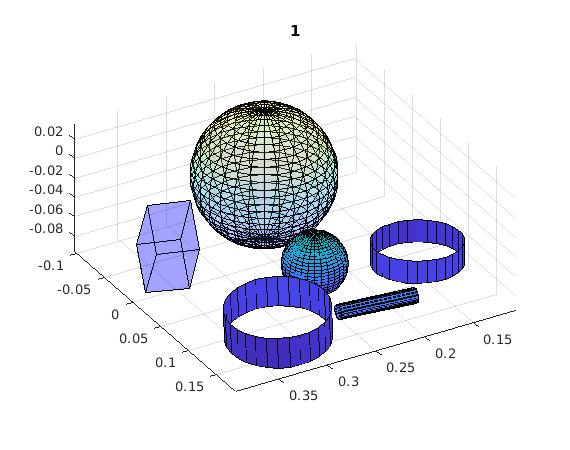}};
     \node[anchor=south east,xshift=-2pt] (MM3) at (MM2.south west)
       {\includegraphics[height=1.1in,clip=true,trim=1.3in 1.4in 1.0in 1.4in]{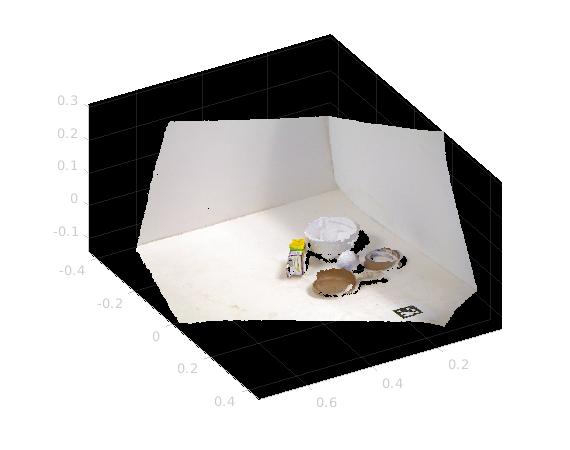}};
    \node[yshift=-2pt] at (-3.8in,-1.3in) {\small (a)};
    \node[yshift=-2pt] at (-2.4in,-1.3in) {\small (b)};
    \node[yshift=-2pt] at (-0.8in,-1.3in) {\small (c)};
    \node[yshift=-2pt] at (1.1in,-1.3in) {\small (d)};
   \end{tikzpicture}
   \caption{Sample visualization for candidate grasp generation, prioritization and selection on single object with multiple primitives (top) and multiple objects with multiple primitives (bottom). Shape decomposition, grasp candidates generation and grasp selection process are as follows.
   \textbf{(a)} {Scene point cloud is reconstructed from depth input for visualization.}
   \textbf{(b)} {Target object is decomposed into a combination of primitive shapes given segmented depth input.}
   \textbf{(c)} {Grasp candidates are generated according to the estimated shape categories, poses and dimensions with invalid grasp eliminated.}
   \textbf{(d)} {The highest scored grasp is selected for final execution.}
     \label{fig:graspdemo}}

\end{figure*}
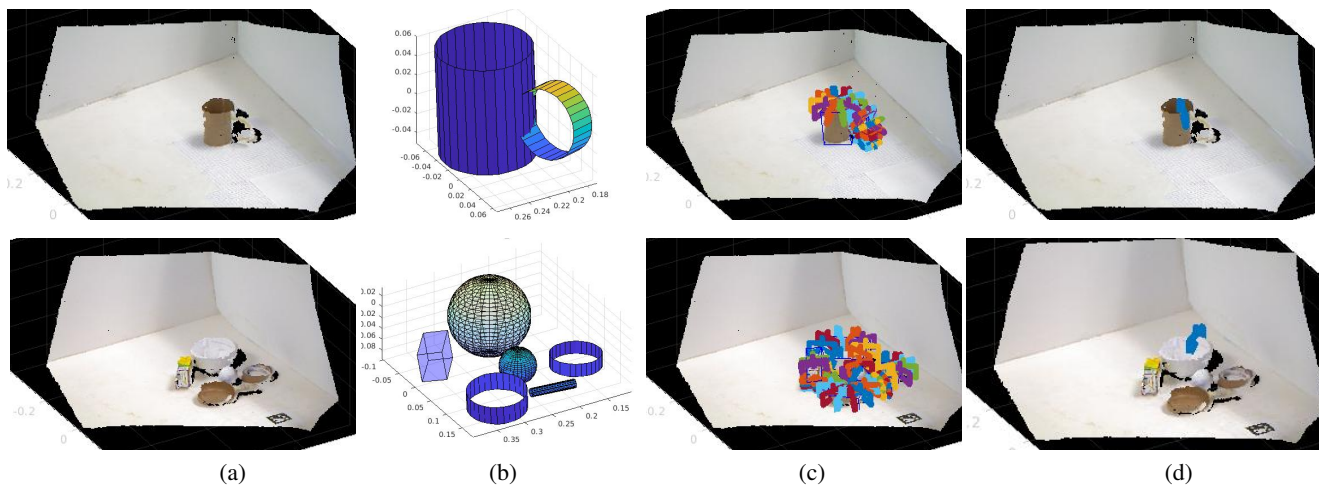

%
%
%
The rotation and translation costs are computed for all grasp candidates, then converted into scores by normalizing them over the range of obtained costs,
\begin{equation} \label{rt_score_eqt}
\begin{split}
  \scoreR(\costR) & = 1 -  \frac{\costR - \costR^{min}}
                                {\costR^{max} - \costR^{min}} \\
  \scoreT(\costT) & = 1 -  \frac{\costT - \costT^{min}}
                                {\costT^{max} - \costT^{min}} 
\end{split}
\end{equation}
where $\cdot^{min}$ and $\cdot^{max}$ superscripts denote the min and
max over all grasps, respectively.

%
We also take grasp stability into consideration. Let $B(\grasp)$
denotes the volume occupied by $\grasp$ when the gripper is fully open and
let $C(\grasp)$ denotes the volumetric region swept out by the gripper
when closing. The occupancy score $\scoreO$, which represents the volume of point cloud within the $C(\grasp)$, is calculated by
\begin{equation} \label{o_score_eqt}
    \scoreO = C(\grasp) \cap \primShape_{j}.
\end{equation}
Both of the dimension score $\scoreD$ and the collision score $\scoreC$
are binary-valued.  $\scoreD$ is set as zero if the estimated shape dimension
exceeds the gripper's maximal opening range while $\scoreC$ is set as zero if
    $B(\grasp) \cap  \primShape_{j} \neq \emptyset.$
The final composite score is:
\begin{equation} \label{score_ranking}
    \begin{aligned}
          \rank(\scoreR, \scoreT, \scoreO, \scoreD, \scoreC) &= \\
        (\lambda_{R} \scoreR + \lambda_{T} \scoreT + &\lambda_{O} \scoreO) \times  \scoreD \times \scoreC
    \end{aligned}
\end{equation}
for $\lambda_R,\lambda_T, \lambda_{O}>0$.

After ordering the grasps according to their grasp prioritization score,
the actual grasp applied is the first one to be feasible when a plan
is made from the current end-effector pose to the target grasp pose.
This final step in the grasp identification process is shown in Figure
\ref{fig:pipeline}(c).
The visualizations for intermediate steps are shown in Figure \ref{fig:graspdemo}.

%
%
\section{Grasping Experiment Setup and Properties}
\label{sec:exp_grasping}
To evaluate how well a PS-CNN based perception module translates to
practice, four grasping experiments were designed and executed.
The first experiment consists of the standard single-object grasping tests,
with objects ranging from an in-class grasping set of 3D-printed primitive
shapes to household object datasets replicating those from the literature 
\citep{Chu_Xu_Vela_2018,morrison2018closing}.  The objects cover known and
unknown types with diverse geometries and physical properties.
It also tests the claim that synthetic training data generated from domain
randomization over primitive shapes is rich enough for real-world use.
Grasping outcomes across all the objects are aggregated
for comparison to published methods that perform similar grasping tests.
The second experiment studies the effect of varied camera angles ranging
from 30 to 75 degrees, thereby testing robustness with respect to the
camera viewpoint. 
The third is a clutter-based robustness test: a multi-object clearing
task.
The fourth experiment is a task-oriented grasping task where a specific
primitive shape must be grasped, as would be done when performing a
specific task with the object. It aims to illustrate that the proposed
method could facilitate higher-level semantic grasping. 
This section describes the manipulator setup, the grasping test sets, and the
experimental evaluation criteria.

\begin{figure}[t]
  \centering
  {\includegraphics[width=0.95\columnwidth]{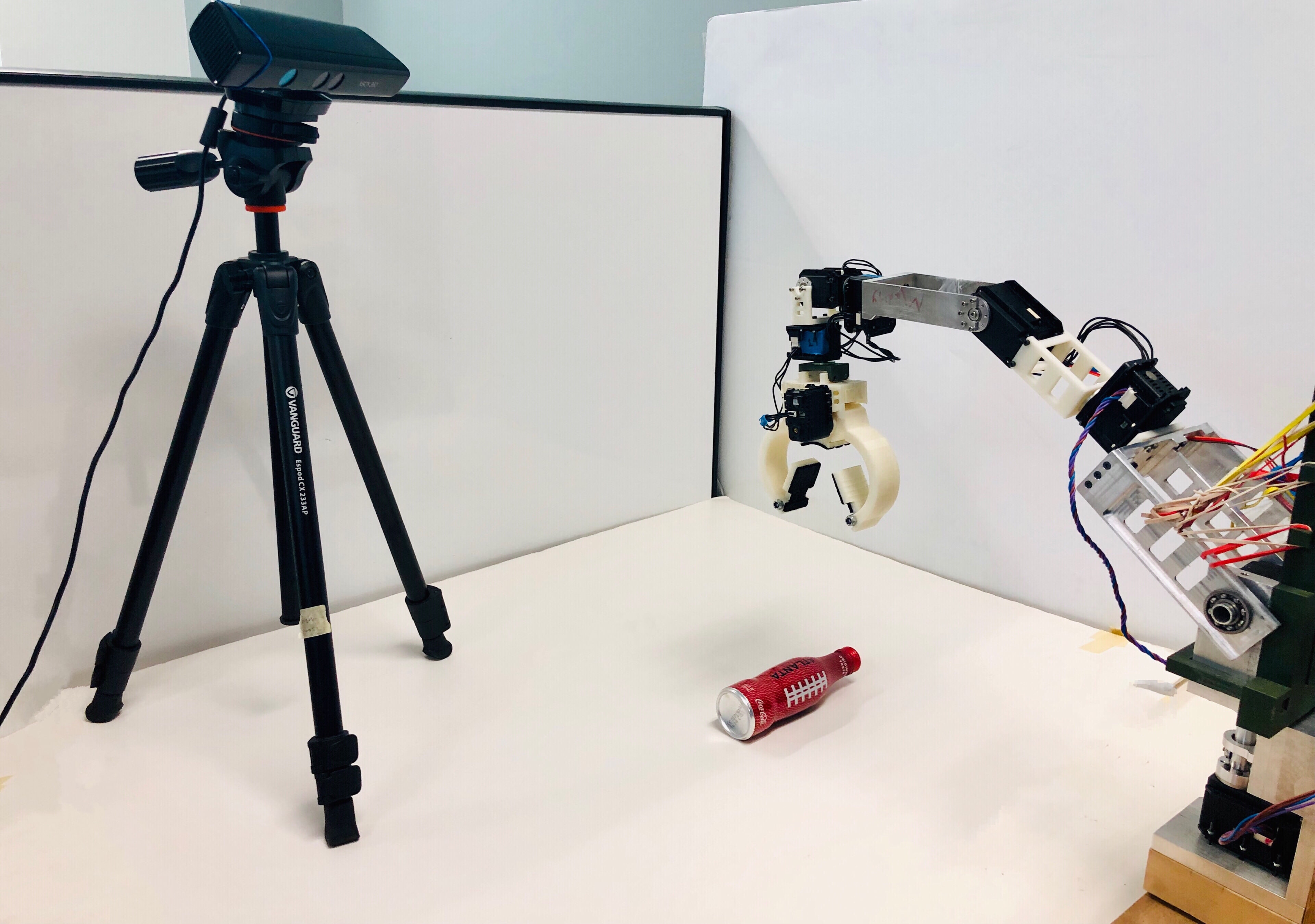}}
  \caption{Experimental setup in the real-world.\label{fig:exp_real_setting}}
\end{figure}

\begin{figure*}[t]
  \centering
  \begin{tikzpicture} [outer sep=0pt, inner sep=0pt]
  \scope[nodes={inner sep=0,outer sep=0}] 
    \node[anchor=north east] (e) at (\textwidth,0in)
    {\includegraphics[height=8.8cm,clip=true,trim=0.0in 1in 1in 2in]{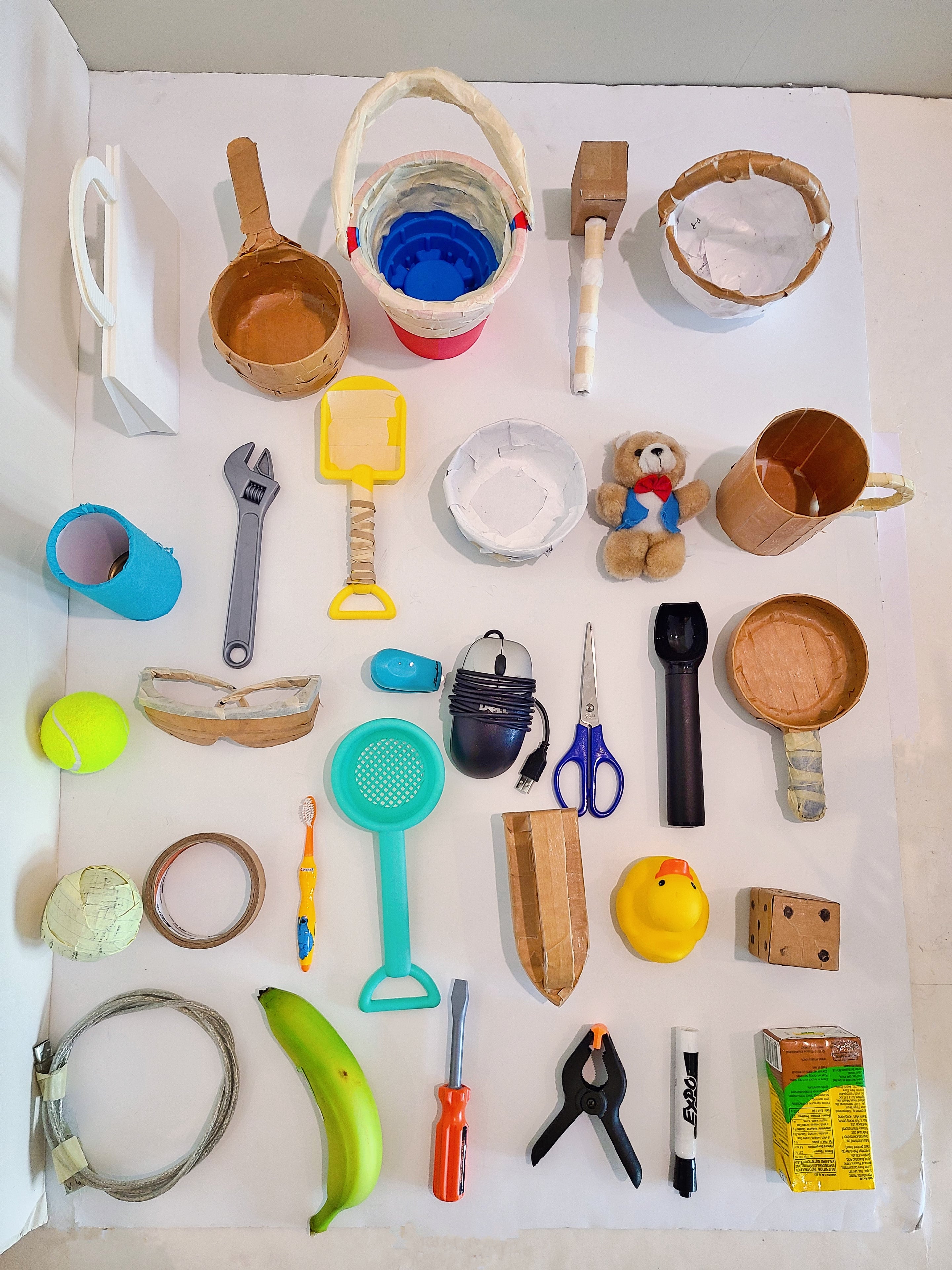}};

    \node[anchor=north east, xshift=-3pt] (f) at (e.north west)
    {\includegraphics[height=4.45cm,clip=true,trim=1in 0.0in 3.8in 2.5in]{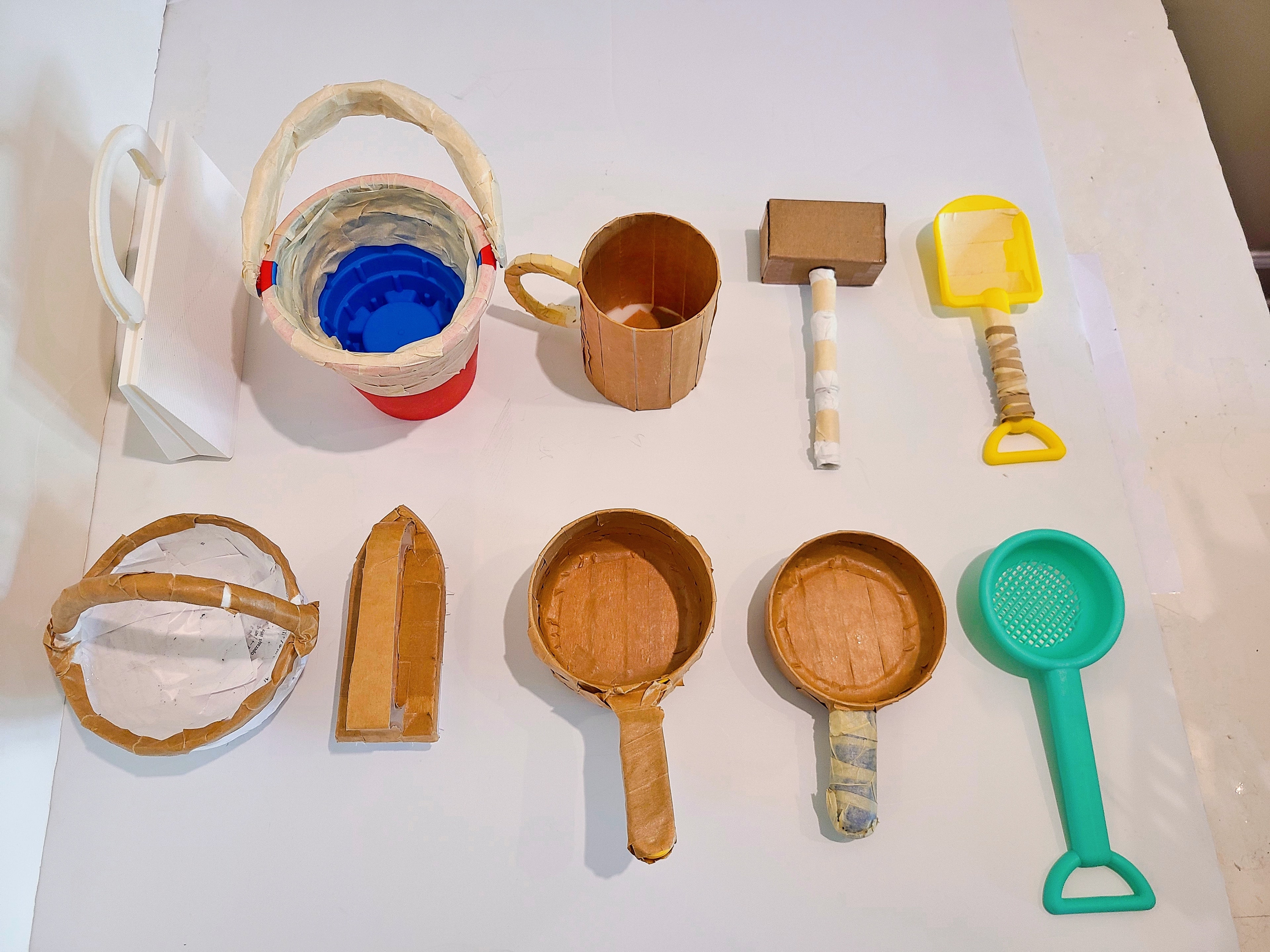}}; 

    \node[anchor=north east,xshift=-3pt] (a) at (f.north west)
    {\includegraphics[width=4cm,clip=true,trim=9in 10in 8in 8in]{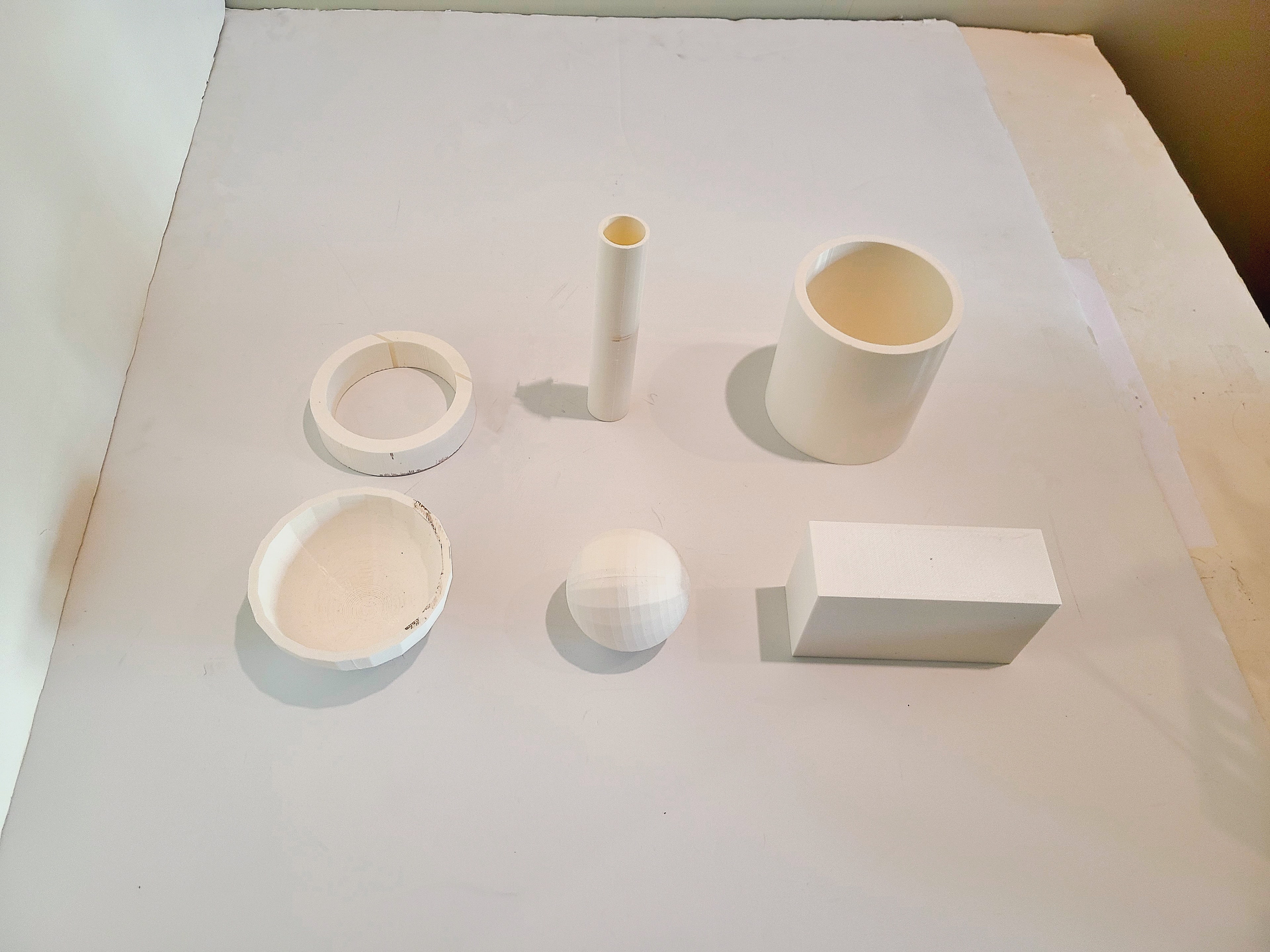}}; 

    \node[anchor=north west, yshift=-3pt] (b) at (a.south west)
    {\includegraphics[width=4cm,clip=true,trim=2in 11in 3in 19in]{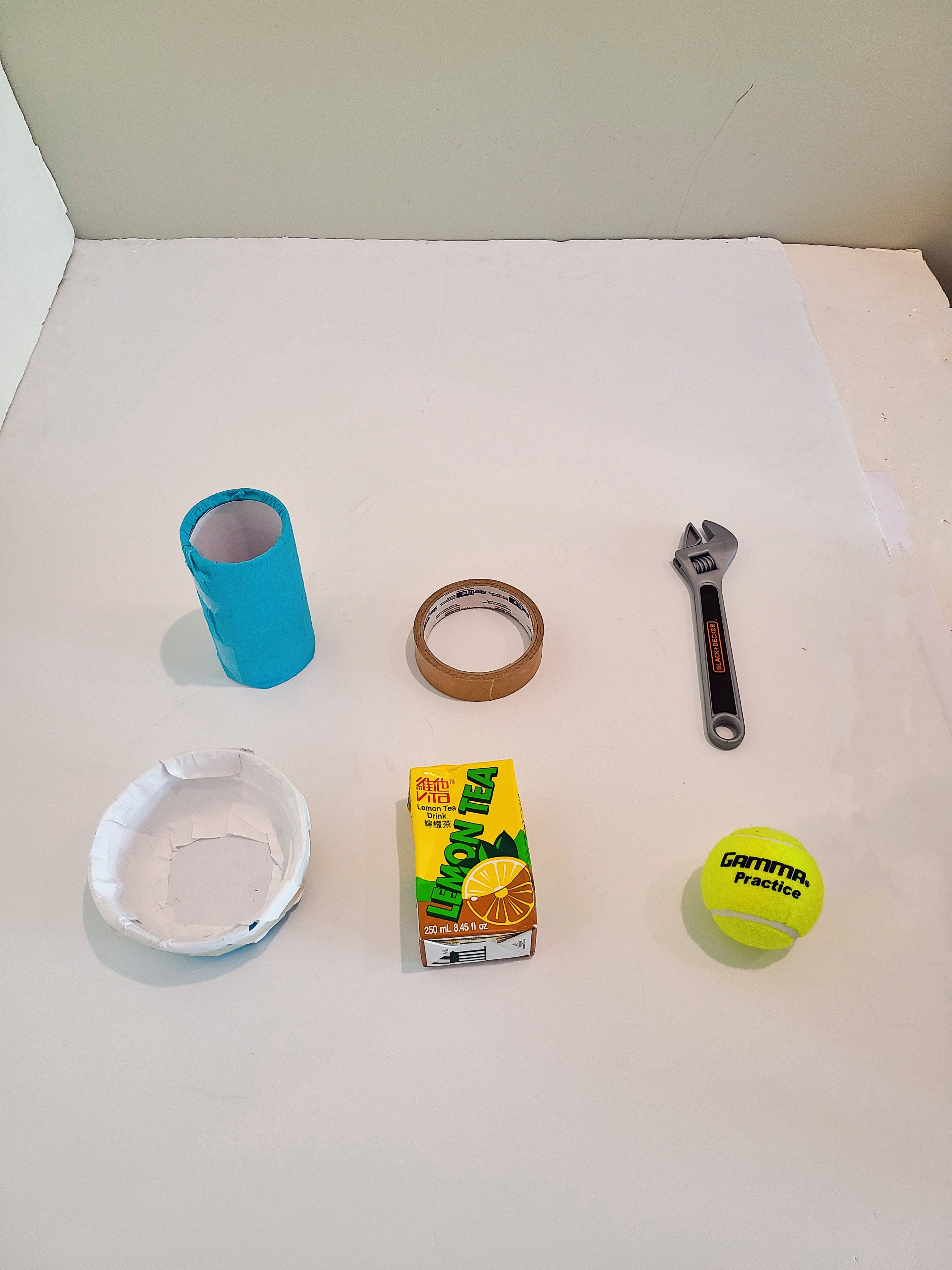}}; 
    
    \node[anchor=south east, xshift=-3pt] (d) at (e.south west)
    {\includegraphics[height=4.25cm,clip=true,trim=0in 2.0in 0in 16in]{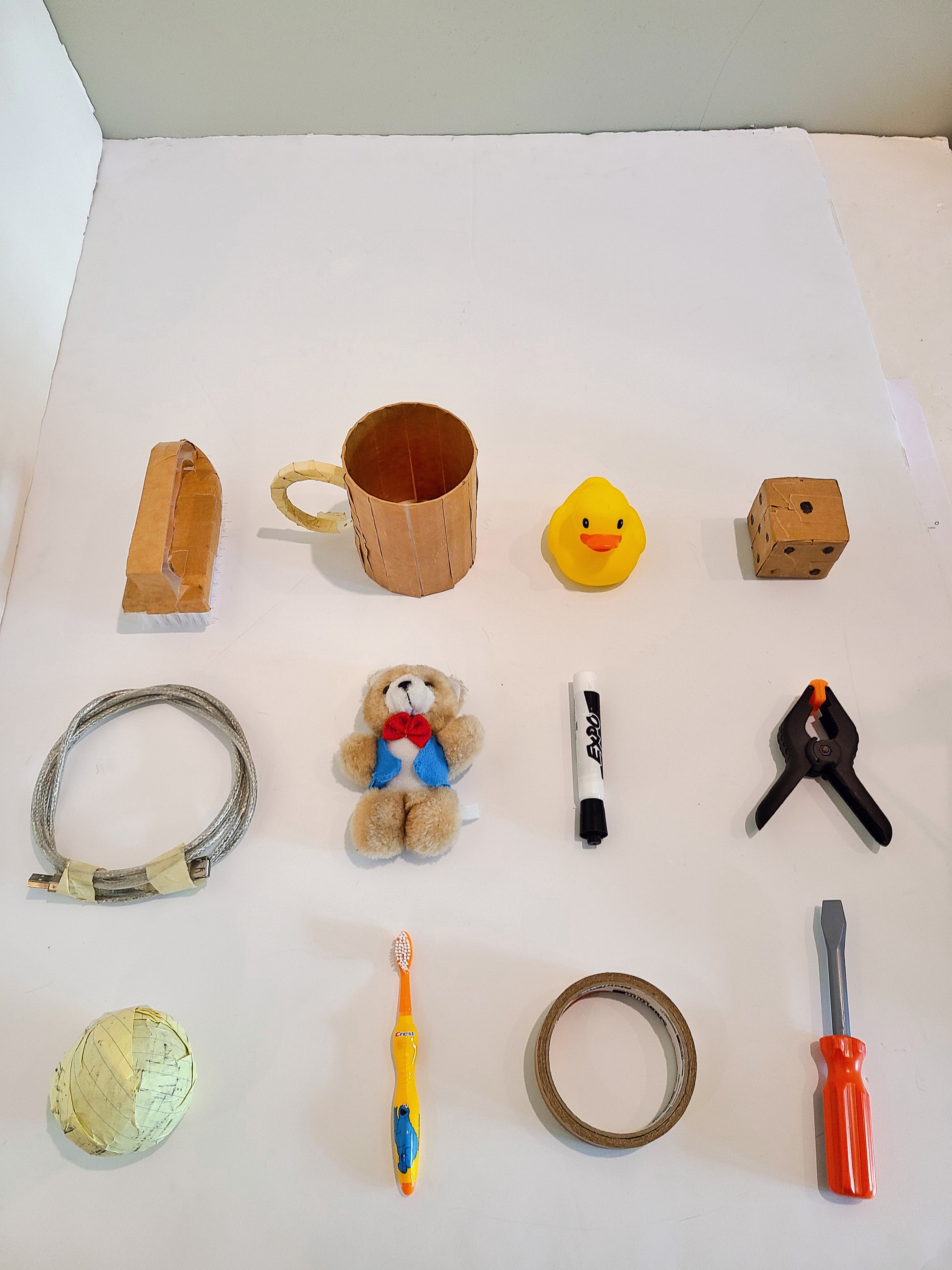}}; 

    \node[anchor=south east, xshift=-3pt] (c) at (d.south west)
    {{\includegraphics[height=3.5cm,clip=true,trim=3.0in 1.25in 4in 6in]{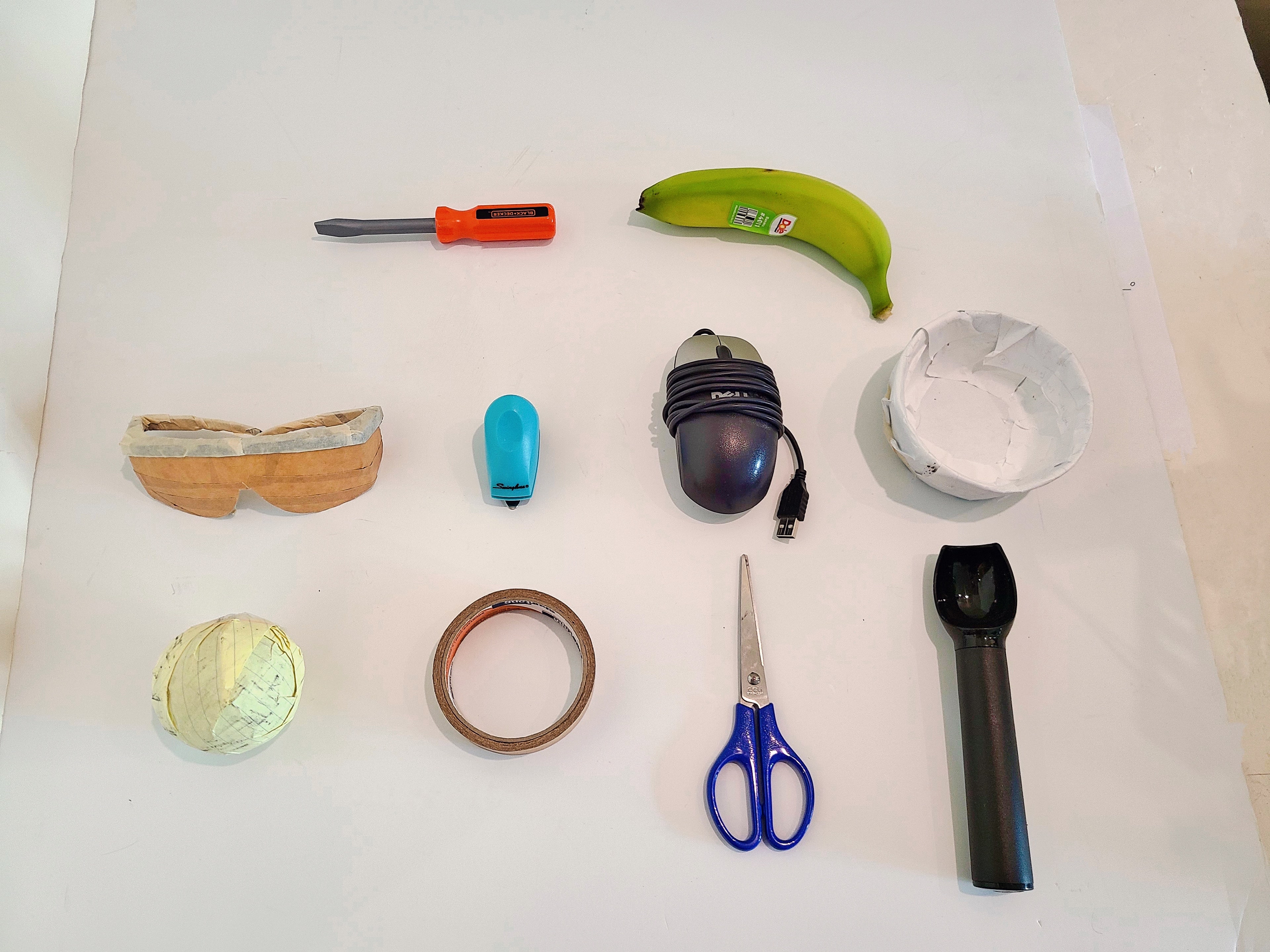}}}; 

  \endscope

  \foreach \n in {a,b,c,d,e,f} 
   {
    \node[anchor=north west,fill=white,draw=white,inner sep=1pt] at (\n.north west) {(\n)};
   }

  \end{tikzpicture}
  \caption{
    Object sets used for grasp experiments: (a) the known object set (3D-printed primitive shapes); (b) the varied angle set; (c) the unknown object set \citep{Chu_Xu_Vela_2018}; (d) the unknown object set \citep{morrison2018closing}; (e) the multi-object grasping set; (f) the task-oriented grasping set.\label{fig:obj_set}}.
\end{figure*}

\subsection{Robotic Arm Experimental Setup and Parameters}
\label{sec:exp_primitive_shapes_grasping}

The eye-to-hand robotic arm and RGB-D camera setup used for the experiments
is shown in Figure \ref{fig:exp_real_setting}.  The camera to manipulator
base frame is established based on an ArUco tag captured by the camera.
Both the described {\algName} pipeline and the implemented baseline methods
are tested on this setup.  
For the training dataset, each grasp set contributed by a grasp family is
discretized according to the dimensions of the gripper so that neighboring
grasps are not too similar.  For grasp ranking, the weights in
\eqref{score_ranking} are set as: 
$\lambda_R = \lambda_T = 0.5, \lambda_O = 0.0025$.
Open-loop execution is performed with the plan of the top grasp via MoveIt!~\citep{gorner2019moveit}.
The total running time, including segmentation, shape fitting, grasp
generation, collision checking, and path planning for a single object
averages 2.69s, of which segmentation takes 0.31s. 

\subsection{Grasping Object Sets}
Several object sets are used across the experiments. For comparison with
recently published methods, the same or similar objects were obtained and
used to define a named dataset class. 
The intent behind the experiments is
to test the value of primitive shapes, and not the challenges associated to
active depth sensor measurements, thus some objects with negative
reflectance properties were covered (i.e., with paper or tape).  
The text below describes these object sets and identifies the corresponding
visual in Figure \ref{fig:obj_set}.

\vspace*{0.25em}
\noindent {\bf Known Object Set.} 
A set of 3D-printed shapes designed to match the training shapes parameter
ranges is used for {\em known objects} testing and evaluation, Figure
\ref{fig:obj_set}(a).

\vspace*{0.25em}
\noindent
{\bf Varied Angle Grasping Set.} The varied angle grasping set, seen in
Fig. \ref{fig:obj_set}(b), comprises of 6 household objects which map to
the 6 different primitive shapes classes.

\vspace*{0.25em}
\noindent
{\bf Unknown Object Set.} 
We collected objects to recreate the object sets used in 
\cite{Chu_Xu_Vela_2018,morrison2018closing}, as shown in 
Figures \ref{fig:obj_set}(c,d). Chu's set \citep{Chu_Xu_Vela_2018} includes
ten commonly used objects collected from the Cornell Dataset, while
Morrison's set \citep{morrison2018closing} includes 12 household objects
selected from the ACRV Picking Benchmark (APB) \citep{leitner2017acrv} and
the YCB Object Set \citep{calli2017yale}.

\vspace*{0.25em}
\noindent
{\bf Multi-object Grasping Set.} To have a more diverse category of test
objects, we combined all the unique objects in the unknown object set, the
primitive shapes set, and the task-oriented grasping set together. 
The object set has 31 objects in total, as shown in \ref{fig:obj_set}(e).

\vspace*{0.25em}
\noindent
{\bf Task-oriented Grasping Set.} 
To better illustrate the idea of shape decomposition, we collect 10
different household objects which could be decomposed into more than 1
primitive shape, see \ref{fig:obj_set}(f).

%
%
\subsection{Grasping Evaluation Metrics}
For the robotic arm testing, only the final outcomes of the grasping tests
are scored.  
A run or attempt is considered to be a success if the target
object is grasped, lifted, and held for at least 10 seconds. The scoring
metric is the success rate (percentage). When possible, success rate
percentages are accompanied by 95\% confidence intervals where each
grasp attempt is considered to be a binary variable following a binomial
distribution.

%
%
\section{Grasping Experiments and Results}
\label{sec:exp_result}
%
%

This section covers the grasping experiments performed and discusses
their outcomes in relation to existing work.  
The experiments test the hypothesis that decomposing objects into
primitive shapes could contribute to a successful grasp recognition
pipeline.  They include primitive shape grasping, general object
grasping, viewpoint robustness, and clutter robustness experiments.
They highlight how effective shape is for establishing candidate grasps
and how shape may complement existing methods.  Lastly, there is a
task-oriented grasping experiment based on the premise that different
shape components perform different functions and should be grasped in
accordance to intended use. Task-oriented grasping creates a pathway for
a long-term investigation into shape as a mechanism to support
task-specialized grasping beyond pick-and-place testing. Each
experiment includes experimental setup details and and testing criteria
for context.

\subsection{Static Object Grasping}
\label{sec:res_static}

%
%
\begin{table*}[t]
  \centering
  \caption{Physical Grasping with 95\%
confidence intervals on Primitive Shapes (Known) \label{tab:exp_known} }
  \begin{tabular}{|l|c|c|c|c|c|}
    \hline
    & \cite{jain2016grasp} & \cite{Chu_Xu_Vela_2018} &  {\PS} v1 & {\PS} v2\\ 
    \hline\hline
    Cuboid
         & 8/10    & 5/10   & 8/10 & 10/10 \\ \hline
    Cylinder
        & 9/10    & 8/10 & 10/10  & 10/10   \\ \hline
    Semi-sphere
       & 5/10    & 7/10  & 9/10 & 10/10    \\ \hline
    Stick
       & 7/10    & 6/10  & 10/10 & 10/10   \\ \hline
    Ring
         & 6/10    & 5/10 & 9/10  & 9/10  \\ \hline
    Sphere
        & 8/10    & 8/10  & 9/10 & 10/10   \\ \hline \hline

  
   Average (\%) & 71.7 $\pm$ 11.4 & 65.0 $\pm$ 12.1 & 91.7 $\pm$ 7.0 & 98.3 $\pm$ 3.3  \\   \hline
      
  \end{tabular}
\end{table*}

\begin{table*}[t]
\begin{minipage}[t]{0.55\linewidth}
\centering
  \caption{Physical Grasping for Chu's Household Set, with 95\%
  confidence intervals \label{tab:exp_chu}}
  \begin{tabular}{ | l | c | c | c |}
    \hline
    {\bf object}  & \cite{Chu_Xu_Vela_2018} &  GKNet  &  {\PS} v2 \\ \hline \hline
Banana      & 7/10  & 10/10 & 10/10 \\ \hline
Glasses     & 8/10  & 10/10 & 8/10  \\ \hline
Ball        & 9/10  & 10/10 & 10/10 \\ \hline
Tape        & 10/10 & 9/10  & 10/10 \\ \hline
Screwdriver & 7/10  & 10/10 & 10/10 \\ \hline
Stapler     & 10/10 & 9/10  & 9/10  \\ \hline
Spoon       & 9/10  & 10/10 & 10/10 \\ \hline
Bowl        & 10/10 & 10/10 & 10/10 \\ \hline
Scissors    & 8/10  & 9/10  & 8/10  \\ \hline
Mouse       & 8/10  & 9/10  & 8/10  \\ \hline \hline
Average (\%)    & 86.0 $\pm$ 6.8 & 96.0 $\pm$ 3.8  & 93.0 $\pm$ 5.0  \\ \hline
  \end{tabular}
    \end{minipage}
%
%
  \begin{minipage}[t]{0.425\linewidth}
\centering
  \caption{Physical Grasping for Morrison's Household Set, with 95\%
  confidence intervals \label{tab:exp_morrison}}
  \setlength{\tabcolsep}{4.5pt}
  \begin{tabular}{ | l | c | l | c |}
    \hline
    {\bf object} & \bf{Acc.} & {\bf object} & \bf{Acc.}            \\ \hline
    Mug          &      10/10     &  Ball        & 10/10                    \\ \hline 
    Brush        &      10/10     &  Toothbrush  &  9/10                    \\ \hline 
    Bear toy     &     8/10     &  Dice        & 10/10                    \\ \hline 
    Tape         &      10/10     &  Duck toy    & 9/10                    \\ \hline 
    Marker       &     9/10     &  Clamp     &  7/10                    \\ \hline 
    Screwdriver  &     10/10     &  Cable       & 10/10                    \\ \hline\hline
    
        \multicolumn{2}{|l|}{{\PS} v2}  &   \multicolumn{2}{c|}{93.3
        $\pm$ 4.5 (\%)}     \\ \hline 
    \multicolumn{2}{|l|}{\cite{morrison2018closing} } &
    \multicolumn{2}{c|}{92.0 $\pm$ 4.9 (\%)}  
    \\ \hline 
    \multicolumn{2}{|l|}{GKNet} & \multicolumn{2}{c|}{95.0 $\pm$
    3.9 (\%)}  \\ \hline

    
     
  \end{tabular}
  \end{minipage}
\end{table*}

The static object grasping experiment evaluates the proposed primitive
shapes grasping recognition pipeline under an ideal setup. 
This section explores the following questions: 
1) Does {\PS} bridge the simulation-reality gap? 
2) Is the primitive shapes concept robust to objects of different type
or form? 
3) What value does the primitive shapes idea have relative to other
approaches to grasping?

The use of depth instead of RGB image inputs should minimize the
sim2real gap when dense depth imagery is available as an input source
and mostly captures the objects of interest. This expectation has been
observed to hold for other depth-only methods, such as
\cite{viereck2017learning}.
Testing starts with the easiest case of {\em known objects}, i.e., 
3D-printed primitive shapes. It continues with grasping experiments for
the Chu and Morrison datasets ~\citep{Chu_Xu_Vela_2018,morrison2018closing}, 
containing {\em unknown objects}.  The two sets cover representative
household objects of different sizes and shapes. Lastly, aggregate
results permit rough comparison with published works.


%
%

\subsubsection{Known Objects (Primitive Shapes).}
The outcomes for the {\em known objects} grasping test are
in Table \ref{tab:exp_known}. 
The baseline implementations include \cite{jain2016grasp},
a primitive-shape-based grasping system for household objects fitting
spherical, cylindrical, and box-like shape primitives; and
\cite{Chu_Xu_Vela_2018}, a typical 2D grasp representation approach.
Also included is the previous primitive shape effort {\PS} v1
\citep{lin2020using}. 

The \cite{jain2016grasp} baseline performed better for shapes easily
approximated by spherical, cylindrical, and box-like primitives
($83.3\%$), but performed worse otherwise ($60.0\%$).  When considering
the 95\% confidence interval of {\PS} v2, all of the baseline methods
were below the interval's lower bound. The improvements to {\PS} v2 led
to a 6.6\% performance boost over {\PS} v1. The primary factor behind
the performance difference is the upgraded post-segmentation
shape-fitting approach.

%
%
\subsubsection{Unkown Objects.}
The outcomes for the {\em unknown objects} grasp test are in 
Tables \ref{tab:exp_chu} and \ref{tab:exp_morrison}.  Included are
published results of the analogous trials for the corresponding baseline
methods~\citep{Chu_Xu_Vela_2018,morrison2018closing,xu2021gknet}.
For compatibility with the {\PS} processing pipeline, the
\cite{Chu_Xu_Vela_2018} results are from the top-1 grasp selection
criteria. 

The first observation is that {\PS} v2 has a 5.3\% and a 3.3\% drop in
success rate relative to the {\em known objects} performance outcomes,
for the Chu and Morrison datasets, respectively.  Combining the 19 unique results from
Tables~\ref{tab:exp_chu} and~\ref{tab:exp_morrison}, {\PS} v2 has a
93.2\% success rate, which is a 5.1\% drop in performance.
The drop does indicate some sensitivity to unknown objects, however
the relative performance of {\PS} v2 to the baseline methods indicates
that the sensitivity is not large. {\PS} v2 lies within the 95\%
confidence interval of the best performing method, GKNet, across the two
datasets. Furthermore, it outperforms the source implementations from
each of the datasets. The {\PS} v2 confidence interval for the Chu
dataset excludes \cite{Chu_Xu_Vela_2018}, which means that it is a more
effective method. For the Morrison dataset, the confidence interval
includes the source implementation but {\PS} v2 has higher performance
(by 1.3\%).

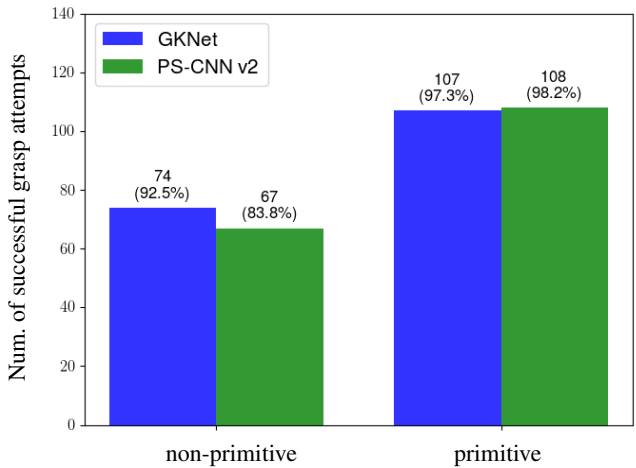
\begin{figure}[t]
  \centering
   \begin{tikzpicture}[inner sep = 0pt, outer sep = 0pt]
    \node[anchor=south west] at (0in,0in)
      {{\includegraphics[height=6cm]{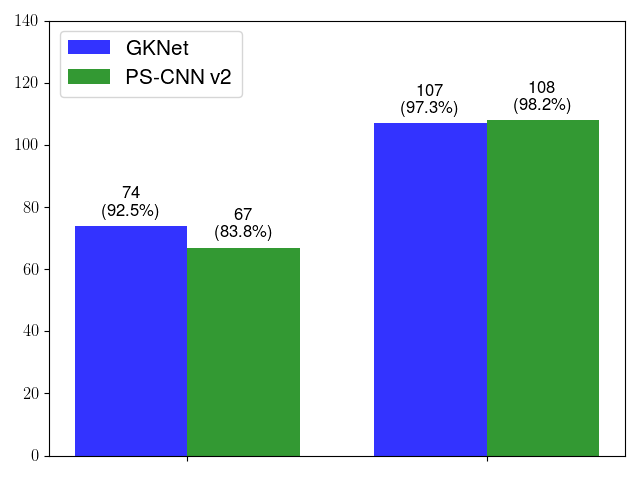}}};
 \node[rotate=90] at (-0.1in,1.2in) {\small Num. of successful grasp attempts};
 \node[anchor=south] at (1in,-0.1in)
      {\small non-primitive};
 \node[anchor=south] at (2.38in,-0.1in)
      {\small primitive};
\end{tikzpicture}    \caption{ Comparison between GKNet~\citep{xu2021gknet} and {\PS} v2 on the unknown object sets, where the 19 unique objects are divided into non-primitive shape-like ones and primitive shape like ones (The number in the bracket represents the corresponding success rate).\label{fig:exp_gknet_comparison}}

\end{figure}

GKNet \citep{xu2021gknet} achieves the best performance on the two
datasets individually, and when considering the 19 unique outcomes
(95.3\%). 
First, the performance difference is 3.2\% (95.3\% vs.~92.1\%), which is
lower than the baseline methods outside of GKNet, indicating
that shape information can be effective at informing grasp options.
GKNet was trained with a variety of objects, some of which look like the
test objects, thus it is expected to have some advantage. These objects
are unknown to the {\PS} but some are known to the GKNet.
Secondly, the results are better interpreted by dividing the objects into
two groups: non-primitive shape and primitive shape objects.  Primitive
shape objects are those that mostly fit one of the primitive shapes
known to {\PS}.  As shown in Figure~\ref{fig:exp_gknet_comparison},
GKNet outperforms {\PS} v2 on the first group while {\PS} v2
outperforms GKNet on the second group (e.g., they lack complex geometry). 
There exists the opportunity to combine both primitive shape and
keypoint strategies to improve general purpose grasping. Evidence for
this can be seen through a composite GKNet+{\PS} v2 success rate as
obtained from selecting the maximum success rate between GKNet and {\PS}
v2.  The composite success rate is 96.84\%, which is 1.58\% higher than
GKNet and 4.74\% higher than {\PS} v2. Shape can play a complementary
role to existing methods.





\begin{figure*}[t]
  \centering
  \begin{minipage}[t]{0.99\columnwidth}
  \centering

  \begin{tikzpicture} [outer sep=0pt, inner sep=0pt]
  \scope[nodes={inner sep=0,outer sep=0}] 
  \node[anchor=north west] (a) at (0in,0in) 
    {\includegraphics[height=3.6cm,clip=true,trim=2.5in 3.3in 4.3in 1.5in]{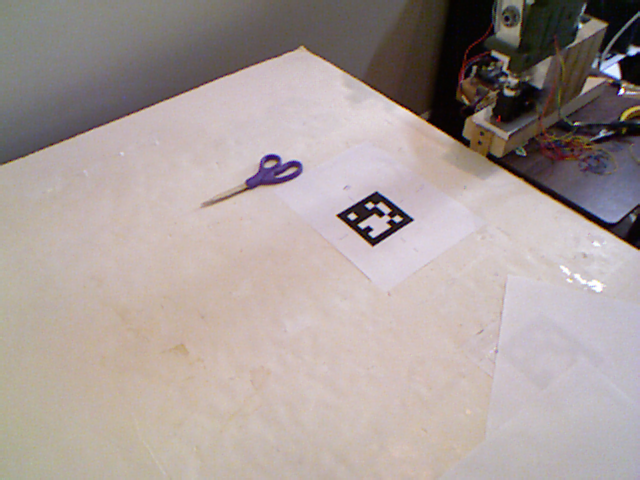}}; 
  \node[anchor=north west, xshift=0.1cm] (b) at (a.north east)
    {{\includegraphics[height=3.6cm,clip=true,trim=2.5in 3.3in 4.3in 1.5in]{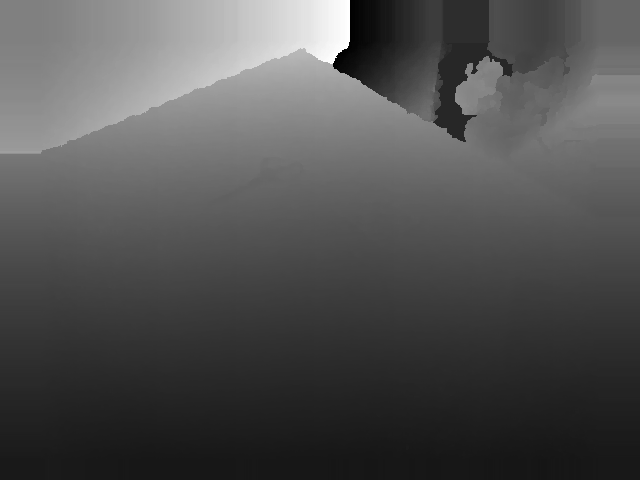}}}; 
  \endscope

  \foreach \n in {a,b} 
   {
    \node[anchor=north west,fill=white,draw=white,inner sep=1pt] at (\n.north west) {(\n)};
   }
  \end{tikzpicture}
  \caption{Scissor failure case. 
    (a) Color image. 
    (b) Depth input after post-processing. Object geometry has low
    signal-to-background differences.
    \label{fig:exp_failure_scissor}}

  \end{minipage}
  \begin{minipage}[t]{0.99\columnwidth}
  \centering

  \begin{tikzpicture} [outer sep=0pt, inner sep=0pt]
  \scope[nodes={inner sep=0,outer sep=0}] 
  \node[anchor=north west] (a) at (0in,0in) 
    {\includegraphics[height=3.6cm,clip=true,trim=3.15in 3in 3.5in 1.7in]{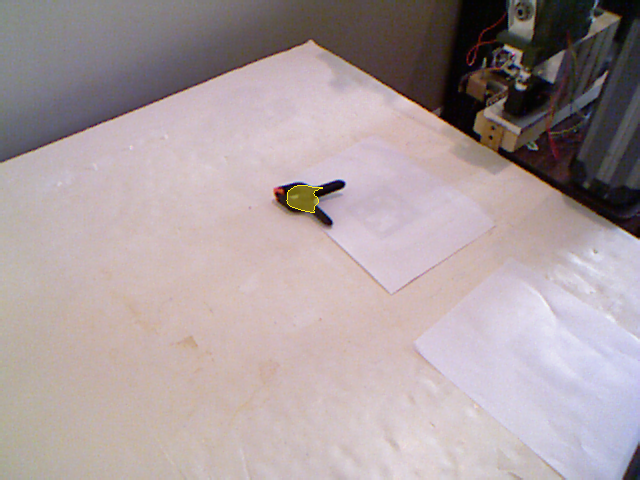}}; 
  \node[anchor=north west, xshift=0.1cm] (b) at (a.north east)
    {{\includegraphics[height=3.6cm,clip=true,trim=3.3in 2.5in 4.1in 2.8in]{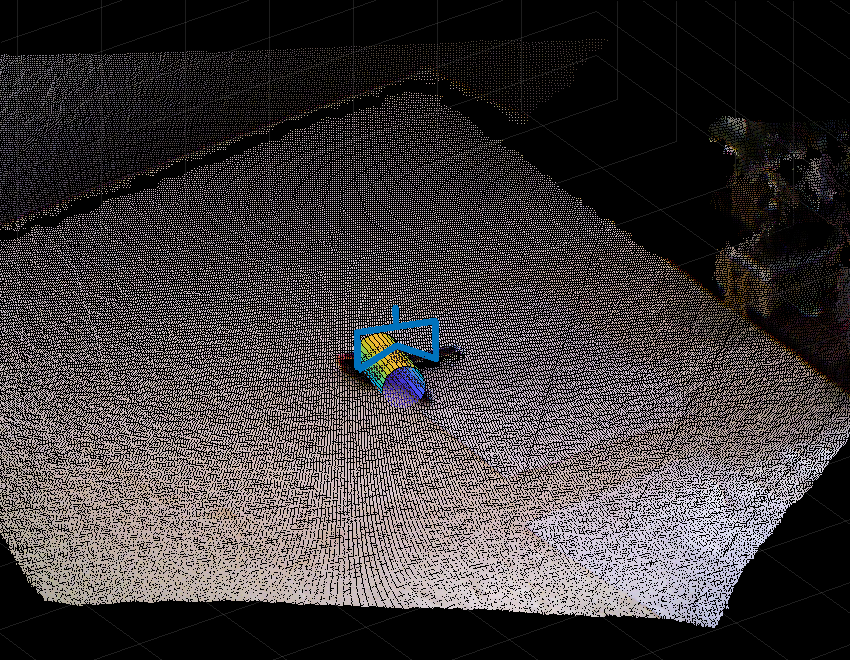}}}; 
  \endscope

  \foreach \n in {a,b} 
   {
    \node[anchor=north west,fill=white,draw=white,inner sep=1pt] at (\n.north west) {(\n)};
   }
  \end{tikzpicture}
  \caption{Clamp failure case. (a) Segmentation result. 
  (b) Shape fitting result with rank-1 grasp. 
  Though the grasp seems feasible, the wide end-effector affects
  grasping success.
  \label{fig:exp_failure_clamp}}

  \end{minipage}
\end{figure*}

\begin{figure}[t]
  \centering
  \begin{tikzpicture} [outer sep=0pt, inner sep=0pt]
  \scope[nodes={inner sep=0,outer sep=0}] 
  \node[anchor=north west] (a) at (0in,0in) 
    {\includegraphics[height=3.9cm,clip=true,trim=4in 5in 2in 15in]{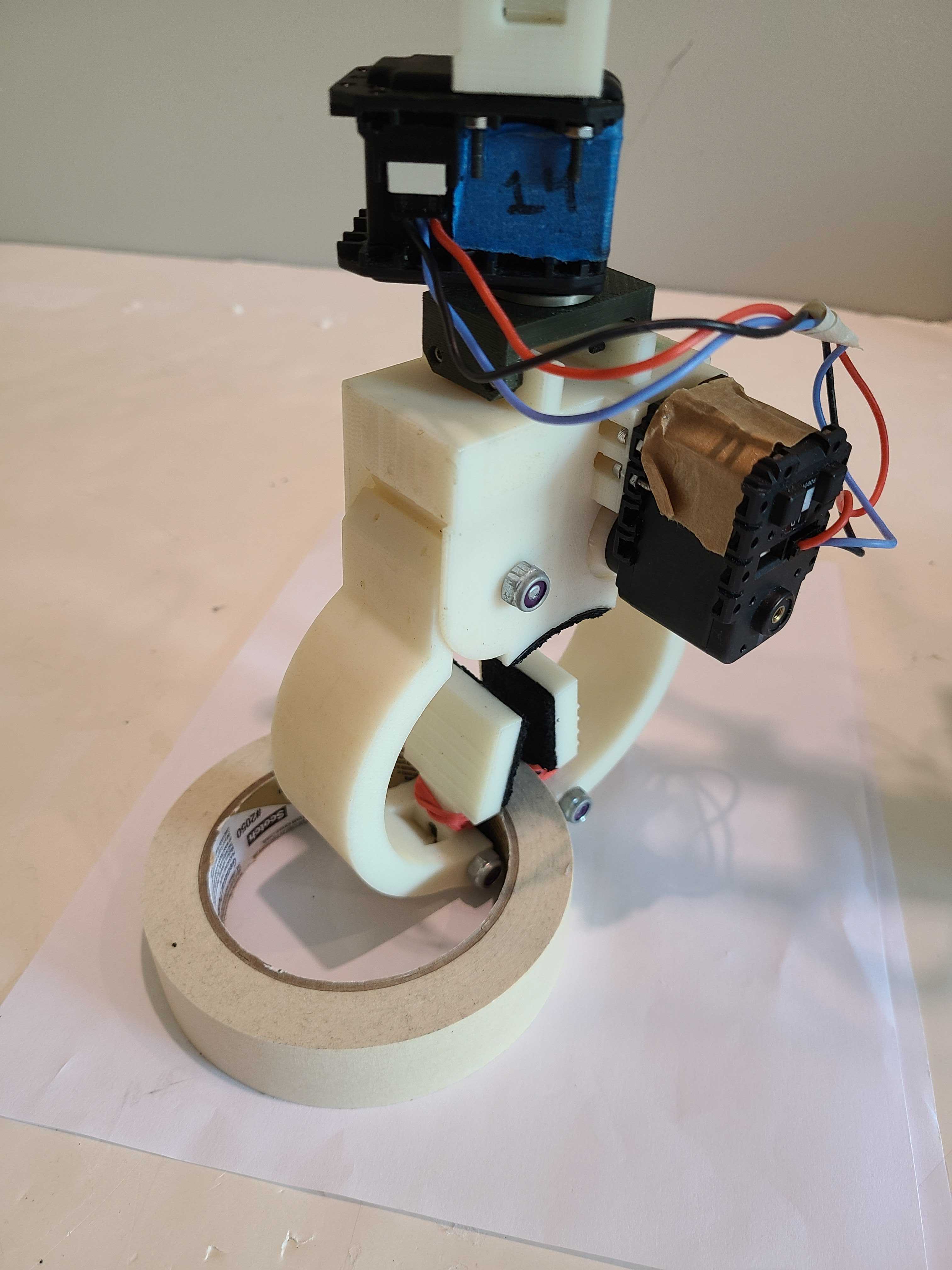}}; 
  \node[anchor=north west, xshift=0.1cm] (b) at (a.north east)
    {{\includegraphics[height=3.9cm,clip=true,trim=0in 0in 0in 0in]{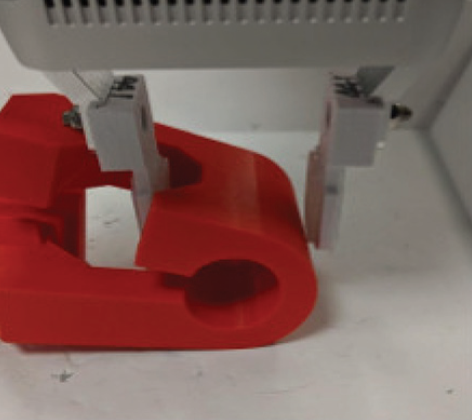}}}; 
  \endscope

  \foreach \n in {a,b} 
   {
    \node[anchor=north west,fill=white,draw=white,inner sep=1pt] at (\n.north west) {(\n)};
   }

  \end{tikzpicture}
  \caption{Comparison on the gripper between (a) ours and (b) 
  the Franka Emika robot used in \cite{morrison2018closing}. 
  }
  \label{fig:exp_gripper}
\end{figure}

To understand some of the lower performing outcomes, this paragraph
reviews a couple of prominent failure cases. The scissors in the Chu
dataset should be grasped well since the cutting parts take the geometry
of a stick when closed. Unfortunately, in some cases the
flatness of the object exposed depth resolution issues.  Figure
\ref{fig:exp_failure_scissor} shows the color and depth images for a
pair of scissors.  The depth variation in the image is not as salient as
the color variation, for the scissors relative to the background.
Objects in the monocular depth image may be missing, have missing parts,
or ambiguous geometry after denoising.  Another example includes the
clamp from the Morrison dataset, depicted in Figure \ref{fig:exp_failure_clamp}.
It has an irregular shape, which is challenging to be abstracted or
decomposed, though the handles are stick-like.  
The algorithm does abstract part of the clamp as a stick, but does so
for the wrong region. In spite of that, the grasp prediction seems
visually reasonable, but the manipulator failed to grasp the object. 
This is attributed to the gripper geometry, which is wider than the
typical parallel-plate gripper and more sensitive to imprecise gripper
positioning.  Compare the gripper used with the Franka Emika robot
\citep{morrison2018closing}, c.f.~Figure \ref{fig:exp_gripper}. Other
reduced grasping outcomes arise from shapes that don't cleanly fit to
the known primitive shape families, but do approximately fit to one
primitive shape class. The best scoring grasp may not be best for the
actual geometry, which opens the opportunity for grasp quality scoring
methods to provide improved scoring values over analytical models based
on idealized geometries.


\subsubsection{Tabletop Grasping of Individual Objects.}
To better place the value of primitive shapes relative to successful
grasping strategies, the success rate statistics and experimental
parameters were collected from published approaches employing only a
single gripper (not suction-based).

{\bf Source Dataset and Setup:}
During the literature review, we noted the testing scenarios for the
cited works and marked which publications provided physical robot
grasping performance outcomes for individual objects on a tabletop.
If single and multi-view results are provided, then the single view
performance outcomes were taken.
The references in Table
\ref{tab:exp_comparison} all performed such an experiment, except for
\cite{viereck2017learning} which used a single-camera visual servoing
approach.  Factors that cannot be accounted for include the robot
manipulator and gripper, the software pipeline, and the actual test
objects.  Even though the test objects may differ, the general household
object test sets tend to be similar.  This is because recent research
efforts obtain their sets from existing work \citep{calli2017yale} or
attempt to recreate the object set
\citep{Chu_Xu_Vela_2018,morrison2018closing}.  The biggest exception
being that~\cite{mahler2017dex, morrison2018closing},
and~\cite{xu2021gknet} also take adversarial objects into consideration.
In regard to the evaluation metric, most adopt a success policy similar
to the one here, that is to successfully grasp and lift the object in
the air. DexNet 2.0 \citep{mahler2017dex} has stricter criteria which
only counts the grasps that are lifted, transported, and held the object
after shaking the hand.  Despite the existing differences, a rough
comparison can still provide some context for the relative performance
of {\PS} v2 on household objects. 

Presented in Table \ref{tab:exp_comparison}, the selected works are
divided into two groups based on the grasp representation or strategy.
One group (2D Grasp) typically employs a 2D grasp representation,
processes the data in image-space, and assumes a top-down grasping
technique. The second group (3D Grasp) employs a 3D grasp
representation, processes the data using point clouds, and has less
constraints on the grasp orientation or approach direction. {\PS} 
belongs in the 3D Grasp group.  Three key properties of the
experiments are reported: the input type, the number of unique objects
tested, and the total number of trials across the objects tested.

{\bf Outcome Analysis:} Adopting the confidence interval of {\PS} v2 as
the selection criteria, we organize the top performing results sequentially:
~\cite{viereck2017learning, xu2021gknet}, {\PS} v1,
~\cite{wang2021double}, and~\cite{asif2017rgb}. They achieve a success rate higher than
the lower bound of the {\PS} v2 confidence interval (90.9\%) while the
remaining approaches lie below the interval and are presumed to achieve
lower performance granted that their grasping experiments were
comparable. {\PS} v2 provides updated and improved success rates versus
{\PS} v1 based on an improved pipeline and evaluation on a larger 
test objects set.

\cite{viereck2017learning} achieved the highest success rate among all.
There are two possible reasons. The first being the limited quantity of
test objects and trials, and the second being the advantage of the
eye-in-hand visual servoing system design. The approach permits multiple
estimates from different views over time, which may offset any grasp
recognition errors associated to the grasp predictions of the initial image.

\begin{table}[t]
        \centering
          \caption{Grasping Comparison from Published Works
    \label{tab:exp_comparison}}
        \begin{threeparttable}
          \setlength\tabcolsep{4pt}
          \begin{tabular}{|l|c|c|c|c|}
          \hline
           \multicolumn{1}{|l|}{\multirow{2}{*}{\bf{Approach}}}  &
            \multicolumn{3}{c|}{\multirow{2}{*}{\bf{Settings}}} &
            \multicolumn{1}{c|}{\bf{Success}} \\
            \multicolumn{1}{|c|}{}                   &
            \multicolumn{3}{c|}{}                   &
            \multicolumn{1}{c|}{\bf{Rate (\%)}} \\ \hline

          &  \multicolumn{1}{c|}{Inp.} &\multicolumn{1}{c|}{Obj.} &
            \multicolumn{1}{c|}{Trials} &
            \\  \hline
                       {\bf 2D Grasp} &  &  &     &
            \\  \hline
          \cite{jiang2011efficient}   & M   & 12  & - & 87.9
          \\ \hline
          \cite{lenz2015deep} & M  & 30  & 100 & 84.0/89.0\tnote{*}\rule{0pt}{2.2ex}
          \\ \hline
          \cite{pinto2016supersizing}  & RGB & 15  & 150 & 66.0
            \\ \hline

           \cite{johns2016deep}  & D
      & 20  & 100 & 80.3

      \\ \hline
          \cite{watson2017real}  &  M  & 10  & -   & 62.0
            \\ \hline
          \cite{mahler2017dex}   & D  & 10  & 50  & 80.0
            \\ \hline
           \cite{asif2017rgb}  & M
      & 35  & 134   & 91.9

      \\ \hline

         \cite{viereck2017learning}  & D
      & 10  & 40   & 97.5

      \\ \hline

         \cite{morrison2018closing} & D  & 20  & 200   & 87.0
            \\ \hline
         \cite{Chu_Xu_Vela_2018}  &  M  & 10  & 100   & 89.0
          \\ \hline
          \cite{satish2019onpolicy}  & D  & 8 & 80    & 87.5
            \\ \hline
          \cite{asif2019densely} & M   & -  & $>$200   & 89.0/90.0\tnote{**}\rule{0pt}{2.2ex}\rule{0pt}{2.2ex}
                             \\ \hline
          \cite{lu2020planning}   & M  & 10  &  30  & 84.0
                           \\ \hline
          \cite{wang2021double}   & RGB  &  43  & 129  & 93.0
            \\ \hline

           \cite{xu2021gknet}  & M  & 30  & 300  & 95.3$\pm$2.4
          \\ \hline
           {\bf 3D Grasp} &  &  &       &
          \\ \hline

           \cite{kopicki2016one}  & PC
      & 45  & 45 & 77.8 
                            \\ \hline 
            
          \cite{ten2018using}  & PC
      & 30  & 214 & 85.0

             \\ \hline
         \cite{liang2019pointnetgpd}   & PC  & 10  & 100   & 82.0
               \\ \hline
          \cite{mousavian20196} & PC & 17 & 51    & 88.0

                       \\ \hline
          \cite{lou2020learning}  & PC &       5  & 50   & 82.5

                     \\ \hline
          \cite{wu2020grasp}  & PC &     20  & 60   & 85.0


        \\ \hline 
          {\PS} v1 & D+PC  &  10  & 100  & 94.0
                           \\ \hline 
         \cite{lundell2021multi}  & M  &  10  & -  & 60.0
                                              \\ \hline 
            \hline 
          {\PS} v2  &  D+PC  & 33\tnote{$\dagger$} & 330  & 94.2$\pm$3.3
            \\ \hline 
        \end{tabular} 
        \begin{tablenotes} 
          \footnotesize 
          \item[] M: RGB-D; D: Depth; PC: Point cloud.
          \item[*] Success rate of 84\%\,/\,89\% achieved on Baxter\,/\,PR2 robot.
           \item[**] Success rate of 89\%\,/\,90\% achieved on ResNet and DenseNet.
        \item[$\dagger$] Combines the unique objects from Tables
        \ref{tab:exp_known}, \ref{tab:exp_chu}, \ref{tab:exp_morrison}, and \ref{tab:exp_task_oriented}.
        \end{tablenotes}
        \end{threeparttable}
        \vspace*{-0.15in}
      \end{table}

\cite{xu2021gknet} and~\cite{wang2021double} focused on a compact grasp
representation consisting of a left-right keypoint pair. Both achieved a
good balance between accuracy and speed, which suggests an advantage of
this smaller representation over the traditional 5-dimensional
rectangle representation of 2D grasp methods.  The earlier comparative
analysis between GKNet and {\PS} v2 suggests that combining the two
methods should improve performance.  In doing so, the keypoint
representation may result in a faster processing pipeline since it would
replace the current pose fitting process.



From Table~\ref{tab:exp_comparison}, recent methods in the 2D Grasp
group have a better success rate (over 90\%) than methods in the 3D
Grasp group (below 90\%), with the exception of {\PS}. The majority of
the methods in the 3D Grasp group involve grasp sampling and anti-podal
scoring methods.  The reduced performance relative to {\PS} suggests
that the grasp sampling strategy may not be densely sampling the
feasible grasp pose space. Evidence for this is comes from the top
performing 2D method \cite{viereck2017learning}, which employs
randomized grasp sampling. 
Moving from 2D to 3D grasp sampling induces a performance gap for 
the 3D Grasp methods.  {\PS} uses pre-determined grasp
families from which grasps are densely sampled in the grasp family's
parameter space.  It also operates with a performance comparable to top
single-view 2D Grasp methods.  Interestingly, these 2D Grasp methods are
based on hypothesizing candidate anti-podal pairs
\cite{xu2021gknet,wang2021double} as opposed to scoring randomly sampled
grasp poses, or on scoring grasps when given the equivalent of an
anti-podal pair.  Incorporating these techniques into the {\PS} network
as a separate grasp prediction and evaluation branch may support
enhanced grasping.

Overall, {\PS} v2 achieves a success rate of 94.2\% and is amongst the
top performing methods.  Testing over a variety of objects validates the
generality and consistency of the {\PS} pipeline. The success rate confirms
the hypothesis that explicit shape recovery can inform how objects should be
grasped.  {\PS} uses deep learning to segment the scene, after which the
remaining pipeline follows traditional point-cloud, model-based fitting and
processing approaches. Consequently, there is an opportunity to include
additional deep learning components for predicting grasp poses conditioned
on the segmentation masks, and to include grasp quality scoring.

\begin{figure}[t]
  \centering
  \includegraphics[width=\columnwidth,clip=true,trim=0in 0.75in 0in 2.25in]{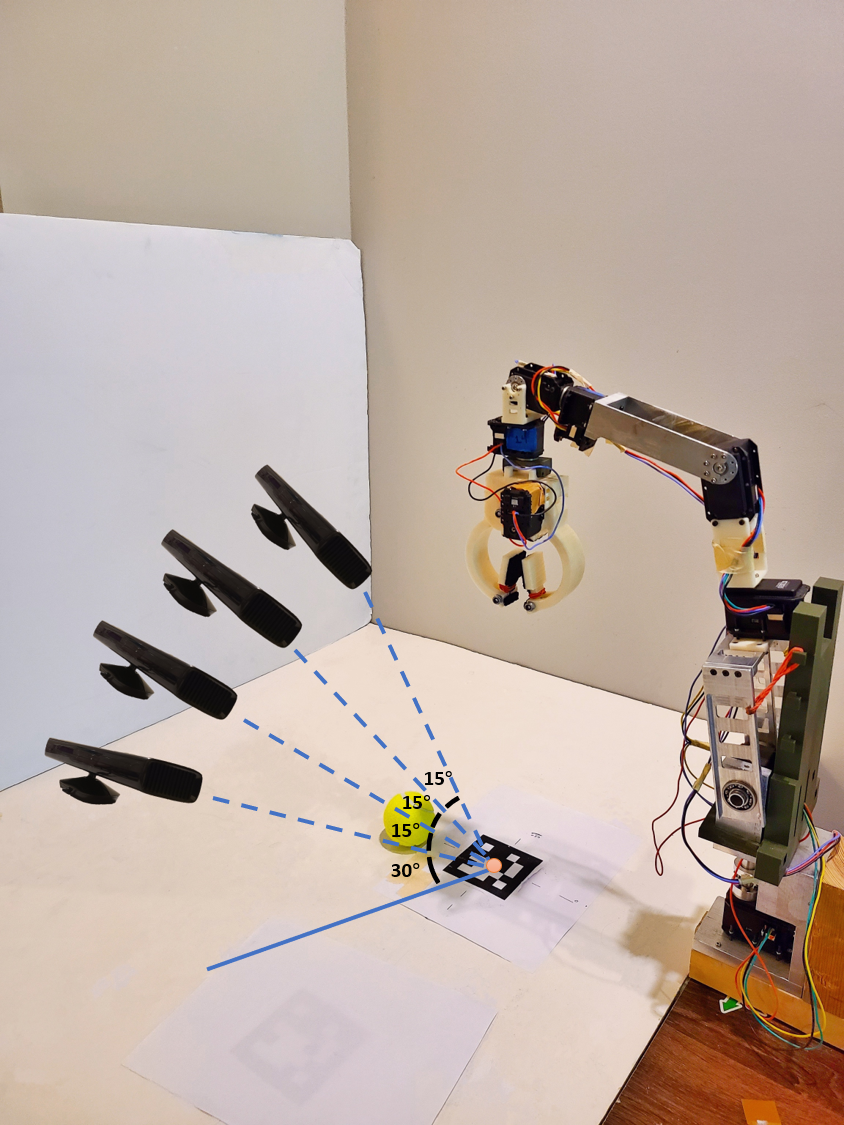}
  \caption{Experimental setup for grasping at varied camera angles (side view), where objects are tested under 30$\degree$, 45$\degree$, 60$\degree$ and 75$\degree$, respectively. 
  \label{fig:exp_variedangle}}
\end{figure}

\subsection{Grasping at Varied Camera Angles}

The {\PS} v2 deployment assumes a particular viewpoint similar to that of a
human looking down towards a manipulation workspace. The training model
included randomization for robustness to viewpoint variance, since it is
not always the case that the a given deployment scenarion can exactly
match the training scenario.
This section evaluates grasping consistency of {\PS} v2 under varied camera
angles.



\subsubsection{Setup.}
This experiment focuses on grasping a single isolated object under different
camera viewing angles. As is shown in Figure~\ref{fig:exp_variedangle},
the view angles vary from 30$\degree$ to 75$\degree$ in 15$\degree$
increments, while the distance from the object to the camera frame is
kept constant across the experiments. The view angle does not go below
30$\degree$ since the effective working area of the Kinect would be too
small under such settings. We do not test objects above 75$\degree$
because the viewpoint fails to capture sufficient information regarding
the sides of objects, which means that shape is difficult to infer and
the limited point cloud information leads to degraded model fitting.
The six objects tested match to different primitive shape classes. 

\subsubsection{Outcome Analysis and Discussion.}
As seen in Table \ref{tab:exp_angle}, {\PS} v2 grasping success rates
for 30$^\circ$ and 60$^\circ$ lie within the 95\% confidence interval of
the 45$^\circ$ case, which demonstrates that {\PS} v2 has robustness to
the camera view within $45^\circ \pm 15^\circ$.  Recall that the
principle angle of the camera is set to be 42.3$\degree$ in the
simulation environment (Section~\ref{datagen}).  The 75$^\circ$ case
lies outside of the 95\% confidence interval, indicating a
non-negligible reduction in performance.  Nevertheless, performance
continues to be competitive relative to methods found in Table 
\ref{tab:exp_comparison}.
The performance drop is mainly
due to the wrench, which is a flat and thin object lying on the ground.
This case is illustrated for a pair of scissors in
Figure~\ref{fig:exp_failure_scissor}, where the depth sensor cannot retrieve
a good depth map due to depth resolution limits.  Depth-based approaches
may benefit from precise estimation and isolation of the tabletop plane
from the target object so that short or flat objects on the surface can
be distinguished.

\begin {table}[t]
  \centering
  \caption {Grasping at varied camera angles with 95\%
confidence intervals on our household set\label{tab:exp_angle}}
  \begin{tabular}{ | l | c | c | c | c |}
    \hline
    \multirow{2}{*}{\bf{object}} & \multicolumn{4}{c|}{\bf{Accuracy (\%)}}           \\ \cline{2-5}
            & 30$\degree$ & 45$\degree$ & 60$\degree$ & 75$\degree$ \\ \hline 
 Bowl       &  10/10    &   9/10   &  10/10    &  10/10    \\ \hline 
 Tape       &  10/10    &  10/10   &  10/10    &  10/10    \\ \hline 
 Juice box  &  10/10    &  10/10   &   9/10    &   9/10    \\ \hline
 Wrench     &  10/10    &  10/10   &  10/10    &   8/10    \\ \hline
 Cup        &   9/10    &  10/10   &  10/10    &   9/10    \\ \hline
 Tennis ball&   9/10    &  10/10   &  10/10    &  10/10    \\ \hline
 \hline
 \multirow{2}{*}{Average (\%)}   & 96.7  &  98.3 &  98.3 & 93.3\\   
      &   \ $\pm$ 4.5 & \ $\pm$ 3.3 & \ $\pm$ 3.3   &  \ $\pm$ 6.3
 \\\hline
  \end{tabular}
\end {table}

\subsection{Multi-Object Grasping}

To gauge how effective the PS-CNN pipeline is at grasping in clutter, a
picking task was implemented with two modes denoted: light clutter and heavy
clutter. Based on the training data and grasp scoring mechanism, light
clutter should not significantly impact success rate, however heavy
clutter might.  The picking task setup was modeled after existing
published works to allow for comparative evaluation. 

\subsubsection{Setup.}
\begin{figure}[t]
    \centering
    \begin{tikzpicture}[inner sep=0pt,outer sep=0pt]
    \node[anchor=south west] (A) at (0in,0in)
      {\includegraphics[width=\columnwidth,clip=true,trim=0in 5in 0in 5in]{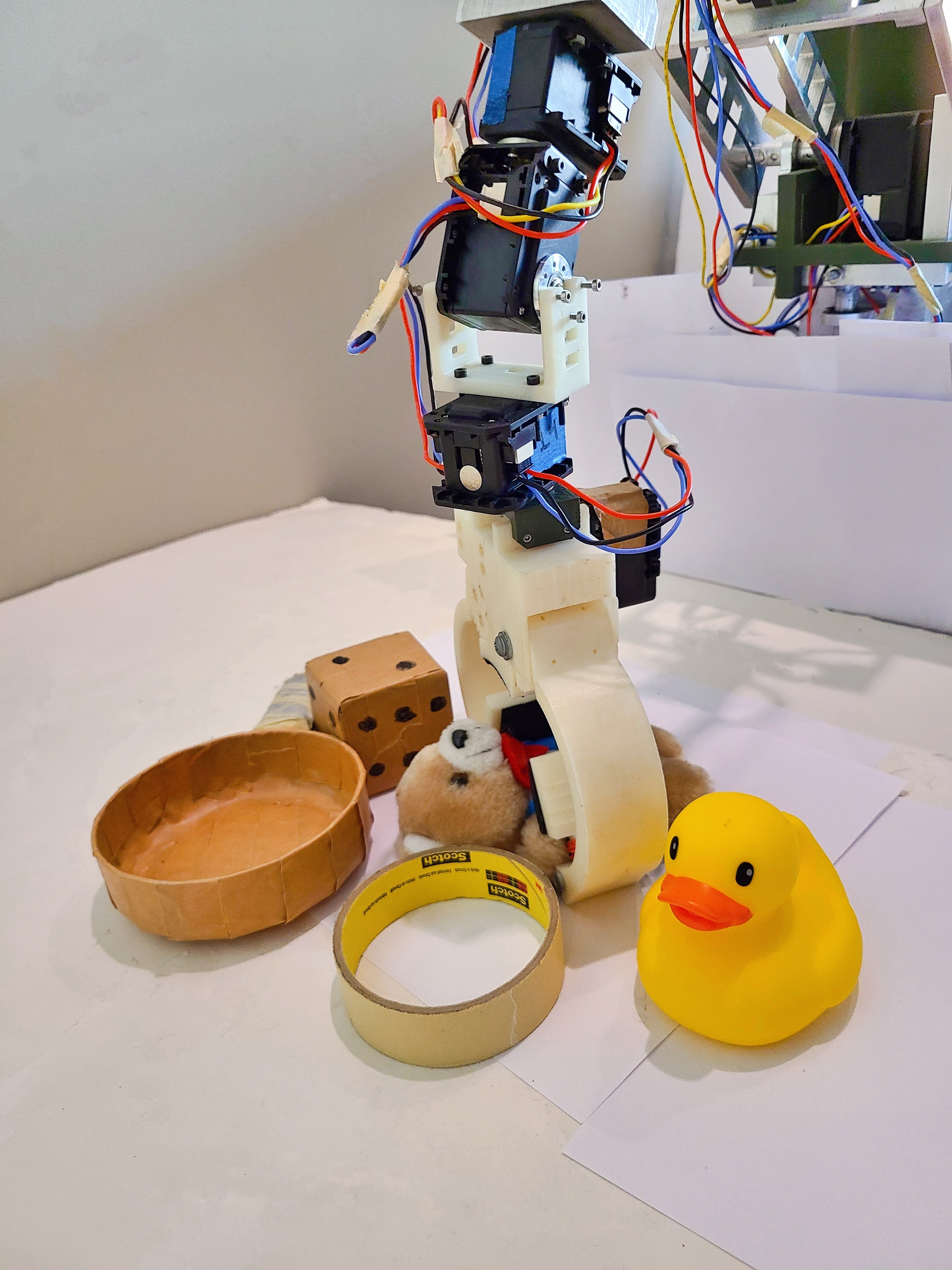}};
    \node[anchor=north west,yshift=1pt,xshift=-1pt,inner sep=1pt,fill=white] at (A.north west)
      {\includegraphics[width=0.4\columnwidth,clip=true,trim=3.5in 2.25in 1.75in 2in]
       {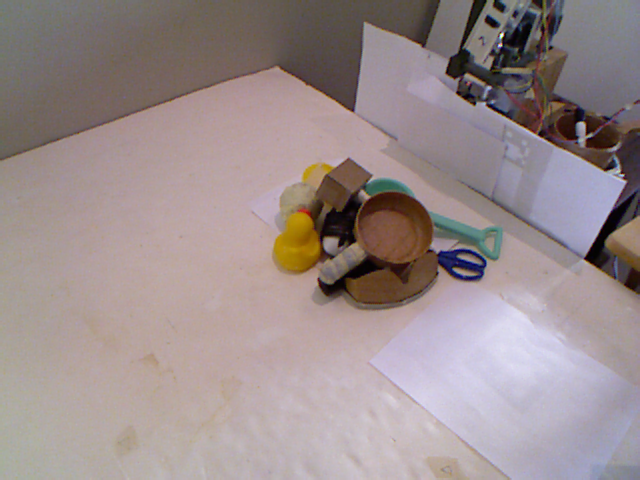}};
    \end{tikzpicture}
    \caption{
    Setup of the multi-object grasping experiment.  Objects are placed close
    to each other in the depicted 5 object, light clutter case. The top-left
    inset image shows the 10 object, heavy clutter case where objects are
    piled.
    \label{fig:exp_multi}}
\end{figure}

From the set of 31 unique objects of the Multi-object Grasping Set
depicted in Figure \ref{fig:obj_set}(e), a smaller set of 5 (or 10)
objects are randomly selected and placed on the workspace close to each
other with random poses, see Figure \ref{fig:exp_multi}.
For each grasp attempt trial, PS-CNN will first capture a depth
image of the scene as the input and then predict the 
{top-1 grasp candidate for each shape.}  Afterwards, the
robot will only execute the grasp command with the highest grasp score
{across the scene.} Following \cite{ni2019new}, the robot has $n+2$
grasp attempts to remove the objects The experiment terminates when all
objects have been picked or the robot has tried 7 (or 12) times. 

\subsubsection{Evaluation.}
The Multi-Object Grasping experiment involves three evaluation metrics,
including: (1) {\bf Success Rate (S)}: the percentage of grasp attempts
that successfully grasp the target; (2) {\bf Object Clearance Percentage
(OC)}: the ratio of number of objects has been grasped successfully to
the total number of objects in the scene; (3) {\bf Completion percentage
(C)}: marked as 1 if all the objects in the scene have been grasped
successfully or 0 if any object was not picked.  Table
\ref{tab:multi_object} compares these statistics to other published
works. There are some differences regarding trial termination criteria
(TC), with $k$G signifying up to $k$ grasp attempts,  $+k$ signifying
$n+k$ allowed grasps for $n$ objects, and  $k$Seq meaning $k$ sequential
failures.  Ordering of the results is from least strict to strictest
termination criteria, in terms of permitting the most amount of
incorrect grasps before termination.

\begin{table*}[t]
\centering
\begin{threeparttable} 
  \caption{Multi-Object Grasping Comparison \label{tab:multi_object}}
  \begin{tabular}{|l|c|c|c|c|c|c|c|}
    \hline
    Method & \#Obj. & \#Sel. & \#Trials & TC\tnote{*}\rule{0pt}{2.2ex} &
    S\tnote{\#}\rule{0pt}{2.2ex} &
    \multicolumn{1}{|l|}{OC\tnote{$\dagger$}\rule{0pt}{2.2ex}} &
    \multicolumn{1}{|l|}{C\tnote{$\ddagger$}\rule{0pt}{2.2ex}} \\ \hline
    \cite{pinto2016supersizing}   
      & 21 & 10 & 5  & None & 38.0 & 100.0 & 100.0 \\ \hline
    
    \multirow{2}{*}{\cite{mahler2017dex} } 
      & 25 & 5  & 20 & 5Seq & 92.0 & 100.0 & 100.0 \\  
      \cline{2-8}& 25 & 10  & 10  & 5Seq & 83.0 & 98.0 &  - \\  \hline
    \multirow{2}{*}{\cite{mahler2017learning} } 
      & 25 & 5  & 20 & 5Seq & 94.0 & 100.0 & 100.0 \\ 
      \cline{2-8}
      & 25 & 10 & 10 & 5Seq & 89.0 & 100.0 & 100.0\\ \hline 
    \multirow{2}{*}{\cite{xu2021gknet} } 
      & 30 & 5  & 20 & 5Seq & 92.1 & 100.0 & 100.0 \\ 
      \cline{2-8}
      & 30 & 10 & 10 & 5Seq & 72.1 & 98.0 & 80.0\\ \hline 
    \cite{gualtieri2016high} 
      & 10 & 10 & 15 & 3Seq & 84.0 & 77.0 & - \\ \hline

    \cite{levine2018learning}    
      & 25 & 25 & 4  & 31G  & 82.1 & 99.0 & 75.0 \\ \hline

    \cite{ni2019new}  
      & 16 & 8  & 15 & +2 & 86.1 & 87.5 & - \\  \hline
     {\PS} v1 (2020) & 10 & 5 & 10  & +2  & 88.9 & 96.0  & 80.0 \\ \hline 
           \cite{corsaro2021learning}    
      & 29 & 13 & 10  & +2  & 80.8 & 80.8 & - \\ \hline
      \hline
      \multirow{2}{*}{{\PS} v2} & 31 & 5 & 20  & +2  & 93.5 & 100.0  & 100.0 \\ 
     \cline{2-8}& 31 & 10 & 5  & +2  & 82.8 & 96.0  &  60.0 \\ \hline
  \end{tabular}
  \vspace*{-1.5em}
  \begin{multicols}{2}
  \footnotesize
  \begin{tablenotes}
    \item[*] TC = Termination Criteria.
    \item[\#] S = Success   Rate (\%).
  \end{tablenotes}
\columnbreak
  \begin{tablenotes}
    \item[$\dagger$] OC = Object Clearance Percentage (\%).
    \item[$\ddagger$] C = Completion percentage (\%).
  \end{tablenotes}
\end{multicols}
  \end{threeparttable}
\end{table*}

\subsubsection{Light Clutter Grasping (5 Objects).}
As recorded in Table \ref{tab:multi_object}, the performance outcomes of
{\PS} v2 improved relative to v1. Importantly the object clearance and
the completion percentage increased to 100\%, indicating success for all
of the object picking and clearing trials. 
Referring to the isolated object grasping results in Table
\ref{tab:exp_comparison} as a normative grasp success rate, the drop in
grasp success rate under clutter for {\PS} v2 is 1.4\%, which indicates
that the method is robust to light clutter.  The grasp success rate lies
within the 95\% confidence interval of the single object case.
Looking at the other rows, {\PS} v2 performs as well as the top
performing grasping methods noted.  It matches the performance of
\cite{mahler2017dex} and \cite{mahler2017learning}, also known as DexNet
1.0 and DexNet 2.0.  These two methods were explicitly designed to
handle grasping in cluttered scenarios and are sampled-based, deep
network grasp quality scoring methods with a top-down grasping
assumption.  In contrast, {\PS} v2 employs more traditional grasp scoring,
which indicates that the primitive shape segmentation does a good job at
isolating distinct regions likely to have good candidate grasp options.
{\PS} v2 performs similar to, but slightly higher than, GKNet
\citep{xu2021gknet}. GKNet is a 2D grasp prediction strategy trained on
manually annotated data of single objects. It employed a grasp scoring
method similar to that of {\PS} v2. These two methods perform comparably
when tasked to grasp in clutter.  As noted in Section
\ref{sec:res_static}, a composite GKNet+PS grasping strategy may provide
complementary strengths and boost performance.



\subsubsection{Heavy Clutter Grasping (10 Objects).}
As a stress test, picking tests with 10 objects piled together were also
performed, see Figure \ref{fig:exp_multi} inset. The footprint of the pile
was approximately equal to the footprint of the five Light Clutter objects.
The DexNet variants \citep{mahler2017dex,mahler2017learning}
should continue to be an upper bound reference on performance
since their deep networks are trained to handle cluttered scenes. The
performance drop for {\PS} v2 is more acute here and lies outside of the
95\% confidence interval for single object grasping. The grasping
pipeline starts to break down. One reason is that the point clouds for
the primitive shape regions are sparser under heavy clutter versus
light clutter, due to occlusion. The shape fitting algorithm starts to
exhibit sensitivity to the point cloud data and the model estimates lose
accuracy.  In comparison to other methods with the $n+2$ termination
criteria \citep{ni2019new,corsaro2021learning}, {\PS} v2 performs well. 
It has a higher object clearance rate compared to these two strategies
while having a grasping success rate in between the two methods. Both
are grasp sampling and scoring methods. {\PS} v2 performs close to
DexNet 1.0 \citep{mahler2017dex} which provides evidence that shape can
serve as a strong prior for where to look for grasp candidates, even in
clutter. However, it has a clear performance gap relative to DexNet 2.0
\citep{mahler2017learning} indicating that there is a limit to shape
alone when occlusion and clutter begin to reduce the information content
and region continuity of primitive shape data in the depth image.
Incorporating scoring methods robust to clutter, as a separate branch or
subsequent process, should improve performance.
We could not test the combination of PS-CNN with a downstream grasp
quality CNN (GQ-CNN), such as DexNet. 
The publicly available version of Dex-Net does not provide the API to do so. 

Lastly, one other source of picking failure was the actual gripper
geometry versus the grasp scoring collision geometry.  
Rather than being a parallel-plate gripper, it has a jaw-like
design, as captured in Figure \ref{fig:exp_gripper}(a). 
The swept volume of the gripper when closing is not perfectly modeled by
the grasp ranking algorithm, resulting in false positives and false
negatives. The former lead to the gripper colliding with other objects
during execution, while the latter reject good candidates in favor or
poorer candidates. The value of DexNet-like learning methods is the
existence of training data based on the actual gripper geometry that
better captures its effect on grasping performance.


%
%
\begin{figure*}[t]
  \centering
  \begin{tikzpicture} [outer sep=0pt, inner sep=0pt]
  \scope[nodes={inner sep=0,outer sep=0}] 
  \node[anchor=north west] (a) at (0in,0in) 
    {\includegraphics[height=3.4cm,clip=true,trim=2.5in 1.5in 0.5in 0.5in]{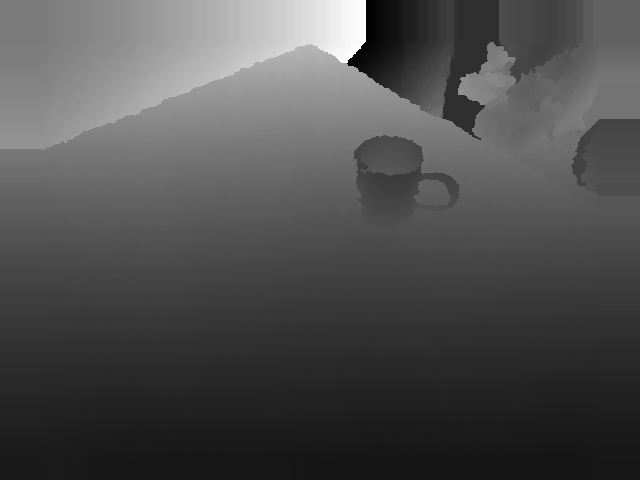}}; 
  \node[anchor=north west, xshift=0.1cm] (b) at (a.north east)
    {{\includegraphics[height=3.4cm,clip=true,trim=2.5in 1.5in 0.5in 0.5in]{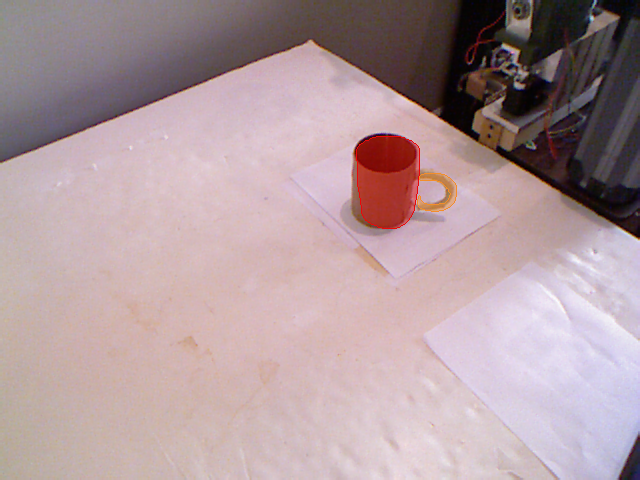}}}; 
  \node[anchor=north west, xshift=0.1cm] (c) at (b.north east)
  {\includegraphics[height=3.4cm,clip=true,trim=2.8in 1.0in 0.3in 1.0in]{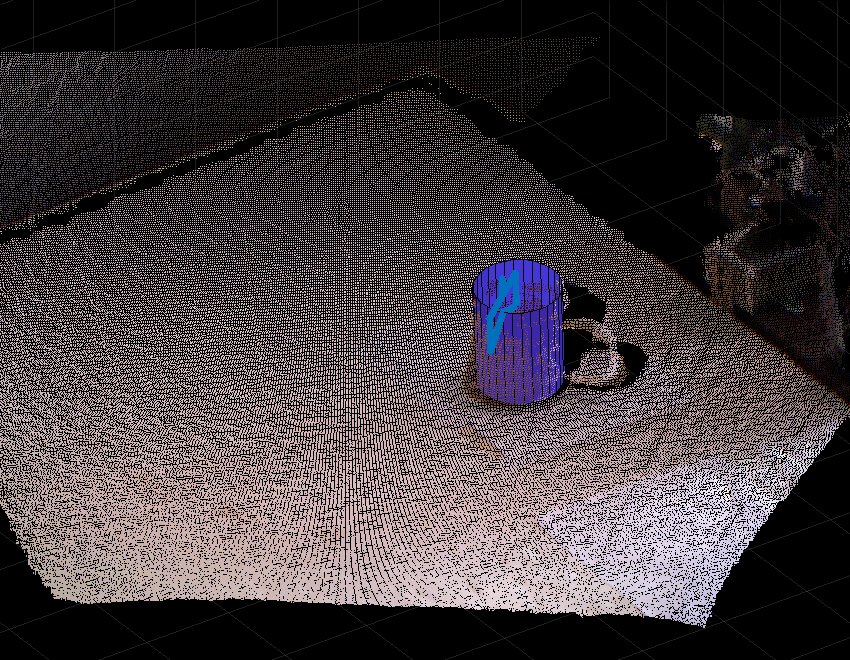}}; 
  \node[anchor=north west, xshift=0.1cm] (d) at (c.north east)
  {\includegraphics[height=3.4cm,clip=true,trim=2.8in 1.0in 0.3in 1.0in]{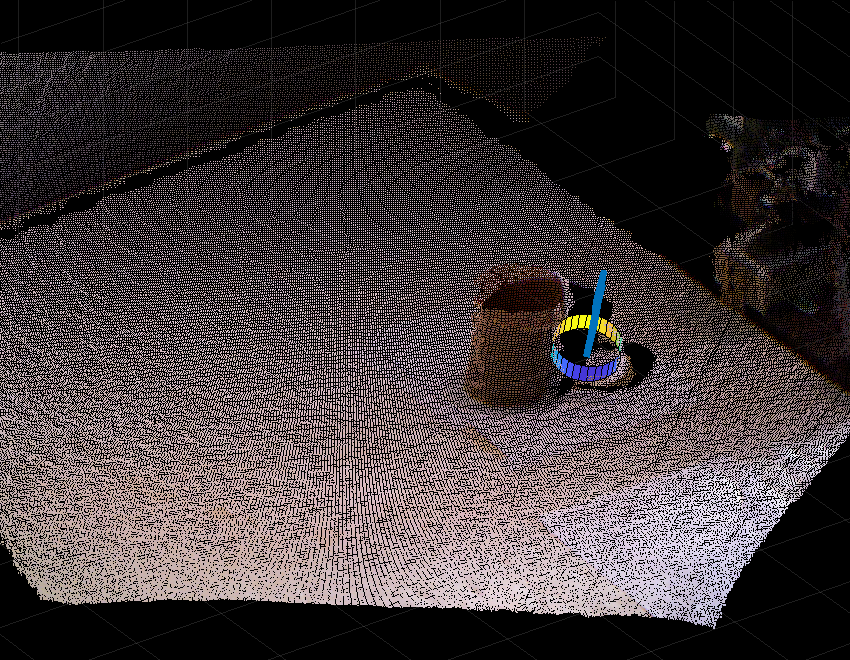}}; 
 
  \endscope

  \foreach \n in {a,b,c,d} 
   {
    \node[anchor=north west,fill=white,draw=white,inner sep=1pt] at (\n.north west) {(\n)};
   }
  \end{tikzpicture}
  \vspace*{-1.0ex}
  \caption{
    Task-oriented grasping. (a) Depth input. (b) Segmented mask. (c)
    Primitive shape model and grasp candidate on main body. (d)
    Primitive shape model and grasp candidate on handle.
    \label{fig:exp_task-oriented_mug}}
\end{figure*}

\begin{table*}[t]
  \caption{Task-Free and Task-Oriented Grasping with 95\%
confidence intervals \label{tab:exp_task_oriented}}
  \centering
     \begin{tabular}{|l|c|c|c||c|c|}
    \hline
    \multirow{2}{*}{Target Objects}
    \rule{0pt}{2ex} & \multicolumn{3}{|c|}{Task-Free} &
    \multicolumn{2}{|c|}{Task-Oriented} \\
    \cline{2-6}\rule{0pt}{2ex}
      & \cite{jain2016grasp} & {\PS} v1 & {\PS} v2
       & {\PS} v1 & {\PS} v2  \\ \hline
    \hline
    Mug         & 6/10   & 9/10  & 10/10  &  9/10  & 10/10 \\ \hline
    Pot         & 3/10   & 10/10 & 10/10 &  8/10  & 10/10  \\ \hline
    Pan         & 5/10   & 10/10 & 10/10  &  8/10  &  9/10 \\ \hline
    Brush       & 5/10   & 6/10 & 10/10  &  4/10  &  9/10 \\ \hline
    Handbag     & 9/10   & 10/10 & 10/10   & 8/10  & 10/10 \\ \hline
    Hammer      & 9/10   & 9/10 & 10/10   & 7/10  & 9/10 \\ \hline
    Basket      & 6/10   & 10/10 & 10/10   & 8/10  & 9/10 \\ \hline
    Bucket      & 7/10   & 8/10 & 9/10 & 8/10  & 8/10 \\ \hline
    Shovel      & 8/10   & 10/10 & 10/10  & 9/10  & 10/10 \\ \hline
    Net spoon   & 7/10   & 10/10 & 9/10   & 8/10  & 8/10 \\ \hline
    \hline
    Average (\%)  
    & 65.0 $\pm$ 9.3 
    & 92.0 $\pm$ 5.3 
    & 97.0 $\pm$ 3.3
    & 73.0 $\pm$ 8.7
    & 93.0 $\pm$ 4.9
    \rule{0pt}{2ex} \\ \hline
             
  \end{tabular}
  
\end{table*}

\subsection{Task-Oriented Grasping}
\label{sec:res_taskoriented}
The added value of segmenting objects according to shape is that the
distinct shape regions may correspond to grasp preferences based on the
task sought to accomplish.  All of the previous experiments discussed
focus on task-free grasping (or simply pick-n-place operations). This is
because the methods do not differentiate object regions. Awareness of
shape moves grasp selection one step closer to semantically meaningful
and task appropriate grasping.

In this set of tests, each connected primitive shape region for the
given object is presumed to correspond to a specific functional part. 
Grasping tests will compare task-free grasping to task-oriented grasping. 
In task-oriented grasping a specific region must be grasped based on the
functionality of the object being grasped.  
Figure \ref{fig:exp_task-oriented_mug} depicts the depth and color images
of a mug.  Task-free grasping would focus any graspable part of a target
object, which would include the container part and the handle.
Task-oriented grasping would prefer to grasp the handle, Figure
\ref{fig:exp_task-oriented_mug}(d), behaving like humans would
under certain use cases.

\subsubsection{Setup.}
The methodology is similar to the static object grasping experiment
(Section~\ref{sec:res_static}).
The test set objects, depicted in Figure \ref{fig:obj_set}(f),
each consist of more than one functional part or primitive shape
category, one of which is defined to be the task-oriented preference
when grasping.  

\subsubsection{Outcome Analysis and Discussion.}
Only \cite{jain2016grasp} and {\PS} v1 were tested as baselines since
they are shape-based.  Table \ref{tab:exp_task_oriented} reports the
performance of the systems on the grasping tests. The method in
\cite{jain2016grasp} simply approximates each object with a single
primitive shape. which renders task-oriented grasping inapplicable.
The inability to capture all of the primitive shape sub-components in
the objects tested degraded this method's performance relative to the
case when objects were uniquely a single primitive shape class, see Table
\ref{tab:exp_known} (a 6.7\% drop in success rate).
In contrast {\PS} v1 and v2 performed closer to their nominal
performance for single primitive shape class objects. This demonstrates
the ability of the {\PS} pipeline to capture the primitive shapes and
select the best one to use for grasping based on the specified scoring
mechanism. 

\begin{figure}[t]
  \centering
  \begin{tikzpicture} [outer sep=0pt, inner sep=0pt]
  \scope[nodes={inner sep=0,outer sep=0}] 
  \node[anchor=north west] (a) at (0in,0in) 
    {\includegraphics[width=4.1cm,clip=true,trim=3.3in 3in 2.5in 1in]{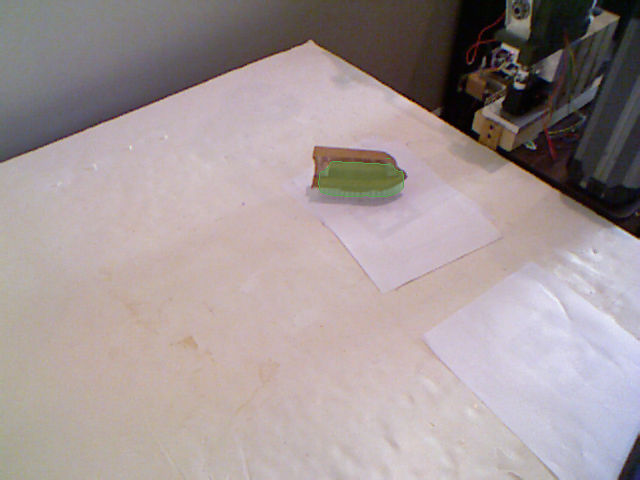}}; 
  \node[anchor=north west, xshift=0.1cm] (b) at (a.north east)
    {{\includegraphics[width=4.1cm,clip=true,trim=3.3in 3in 2.5in 1in]{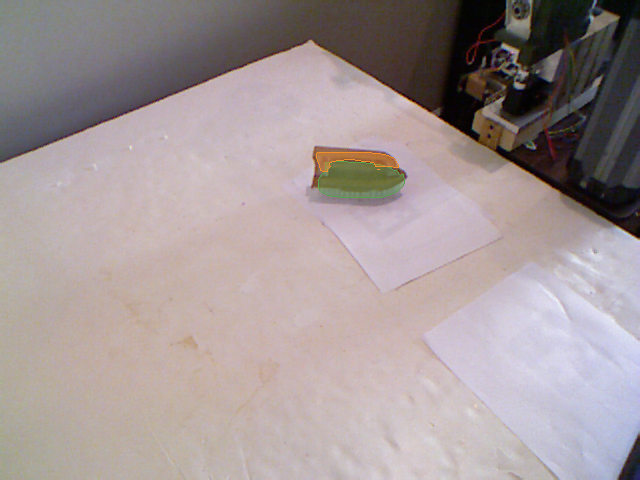}}}; 
  \node[anchor=north west, yshift=-0.1cm] (c) at (a.south west)
  {\includegraphics[width=4.1cm,clip=true,trim=3.4in 1.5in 2.3in 1.5in]{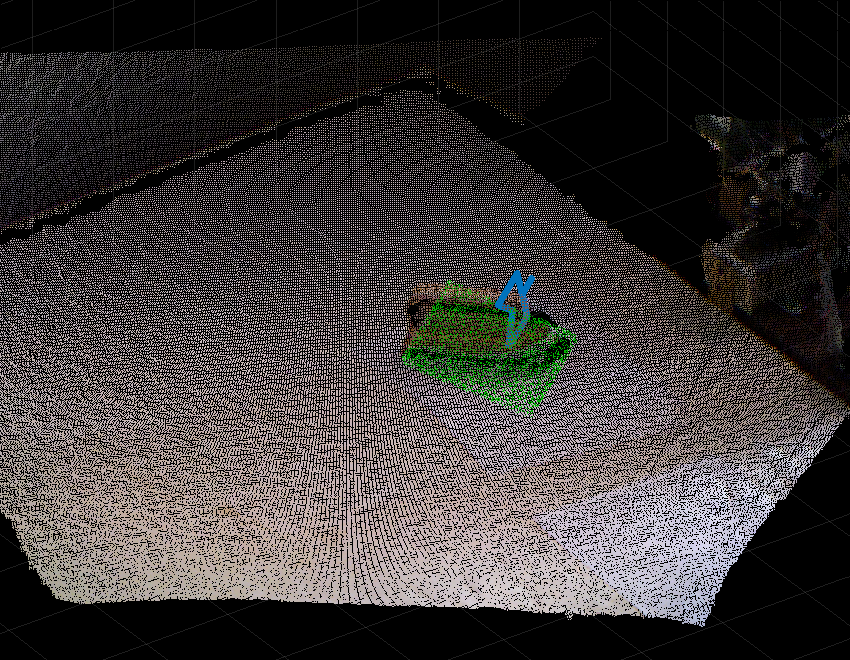}}; 
  \node[anchor=north west, xshift=0.1cm] (d) at (c.north east)
  {\includegraphics[width=4.1cm,clip=true,trim=3.4in 1.5in 2.3in 1.5in]{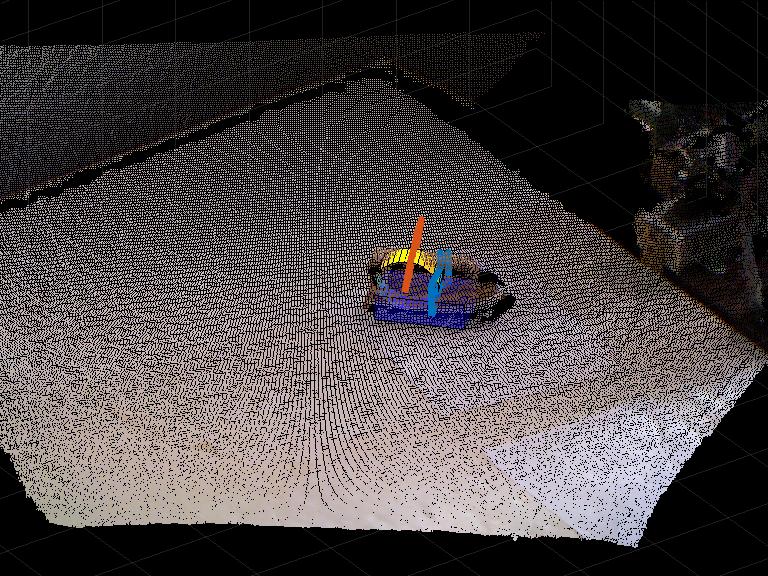}}; 
  \endscope
  \foreach \n in {a,b,c,d} 
   {
    \node[anchor=north west,fill=white,draw=white,inner sep=1pt] at (\n.north west) {(\n)};
   }
  \end{tikzpicture}
  \caption{
    Comparison between {\PS} v1 and v2 on the brush. 
    (a) Segmentation from v1. 
    (b) Segmentation from v2. 
    (c) Top grasp candidate of v1. 
    (d) Top grasp candidates of v2.
    \label{fig:exp_task-oriented_brush}} \end{figure}

When moving to the task-oriented case {\PS} v2 has a clear advantage
over {\PS} v1. Not only is the task-free success rate higher, but the
performance drop in the task-oriented case is lower, at a 4\% drop
versus a 19\% drop. The improved training corpus and upgraded shape
fitting module contributed to the performance boost. The difference is
best observed for the brush test object, which had the largest
improvement for the task-free and task-oriented cases.  
Figure \ref{fig:exp_task-oriented_brush}(a,c) shows that
{\PS} v1 failed to distinguish the handle and the base of the brush 
while {\PS} v2 could differentiate the two regions. 
The improvements suggest that more research placed on the precision
grasping of specific object parts or regions can contribute to improved
general purpose grasping.  Additionally, the ability to achieve
task-oriented grasping through shape differentiation permits follow-up
investigations into realizing more advanced semantic grasping.

%
%
\section{Conclusion}
This paper investigated the value of primitive shape awareness for robotic
grasping.  To that end, it described a segmentation-based shape primitive
grasping pipeline that leverages recent advances in deep learning to
decompose objects into multiple primitive shapes regions then recover their
corresponding shape models. Prior knowledge of the shape primitive permits
dense generation of grasp configurations from shape-specific, parametrized
grasp families. Once the primitive shape object regions are known, the
pipeline employs classical grasping paradigms to filter, rank, and execute
the grasp operation. The primitive shapes deep network ({\PS}) shows that
high-performance grasp candidates can be learned from simulated visual data
of primitive shapes without collecting a large-scale dataset of CAD models
and simulating grasp attempts.

In static object grasping experiments, {\PS}-enabled grasping achieves
a 94.2\% grasp success rate amongst the top-performing methods. 
Multi-object grasping in light clutter (5 objects per scene) has
a 93.5\% grasp success rate and a 100\% completion rate.
Grasping trials with the camera set at varied angles quantifies robustness
to view angles. 
These outcomes indicate that {\PS} successfully bridges the gap between the
synthetic primitive shape training data and real-world test objects, showing
generalizability to different object types or forms. 
Moreover, our segmentation-based approach facilitates task-specific grasping
on the objects composed of multiple shapes that have functional meaning or
purpose. It goes beyond simple pick-and-place testing since there is
semantic meaning behind grasp constraints. Grasping experiments show a 97\%
grasp success rate in task-free grasping and 93\% for task-oriented
grasping. More advanced tasks could be enabled through task-aware grasping
strategies, including human-robot interaction tasks.

{\PS} is a first attempt to use deep networks to explicitly extract an
object's primitive shape composition for grasping. 
Consequently, there is room to improve.
The first direction is regarding the performance gap between objects
matching to known primitive shapes versus objects with non-primitive or more
complex geometry. 
The second is regarding the trade-off between precision and speed, as 
{\PS} takes 2.69s to generate a grasp plan per object on average. 
The bottleneck lies in the RANSAC-based shape fitting part, where tens
of thousands of proposals are filtered and ranked. A better idea would be to
leverage the power of deep networks to predict rough pose and sizes as the
initialization. 
In doing so, these modifications have the potential to further improve
the grasping success rate by including related grasping tasks, such as
grasp prediction and scoring, as additional task-related deep network
branches.  

To promote exploration into primitive shapes and reproduction of outcomes,
the source code is publicly available \citep{graspPrimShape}.

\begin{funding}
This work supported in part by the National Science Foundation under Award \#1605228 and \#2026611. Any opinions, findings, and conclusions or recommendations expressed in this material are those of the author(s) and do not necessarily reflect the views of the National Science Foundation.
\end{funding}

\bibliographystyle{SageH}
\bibliography{main}

\end{document}